\documentclass[letterpaper]{article} 
\usepackage{aaai24}  
\usepackage{times}  
\usepackage{helvet}  
\usepackage{courier}  
\usepackage[hyphens]{url}  
\usepackage{graphicx} 
\usepackage{natbib}  
\usepackage{caption} 
\usepackage{algorithm}
\usepackage{algorithmic}

\usepackage{newfloat}
\usepackage{listings}
\DeclareCaptionStyle{ruled}{labelfont=normalfont,labelsep=colon,strut=off} 
\floatstyle{ruled}
\newfloat{listing}{tb}{lst}{}
\floatname{listing}{Listing}
\title{ColNeRF: Collaboration for Generalizable Sparse Input Neural Radiance Field}

\author{
    Zhangkai Ni\textsuperscript{\rm 1},
    Peiqi Yang\textsuperscript{\rm 1},
    Wenhan Yang\textsuperscript{\rm 2}, 
    Hanli Wang\textsuperscript{\rm 1}\thanks{Corresponding author},
    Lin Ma\textsuperscript{\rm 3},
    Sam Kwong\textsuperscript{\rm 4}
}
\affiliations{
    \textsuperscript{\rm 1} Tongji University
    \textsuperscript{\rm 2} Peng Cheng Laboratory
    \textsuperscript{\rm 3} Meituan 
    \textsuperscript{\rm 4} City University of Hong Kong \\
    \{zkni, 2233007, hanliwang\}@tongji.edu.cn, yangwh@pcl.ac.cn, forest.linma@gmail.com, cssamk@cityu.edu.hk

}

\usepackage{bibentry}
\usepackage{amsmath}
\usepackage{multirow}
\usepackage{hhline}
\usepackage{booktabs}
\usepackage{enumitem}
\usepackage{makecell}
\usepackage{colortbl}
\usepackage{amsfonts,amssymb}
\usepackage[table]{xcolor}
\usepackage{color}
\usepackage{diagbox}
\usepackage{tikz}
\usepackage{float}
\usepackage{subfigure}
\usetikzlibrary{decorations.fractals,spy}
\usepackage{tikz}
\usetikzlibrary{spy}

\begin{document}
\def\imgratio{0.85}

\maketitle

\begin{abstract}
Neural Radiance Fields (NeRF) have demonstrated impressive potential in synthesizing novel views from dense input, however, their effectiveness is challenged when dealing with sparse input. 
Existing approaches that incorporate additional depth or semantic supervision can alleviate this issue to an extent.
However, the process of supervision collection is not only costly but also potentially inaccurate, leading to poor performance and generalization ability in diverse scenarios.
In our work, we introduce a novel model: the Collaborative Neural Radiance Fields (ColNeRF) designed to work with sparse input. 
%
The collaboration in ColNeRF includes both the cooperation between sparse input images and the cooperation between the output of the neural radiation field.
Through this, we construct a novel collaborative module that aligns information from various views and meanwhile imposes self-supervised constraints to ensure multi-view consistency in both geometry and appearance.
A Collaborative Cross-View Volume Integration module (CCVI) is proposed to capture complex occlusions and implicitly infer the spatial location of objects.
Moreover, we introduce self-supervision of target rays projected in multiple directions to ensure geometric and color consistency in adjacent regions.
Benefiting from the collaboration at the input and output ends, ColNeRF is capable of capturing richer and more generalized scene representation, thereby facilitating higher-quality results of the novel view synthesis.
Our extensive experimental results demonstrate that ColNeRF outperforms state-of-the-art sparse input generalizable NeRF methods.
Furthermore, our approach exhibits superiority in fine-tuning towards adapting to new scenes, achieving competitive performance compared to per-scene optimized NeRF-based methods while significantly reducing computational costs.
Our code is available at: \url{https://github.com/eezkni/ColNeRF}.
\end{abstract}

\begin{figure}[t]
\centering
\includegraphics[width=0.95\linewidth]{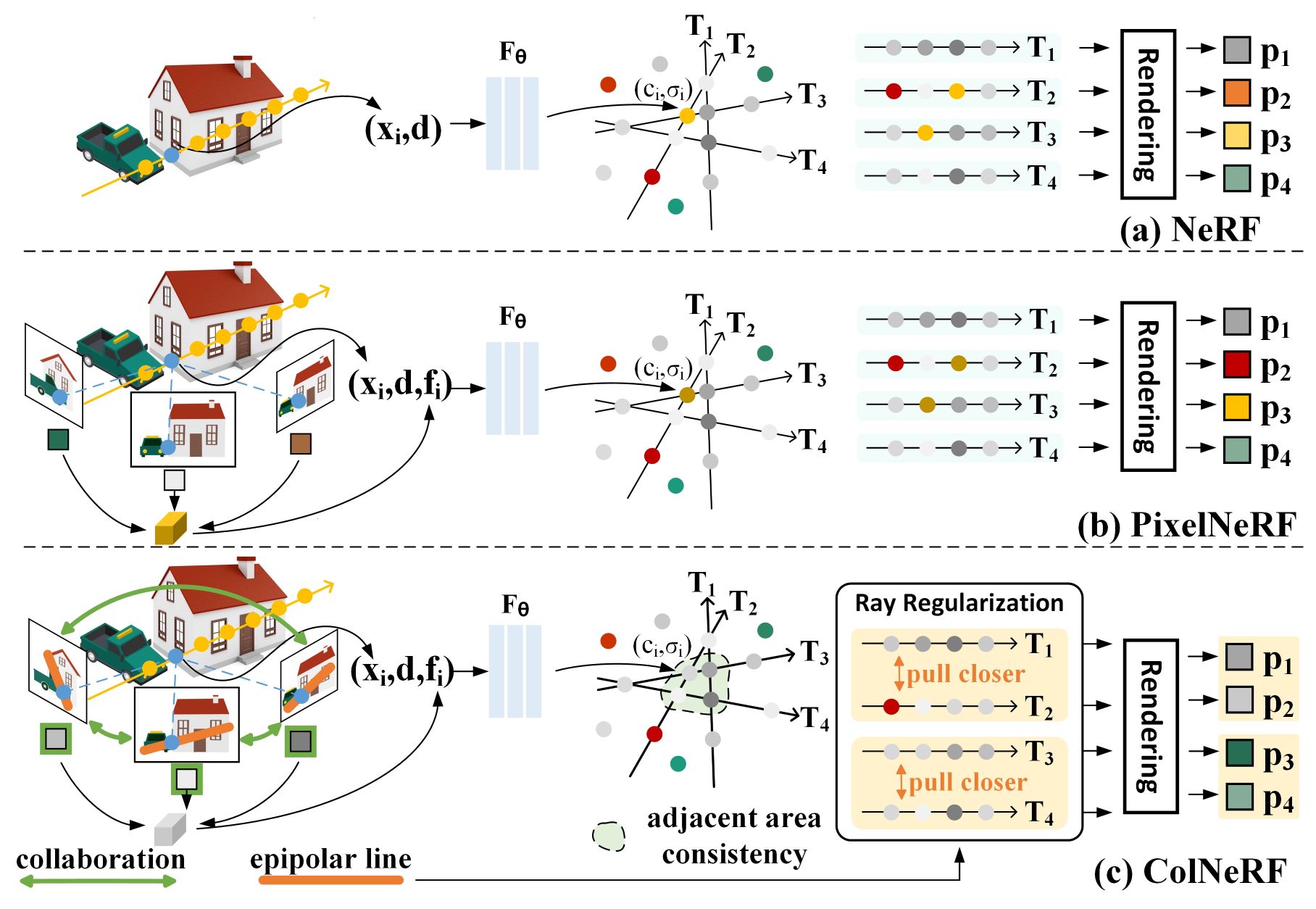}
\caption {
\textbf{Comparing previous approaches (a), (b) with our method (c)}:
Previous approaches heavily depend on a learned neural radiance field ($F_\theta$) for synthesis. However, these approaches result in undesirable outcomes with limited utilization of source view features and their interrelationships.
ColNeRF leverages collaboration at both input and output ends, providing richer supervision for training $F_\theta$.
}
\label{figure1}
\end{figure}

\section{Introduction}

Novel view synthesis aims to generate new view images of a scene based on a set of source images~\cite{zhu2018learning}. A prominent technique in this field is the Neural Radiance Field (NeRF)~\cite{mildenhall2021nerf}, which learns an implicit neural representation of the scene.
NeRF takes a 5D vector as input, comprising a 3D location ${\rm x} = (x,y,z)$ and a 2D viewing direction ${\rm d}=(\theta,\phi)$ for each point, and estimates the corresponding radiance value $(c,\sigma)$. The RGB value of a target pixel is then rendered by accumulating radiance from $N$ sampled points along the target ray.
NeRF brings revolutionary advancements to novel view synthesis, finding versatile applications in virtual reality, autonomous vehicles, robotics, and beyond.

The excellent performance of NeRF comes with a price, i.e. with a large amount of high-quality input source images used for training. However, acquiring a substantial number of RGB images along with their corresponding accurate camera parameters necessitates a complex process of calibrating. In real-world scenarios, not only is this difficult to execute, but the accuracy of the results obtained is questionable.
In scenarios where input images are limited, the novel view results generated by NeRF are degraded due to the lack of dense supervision.
Moreover, the optimization of NeRF is typically conducted independently for each scene, resulting in notable time inefficiency.
Significant research efforts have been put into addressing these issues. 
%
An intuitive strategy to improve geometric accuracy of sparse input NeRF is by supervising the generated density values $\sigma$ for sampled points. However, obtaining ground truth for all these points is unfeasible.
An alternative strategy involves incorporating auxiliary supervisory information during training, such as depth for geometry~\cite{deng2022depth, wang2023sparsenerf} or semantic cues for appearance~\cite{jain2021putting}. However, these supervisory signals themselves might be inaccurate, which limits the potential effectiveness of this route. 
Our work also aims to enhance the generalization ability of NeRF, by training a model that can infer across different scenes with sparse source views. This allows us to flexibly handle situations where training data for certain scenes are limited, while maintaining photo-realistic rendering results.
Approaches like MVSNeRF~\cite{chen2021mvsnerf} and PixelNeRF~\cite{yu2021pixelnerf} have demonstrated improved generalization capabilities by pre-training their models on a diverse multi-view image dataset with various scenes. 
PixelNeRF integrates pixel features from source views to enhance network capabilities, but inaccuracies arise due to inconsistent correspondence between 2D pixels in source images and the queried 3D location.
%
Therefore, these methods inevitably lead to imprecise modeling, they fail to maintain consistency in geometry and appearance across various views, as they do not properly consider the correlation and cooperation among different viewpoints.

To address these challenges, we propose a generalizable sparse input neural radiance field (ColNeRF), a novel approach that gets rid of the need for additional supervision, constructing the precise and generalized model with the consideration of the collaboration among input source views.
Specifically, this method involves the extraction of feature volumes from source images using a pre-trained encoder, followed by the application of cross-view volume fusion to adaptively integrate these features.
The exact spatial locations of relevant patches can be determined by matching and reprojecting them into 3D space using camera parameters. 
This spatial transformation is achieved implicitly through an attention mechanism, serving as a trainable aggregation function that selectively emphasizes important features within source views~\cite{10.1145/3394171.3413839}. This mechanism also corrects features of occluded regions by incorporating information from corresponding parts in alternative viewpoints.
Furthermore, the collaboration also pays attention to the output end, where the constraint is enforced in both geometry and appearance reconstruction.  
For geometric regularization, we adopt a self-supervised approach~\cite{9204448} that aims to minimize discrepancies between predicted depths of adjacent target rays. 
For appearance regularization, we leverage the insight that the most relevant regions within the source views for each target ray should ideally align with their corresponding epipolar lines.
It is notable that we train a single model with potent generalization capabilities applicable to all scenes.
In summary, our main contributions can be summarized as follows:
\begin{itemize}
\item We propose ColNeRF to integrate multi-view compensation and consistency into NeRF at input/output ends, making ColNeRF outperform other generalizable NeRF methods with sparse input, and comparable to scene-specific NeRF approaches with reduced complexity.
\item We introduce self-supervised ray regularization to effectively enforce multi-view consistency for effective model guidance, which leads to more accurate geometry and appearance reconstruction. 
\item ColNeRF achieves superior performance over state-of-the-art generalizable NeRF methods in sparse scenarios and offers efficient adaptability to new scenes via fine-tuning, showcasing comparable results to scene-specific NeRF approaches with reduced computational burden.
\end{itemize}

\section{Related Works}
\subsubsection{Preliminary of NeRF.}
NeRF generated novel view images through an implicit 5D neural radiation field construction process denoted as ${\rm F}({\rm \gamma}({\rm x}),{\rm\gamma}({\rm d}))=(c,\sigma)$, where ${\rm \gamma}(\cdot)$ signifies the position encoding procedure~\cite{mildenhall2021nerf}, ${\rm x} = (x,y,z)$ denotes a 3D locationand and ${\rm d}=(\theta,\phi)$ denotes a 2D viewing direction. The output $c$ represents RGB values and $\sigma$ denotes volume density, which can be understood as the probability of a ray terminating at a given particle. 
The volumetric radiation field produces 2D images via pixel-wise rendering:
\begin{equation}
{\rm C}({\rm r})=\int_{t_n}^{t_f}T(t){\rm \sigma}({\rm r}(t)){\rm c}({\rm r}(t), {\rm d})dt,
\end{equation}
where $T(t)={\rm exp}(-\int_{t_n}^{t}{\rm \sigma}({\rm r}(s))ds)$, representing accumulated transmittance indicating ray traversal probability from $t_n$ to $t$ without encountering particles. Here, ${\rm c}({\rm r}(t), {\rm d})$ and ${\rm \sigma}({\rm r}(t))$ denote color and volume density at the sampled point along ray ${\rm r}$ at distance $t$. The radiance field optimization involves minimizing mean squared error between rendered and ground truth colors:
\begin{equation}
\mathcal{L}_{rec}=\sum\limits_{{\rm r}\in{\rm R}({\rm P})} \left\|{\rm \hat{C}}({\rm r})-{\rm C}({\rm r})\right\|_{2}^{2},
\end{equation}
where ${\rm R}({\rm P})$ is the set of all camera rays of target pose P.

\begin{figure*}[t]
  \centering
  \includegraphics[width=0.96\textwidth]{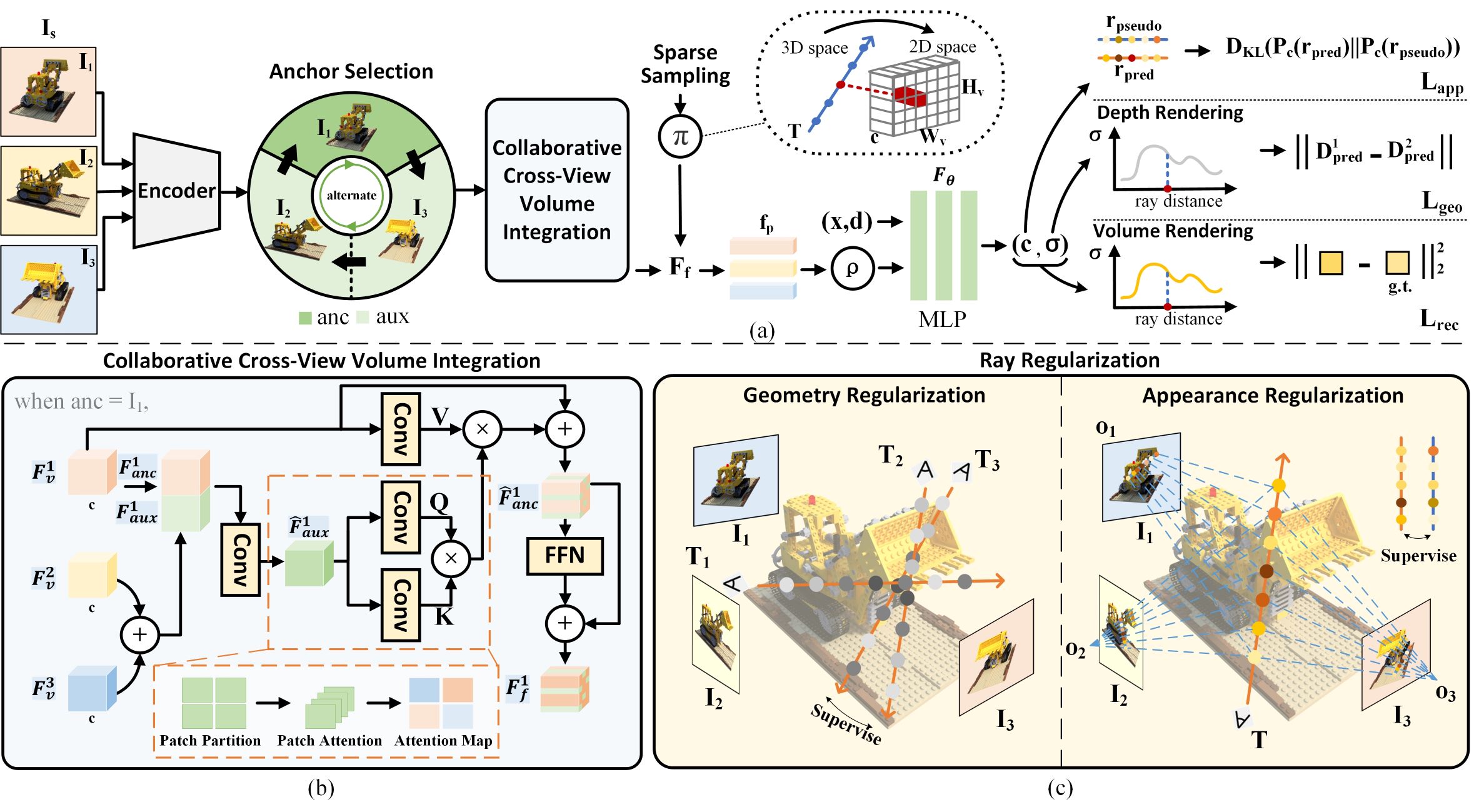}
  \caption{
  \textbf{
  The architecture of proposed ColNeRF.
  (a): The overview pipeline.
  (b) and (c): Collaborative Cross-View Volume Integration (CCVI) and Ray Regularization, \textit{i.e.} collaborative input fusion and output constraint.}
  ColNeRF consists of four key steps: 
  1) Feature volumes are extracted and processed with Anchor Selection (anc: anchor; aux: auxiliary) and Collaborative Cross-View Volume Integration (CCVI) to yield affined feature volumes $F_f$.
  2) Points are sampled and projected ($\pi$) onto $F_f$ to derive local features $f_p$. $f_p$ are subsequently averaged ($\rho$) and fed into MLP $F_{\theta}$, alongside $({\rm x}, {\rm d})$, for predicting radiance values $(c,\sigma)$.
  3) Ray Regularization is employed for predicted output alone for each target ray, encompassing both geometric and appearance aspects.
  4) Volume rendering produces final RGB values for pixels in novel views.
  }
  \label{figure2}
\end{figure*}

\subsubsection{Sparse Input NeRF.}
Researchers that pursue accurate reconstructed results with a reduced number of input views (\textit{i.e.} sparse input) have garnered significant attention~\cite{chen2023bidirectional,xu2022behind}. The challenge of sparse input 3D reconstruction arises from the complex task of maintaining consistency in both geometric shape and appearance. 
Historically, approaches predominantly relied on additional depth or semantics-based supervisory cues to infer occluded regions. 
A pioneering effort by PixelNeRF first integrated pixel features of source images into vanilla NeRF that only use position information.
Additionally, IBRNet~\cite{wang2021ibrnet}, SRF~\cite{chibane2021stereo} and MatchNeRF~\cite{chen2023explicit} contributed to scene reconstruction through the feature alignment of projected points from diverse perspectives.
Besides, researchers have also explored novel forms of explicit 3D representation established from sparse images~\cite{fang2023one}, such as voxel mesh~\cite{maturana2015voxnet, sun2022direct,huang20193d,deng2021voxel}, multiplane images (MPI)~\cite{li2021mine, fontaine2022photonic}, or layered depth images (LDI)~\cite{tulsiani2018layer, shih20203d}.
To address the challenge of inaccurate geometric information under sparse input settings, regularization methods targeting volume density have been introduced~\cite{somraj2023VipNeRF, somraj2023simplenerf}. 
For instance, Lombardi et al.~\cite{lombardi2019neural} enforced zero volume density for the near camera plane using masks, while InfoNeRF~\cite{kim2022infonerf} narrows the distribution of $\sigma$ within the front and back halves of the same ray, which is more suitable for cases where objects are located in the middle of a scene. RegNeRF~\cite{niemeyer2022regnerf} applies depth constraints on sampled image patches, which is evidently unsuitable for a cross-scene training strategy. An innovative approach taken by FreeNeRF~\cite{yang2023freenerf} involved the regularization of position encoding frequency for 5D inputs, yielding noteworthy outcomes.

\section{Collaborative Neural Radiance Fields}
\label{sec_method}

\subsection{Motivation}
Given a limited set of source images along with the corresponding camera extrinsics $\{({\rm I}_i\in{\rm \mathbb{R}}^{H\times W \times 3},{\rm P}_i\in{\rm \mathbb{R}}^{3 \times 4} )\}$, we aim to address the following two issues:
\begin{itemize}
    \item \textbf{Limited Effectiveness.} When the input is sparse, incorporating auxiliary supervision such as depth or semantic cues can improve NeRF's performance to an extent.
    However, these guidance might be not reliable and difficult to obtain, which reduces the effectiveness.
    \item \textbf{Limited Generalization.} As most methods take the one-scene-one-model paradigm.
    Although pre-training on diverse scenes can improves models' generalization ability, previous works have not fully consider the collaborative relationship of different views systematically, which hinders performance improvement.
\end{itemize}
Our core goal is to develop a collaborative NeRF model with the capacity for \textbf{cross-scene generalization} and rendering \textbf{high-quality} results with \textbf{multi-view consistency} when taking \textbf{sparse input}, without utilizing any auxiliary supervision.
In the following, we introduce the overall framework of our approach and then detailed our two contributions.

\begin{table*}[t]
\small
\tabcolsep 0.03cm
\begin{tabular}{l|c|ccc|ccc|ccc|ccc}
    \toprule
      \multirow{2}{*}{Method}  & 
      \multirow{2}{*}{Setting} &
      \multicolumn{3}{c|}{PSNR ↑} &
      \multicolumn{3}{c|}{SSIM ↑} &
      \multicolumn{3}{c|}{LPIPS ↓} &
      \multicolumn{3}{c}{Average ↓} \\
     & & 3-view & 6-view & 9-view & 3-view & 6-view & 9-view & 3-view & 6-view & 9-view & 3-view & 6-view & 9-view  \\
      
    \midrule
    DietNeRF (ICCV 2021)& \multirow{5}{*}{\makecell{Trained on DTU\\and \\Optimized per Scene}} & 10.01 & 18.70 & 22.16 & 0.354 & 0.668 & 0.740 & 0.574 & 0.336 & 0.277 & 0.383 & 0.149 & 0.098  \\
    DS-NeRF (CVPR 2022) & & 16.50 & 20.50 & - & 0.540 & 0.730 & - & 0.480 & 0.310 & - & 0.194 & 0.113 & - \\
    InfoNeRF (CVPR 2022) & & 11.23 & - & - & 0.445 & - & - & 0.543 &- & - & 0.312 & - & - \\
    RegNeRF (CVPR 2022) & & 15.33 & 19.10 & 22.30 & 0.621 & 0.757 & 0.823 & \underline{0.341} &  \textbf{0.233} &  \textbf{0.184} & 0.189 & 0.118 & \underline{0.079} \\
    FreeNeRF (CVPR 2023) & & 18.02 & \underline{22.39} &  \textbf{24.20} & \underline{0.680} & \underline{0.779} &  \textbf{0.833} &  \textbf{0.318}  &  \underline{0.240} &  \underline{0.187} & 0.146  &  \underline{0.094}  &  \textbf{0.068}  \\ \midrule
    SRF (CVPR 2021) & \multirow{4}{*}{\makecell{Trained on DTU\\and \\Not Optimized per Scene}} & 15.84 & 17.77 & 18.56 & 0.532 & 0.616 & 0.652 & 0.482 & 0.401 & 0.359 & 0.207 & 0.162 & 0.145 \\
    MVSNeRF (ICCV 2021)  & & 16.33 & 18.26 & 20.32 & 0.602 & 0.695 & 0.735 & 0.385 & 0.321 & 0.280 & 0.184 & 0.146 & 0.114 \\
    PixelNeRF (CVPR 2021) & &\underline{18.74}&21.02&22.23 &0.618&0.684&0.714&0.401&0.340&0.323&\underline{0.142}&0.119&0.105\\
    \textbf{ColNeRF (Ours)} & & \textbf{19.55} & \textbf{22.94} & \underline{23.93} &  \textbf{0.716} &  \textbf{0.797} & \underline{0.824}  & 0.362 & 0.317 & 0.298 &  \textbf{0.129} &  \textbf{0.090} & \underline{0.079}\\
    \bottomrule
\end{tabular}
\caption{
\textbf{Quantitative comparison on DTU.} 
Our model demonstrates superior performance in sparse input synthesizing compared to most existing methods. Our direct baseline is PixelNeRF. For ease of identification, the entries with the best and second-best performances are respectively highlighted in bold and underscored with an underline.
}
\label{Tab1}
\end{table*}

\begin{table*}[t]
\small
\tabcolsep 0.02cm
\begin{tabular}{l|c|ccc|ccc|ccc|ccc}
    \toprule
      \multirow{2}{*}{Method}  & 
      \multirow{2}{*}{Setting} & 
      \multicolumn{3}{c|}{PSNR ↑} &
      \multicolumn{3}{c|}{SSIM ↑} &
      \multicolumn{3}{c|}{LPIPS ↓} &
      \multicolumn{3}{c}{Average ↓} \\
     & &3-view & 6-view & 9-view & 3-view & 6-view & 9-view & 3-view & 6-view & 9-view & 3-view & 6-view & 9-view  \\
      
    \midrule
    DietNeRF (ICCV 2021)& \multirow{3}{*}{\makecell{Trained on LLFF\\and \\Optimized per Scene}} & 14.94 & 21.75 &24.28 &0.370 &0.717&0.801&0.496 &0.248 &0.183&0.240 &0.105 &0.073  \\
    RegNeRF (CVPR 2022)& & 19.08 &23.10 &\underline{24.86} &\underline{0.587} &\underline{0.760}&\underline{0.820} &0.336&\underline{0.206}&\underline{0.161}&0.149 &\underline{0.086}&\underline{0.067} \\
    FreeNeRF (CVPR 2023)& & \underline{19.63} &\textbf{23.73} &\textbf{25.13} & \textbf{0.612} & \textbf{0.779}& \textbf{0.827}& \textbf{0.308}& \textbf{0.195}& \textbf{0.160} & \underline{0.134}& \textbf{0.075} & \textbf{0.064} \\ \midrule
    SRF ft (CVPR 2021) & \multirow{4}{*}{\makecell{Trained on DTU\\and\\Not Optimized per Scene}} & 17.07 & 16.75 & 17.39 & 0.436 & 0.438 & 0.465 & 0.529 & 0.521 & 0.503 & 0.203 & 0.207 & 0.193 \\
    MVSNeRF ft (ICCV 2021) & &17.88 & 19.99 & 20.47 & 0.584 & 0.660 & 0.695 & \underline{0.327} & 0.264 & 0.244 & 0.157 & 0.122 & 0.111 \\
    PixelNeRF ft (CVPR 2021) & &16.17 & 17.03 & 18.92 & 0.438 & 0.473 & 0.535 & 0.512 & 0.477 & 0.430 & 0.217 & 0.196 & 0.163 \\
    \textbf{ColNeRF ft (Ours)} & & \textbf{20.97} & \underline{23.32} & 23.52 & \underline{0.587} & 0.747 & 0.762 & 0.447 & 0.295 & 0.280 & \textbf{0.132} & 0.088 & 0.084\\
    \bottomrule
\end{tabular}
\caption{\textbf{Quantitative comparison on LLFF}. We generalize the pre-trained model to LLFF dataset and conduct 15K, 10K, and 5K fine-tuning iterations for each scene with 3, 6, and 9 views (all fewer than Pixel-NeRF's default 20K fine-tune steps). Although methods like FreeNeRF may produce better results, they train separate models on each scene for 250K iterations. In contrast, our method trains a single model for all scenes and achieves comparable results with much less fine-tuning cost.}
\label{Tab2}
\end{table*}

\begin{figure*}[t] 
    \centering
    \small
    \begin{minipage}{\textwidth}%
        \makebox[0.16\linewidth]{PixelNeRF}
        \makebox[0.16\linewidth]{\textbf{Ours}}
        \makebox[0.16\linewidth]{Ground Truth}
        \hspace{1.5mm}
        \makebox[0.16\linewidth]{PixelNeRF}
        \makebox[0.16\linewidth]{\textbf{Ours}}
        \makebox[0.16\linewidth]{Ground Truth}
        \\
        \begin{tikzpicture}
            [spy using outlines={rectangle, magnification=2, size=1cm}]
            \node[inner sep=0pt, outer sep=0pt] at (0,0){\includegraphics[height=0.12\linewidth]{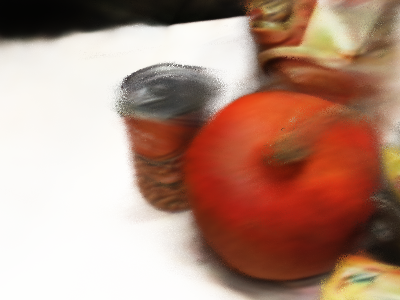}};
            \spy [red] on (0.65,0.1) in node at (-0.9, -0.55);
        \end{tikzpicture}
        \begin{tikzpicture}
            [spy using outlines={rectangle, magnification=2, size=1cm}]
            \node[inner sep=0pt, outer sep=0pt] at (0,0){\includegraphics[height=0.12\linewidth]{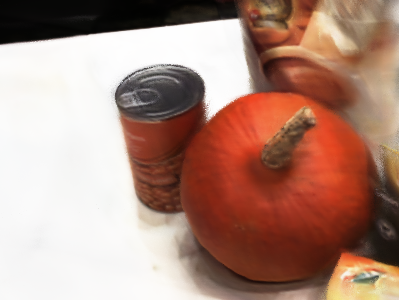}};
            \spy [red] on (0.65,0.1) in node at (-0.9, -0.55);
        \end{tikzpicture}
        \begin{tikzpicture}
            [spy using outlines={rectangle, magnification=2, size=1cm}]
            \node[inner sep=0pt, outer sep=0pt] at (0,0){\includegraphics[height=0.12\linewidth]{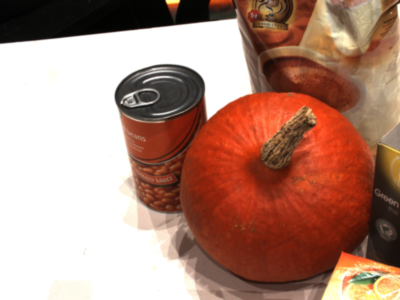}};
            \spy [red] on (0.65,0.1) in node at (-0.9, -0.55);
        \end{tikzpicture}
        \hspace{1.5mm}
        \begin{tikzpicture}
            [spy using outlines={rectangle, magnification=2, size=1cm}]
            \node[inner sep=0pt, outer sep=0pt] at (0,0){\includegraphics[height=0.12\linewidth]{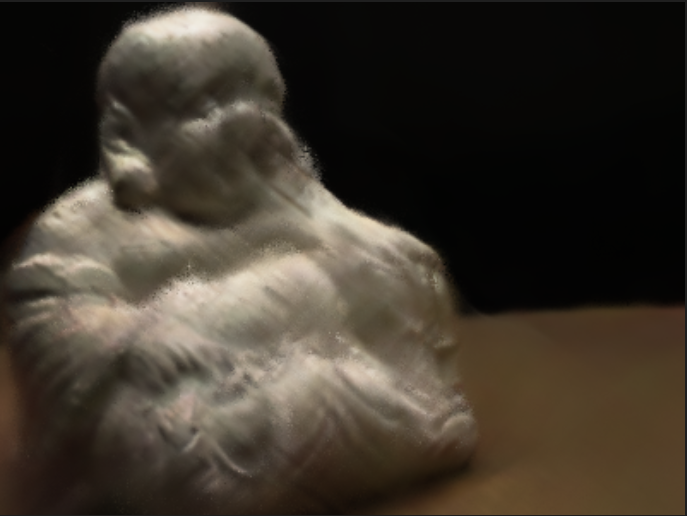}};
            \spy [red] on (-0.38,0.4) in node at (0.9, 0.55);
        \end{tikzpicture}
        \begin{tikzpicture}
            [spy using outlines={rectangle, magnification=2, size=1cm}]
            \node[inner sep=0pt, outer sep=0pt] at (0,0){\includegraphics[height=0.12\linewidth]{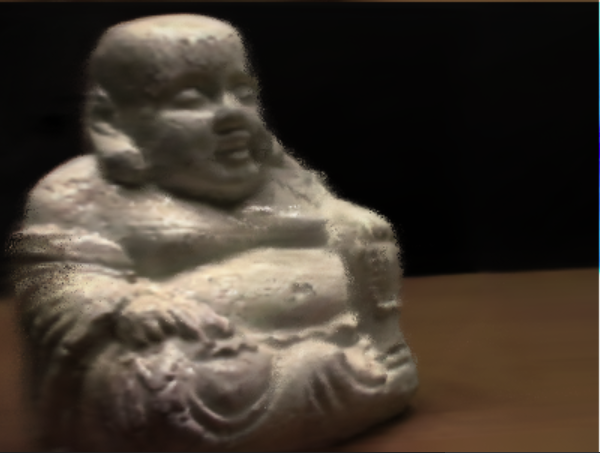}};
            \spy [red] on (-0.38,0.4) in node at (0.9, 0.55);
        \end{tikzpicture}
        \begin{tikzpicture}
            [spy using outlines={rectangle, magnification=2, size=1cm}]
            \node[inner sep=0pt, outer sep=0pt] at (0,0){\includegraphics[height=0.12\linewidth]{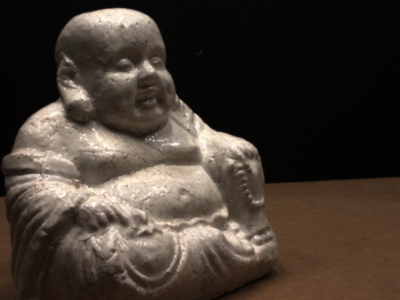}};
            \spy [red] on (-0.38,0.4) in node at (0.9, 0.55);
        \end{tikzpicture}
        \\
        \makebox[\linewidth]{(a) 3 Input View}
        \\
        \begin{tikzpicture}
            [spy using outlines={rectangle, magnification=4, size=1cm}]
            \node[inner sep=0pt, outer sep=0pt] at (0,0){\includegraphics[height=0.12\linewidth]{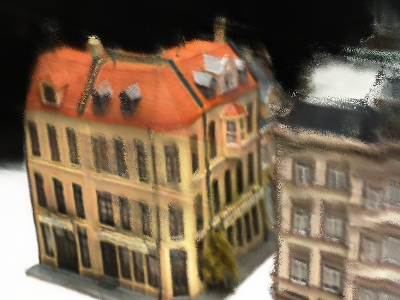}};
            \spy [red] on (1.28,0.28) in node at (-0.9, -0.55);
        \end{tikzpicture}
        \begin{tikzpicture}
            [spy using outlines={rectangle, magnification=4, size=1cm}]
            \node[inner sep=0pt, outer sep=0pt] at (0,0){\includegraphics[height=0.12\linewidth]{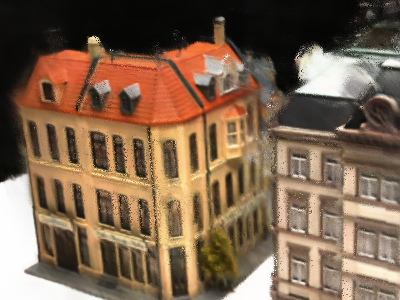}};
            \spy [red] on (1.28,0.28) in node at (-0.9, -0.55);
        \end{tikzpicture}
        \begin{tikzpicture}
            [spy using outlines={rectangle, magnification=4, size=1cm}]
            \node[inner sep=0pt, outer sep=0pt] at (0,0){\includegraphics[height=0.12\linewidth]{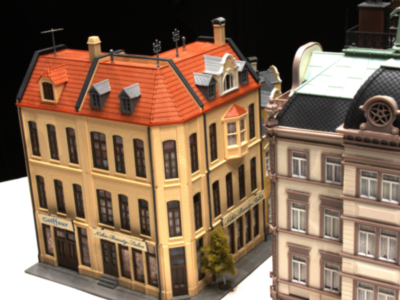}};
            \spy [red] on (1.28,0.28) in node at (-0.9, -0.55);
        \end{tikzpicture}
        \hspace{1.5mm}
        \begin{tikzpicture}
            [spy using outlines={rectangle, magnification=2.5, size=1cm}]
            \node[inner sep=0pt, outer sep=0pt] at (0,0){\includegraphics[height=0.12\linewidth]{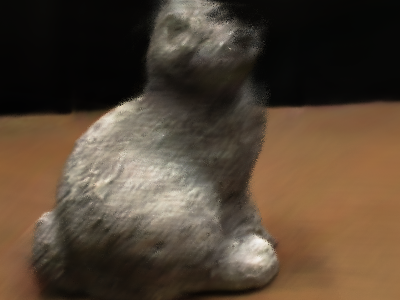}};
            \spy [red] on (0.25,0.85) in node at (0.9, -0.55);
        \end{tikzpicture}
        \begin{tikzpicture}
            [spy using outlines={rectangle, magnification=2.5, size=1cm}]
            \node[inner sep=0pt, outer sep=0pt] at (0,0){\includegraphics[height=0.12\linewidth]{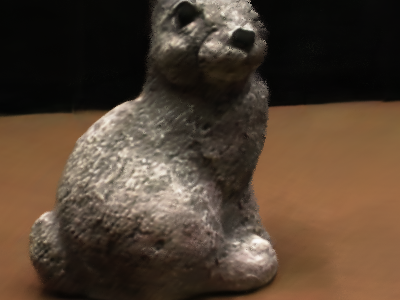}};
            \spy [red] on (0.25,0.85) in node at (0.9, -0.55);
        \end{tikzpicture}
        \begin{tikzpicture}
            [spy using outlines={rectangle, magnification=2.5, size=1cm}]
            \node[inner sep=0pt, outer sep=0pt] at (0,0){\includegraphics[height=0.12\linewidth]{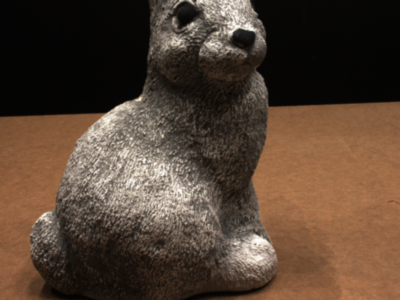}};
            \spy [red] on (0.25,0.85) in node at (0.9, -0.55);
        \end{tikzpicture}
        \\
        \makebox[\linewidth]{(b) 6 Input View}

    \end{minipage}

    \caption{
    \textbf{Qualitative comparison on DTU between PixelNeRF and ColNeRF}, we present results under 3 and 6 input views setting. PixelNeRF's direct use of the average pixel feature from each source view often results in blurriness or shape distortion.
    }
    \label{DTU}
\end{figure*}

\subsection{Overview}
Our system pipeline is depicted in Fig.~\ref{figure2}. At first, we use a pre-trained encoder \textbf{ResNet34}~\cite{he2016deep} to extract feature volumes ${\rm F}_v \in {\rm \mathbb{R}}^{H_v \times\ W_v \times d}$ from the source views. $H_v$, $W_v$, and $d$ respectively represent the height, width, and the channel dimension. Subsequently, we employ a collaborative cross-view attention mechanism to integrate these feature volumes and obtain fused results ${\rm F}_{f}$ of $N$ source views.

The next step involves sampling $N_r$ target rays for training. To train a generalized model applicable across a variety of scenes, we adopt a strategy that randomly selects scene and emits rays into it with a scattered pattern.
Subsequently, we sample $N_p$ points on each target ray. For these sampled points, their camera parameters enable us to project them onto each source image. We then extract their corresponding pixel features ${\rm f}_p$ from the affined feature volume ${\rm F}_{f}$ of each view using bilinear interpolation, the local feature of 3D point $\rm x$ in the $i$-th source view is obtained as follows:
\begin{equation}
{\rm f}_{p}^{i} = {\rm Interpolate} ( {\rm F}_{f}^i({\rm \Pi}({\rm x}) ) \in {\rm \mathbb{R}}^{d}.
\end{equation}

The local pixel features ${\rm f}_p$ are then input into the Neural Radiance Field ${\rm F}_{\theta}$ along with the coordinates ${\rm x}$ and view direction ${\rm d}$ to yield the color $c$ and density $\sigma$:
\begin{equation}
    {\rm F}({\rm \gamma}({\rm x}),{\rm \gamma}({\rm d}),{\rm \rho}(\{{\rm f}_{p}^{i}\}_{i=1}^N))=(c,\sigma),
\end{equation}
where ${\rm \rho}$ denotes the averaging operation, and $N$ denotes the number of source views. 

Finally, as illustrated in Eqn.~(1), we employ principles from classical volume rendering to aggregate the final RGB values ${\rm C}({\rm r})$. The training loss function of our model includes three parts. One is the reconstruction loss, which is identical to that in Eqn.~(2). The remaining two components originate from the ray regularization module:
\begin{equation}  \mathcal{L}_{total}=\mathcal{L}_{rec}+\lambda_1\mathcal{L}_{geo}+\lambda_2\mathcal{L}_{app}.
\end{equation}
The loss weights $\lambda_1$ and $\lambda_2$ are set to 1$e$-4 and 2$e$-4 respectively throughout our experiments.

\subsection{Collaborative Cross-View Volume Integration}
\label{Global Feature Fusion}
Before being fed into the MLP, we enrich sparse information by fusing multi-view source images. This process helps identify corresponding regions cross different views. It then utilizes information from these perspectives to correct potential biases in the source image.

Additionally, this process ensures geometric consistency across multiple views of the same scene. Unlike previous approaches that focus on pixel-level features from different perspectives, our strategy integrates at the patch level. This enables us to generate $N$ affined feature volumes that capture information from other source views. Each of these volumes can be denoted as ${\rm F}_{f}^{i}\in{\rm \mathbb{R}}^{H_v \times\ W_v \times d}$:
\begin{equation}
    {\rm F}_{f}^{i} = {\rm CCVI}({\rm F}_{anc}^{i}, {\rm F}_{aux}^{i}),
\end{equation}
${\rm F}_{anc}^{i}$ represents current anchor feature volume, each source view takes turns as the anchor: ${\rm F}_{anc}^{i}$ = ${\rm F}_{v}^{i}$, while ${\rm F}_{aux}^{i}$ represents the summation of other auxiliary feature volumes:
\begin{equation}
    {\rm F}_{aux}^{i} = \sum_{j=1}^{i-1}{\rm F}_{v}^{j} + \sum_{j=i+1}^{N}{\rm F}_{v}^{j}.
\end{equation}
The transformer block in CCVI is computed as:
\begin{equation}
    \begin{aligned}
    \hat{{\rm F}}_{anc}^{i} &={\rm AVGI}({\rm F}_{anc}^{i},{\rm \hat{F}}_{aux}^{i})+{\rm F}_{anc}^{i}, \\
    {\rm F}_{f}^{i} &= {\rm FFN}(\hat{{\rm F}}_{anc}^{i})+\hat{{\rm F}}_{anc}^{i}, \\
    {\rm \hat{F}}_{aux}^{i} &= {\rm Conv}([{\rm F}_{anc}^{i},{\rm F}_{aux}^{i}]),
    \end{aligned} \\
\end{equation}
${\rm AVGI}(\cdot)$ denotes Auxiliary Volume Guided Integration, ${\rm FFN}(\cdot)$ denotes a Feed-Forward Network, ${\rm Conv}(\cdot)$ denotes a convolutional layer for dimension reduction.
${\rm \hat{F}}_{aux}^{i}$ are used as auxiliary volume for current anchor feature volume ${\rm F}_{anc}^i$:
\begin{equation}
{\rm AVGI}({\rm F}_{anc},{\rm \hat{F}}_{aux})={\rm Softmax}\left(\frac{{\rm Q}{\rm K}^T}{\sqrt{{d}_{k}}}\right){\rm V},
\end{equation}
where 
\begin{equation*}
\begin{aligned}
&{\rm Q}={\rm W}_{Q} ({\rm \hat{F}}_{aux}), \\&{\rm K}={\rm W}_{K} ({\rm \hat{F}}_{aux}),\\
&{\rm V}={\rm W}_{V} ({\rm F}_{anc}),
\end{aligned}
\end{equation*}
${\rm W}_{Q}$,${\rm W}_{K}$ and ${\rm W}_{V}$ are learnable transformations, ${\rm d}_{k}$ is the feature channel dimension of ${\rm Q}$ and ${\rm K}$.
Each view is processed individually to obtain fused feature volumes$\{{\rm F}_{f}^{i}\}_{i=1}^N$.

\begin{figure*}[t] 
    \centering
    \small
    
    \begin{minipage}{\textwidth}%
        \makebox[0.195\linewidth]{RegNeRF}
        \makebox[0.195\linewidth]{FreeNeRF}
        \makebox[0.195\linewidth]{PixelNeRF ft}
        \makebox[0.195\linewidth]{\textbf{Ours}}
        \makebox[0.195\linewidth]{Ground Truth}
        \\
        \begin{tikzpicture}
            [spy using outlines={rectangle, magnification=3.5, size=1cm}]
            \node[inner sep=0pt, outer sep=0pt] at (0,0){\includegraphics[height=0.146\linewidth]{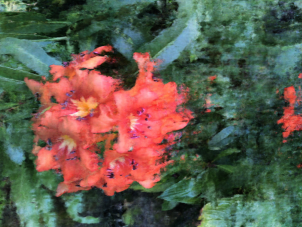}};
            \spy [red] on (0.25,0.05) in node at (1.22, -0.78);
        \end{tikzpicture}
        \begin{tikzpicture}
            [spy using outlines={rectangle, magnification=3.5, size=1cm}]
            \node[inner sep=0pt, outer sep=0pt] at (0,0){\includegraphics[height=0.146\linewidth]{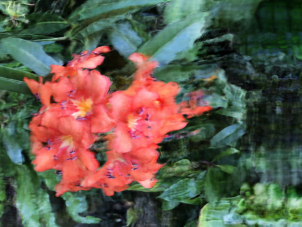}};
            \spy [red] on (0.25,0.05) in node at (1.22, -0.78);
        \end{tikzpicture}
        \begin{tikzpicture}
            [spy using outlines={rectangle, magnification=3.5, size=1cm}]
            \node[inner sep=0pt, outer sep=0pt] at (0,0){\includegraphics[height=0.146\linewidth]{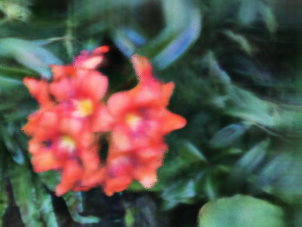}};
            \spy [red] on (0.25,0.05) in node at (1.22, -0.78);
        \end{tikzpicture}
        \begin{tikzpicture}
            [spy using outlines={rectangle, magnification=3.5, size=1cm}]
            \node[inner sep=0pt, outer sep=0pt] at (0,0){\includegraphics[height=0.146\linewidth]{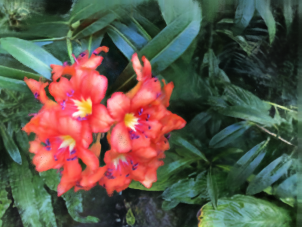}};
            \spy [red] on (0.25,0.05) in node at (1.22, -0.78);
        \end{tikzpicture}
        \begin{tikzpicture}
            [spy using outlines={rectangle, magnification=3.5, size=1cm}]
            \node[inner sep=0pt, outer sep=0pt] at (0,0){\includegraphics[height=0.146\linewidth]{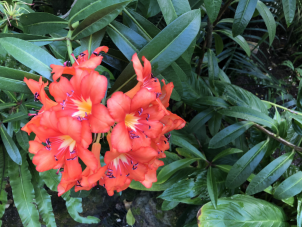}};
            \spy [red] on (0.25,0.05) in node at (1.22, -0.78);
        \end{tikzpicture}
        \\
        \begin{tikzpicture}
            [spy using outlines={rectangle, magnification=3.5, size=1cm}]
            \node[inner sep=0pt, outer sep=0pt] at (0,0){\includegraphics[height=0.146\linewidth]{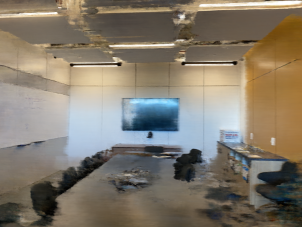}};
            \spy [red] on (1.05,-1.1) in node at (1.22, 0.78);
        \end{tikzpicture}
        \begin{tikzpicture}
            [spy using outlines={rectangle, magnification=3.5, size=1cm}]
            \node[inner sep=0pt, outer sep=0pt] at (0,0){\includegraphics[height=0.146\linewidth]{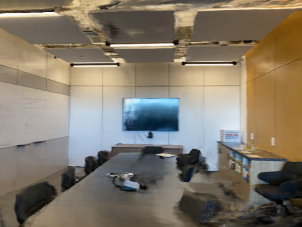}};
           \spy [red] on (1.05,-1.1) in node at (1.22, 0.78);
        \end{tikzpicture}
        \begin{tikzpicture}
            [spy using outlines={rectangle, magnification=3.5, size=1cm}]
            \node[inner sep=0pt, outer sep=0pt] at (0,0){\includegraphics[height=0.146\linewidth]{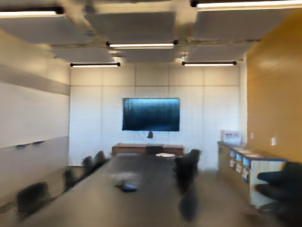}};
           \spy [red] on (1.05,-1.1) in node at (1.22, 0.78);
        \end{tikzpicture}
        \begin{tikzpicture}
            [spy using outlines={rectangle, magnification=3.5, size=1cm}]
            \node[inner sep=0pt, outer sep=0pt] at (0,0){\includegraphics[height=0.146\linewidth]{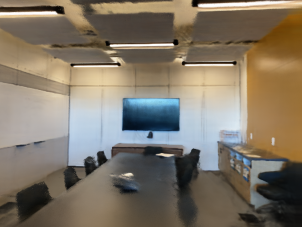}};
            \spy [red] on (1.05,-1.1) in node at (1.22, 0.78);
        \end{tikzpicture}
        \begin{tikzpicture}
            [spy using outlines={rectangle, magnification=3.5, size=1cm}]
            \node[inner sep=0pt, outer sep=0pt] at (0,0){\includegraphics[height=0.146\linewidth]{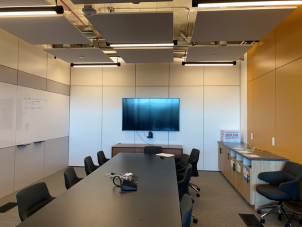}};
            \spy [red] on (1.05,-1.1) in node at (1.22, 0.78);
        \end{tikzpicture}

    \end{minipage}

    \caption{
    \textbf{Qualitative results on LLFF under 3 input views setting.} FreeNeRF is among the best-performing methods for per-scene optimization, but it exhibits noticeable noise issues due to the inaccurate encoding of high-frequency information. PixelNeRF's results suffer from apparent blurriness when compared to our method.
    }
    \label{LLFF}
\end{figure*}

\begin{figure}[t] 
    \centering
    \small
    
    \begin{minipage}{\textwidth}%
        \makebox[0.153\linewidth]{NeRF}
        \makebox[0.153\linewidth]{DS-NeRF}
        \makebox[0.153\linewidth]{\textbf{Ours}}
        \\
        \includegraphics[height=0.113\textwidth]{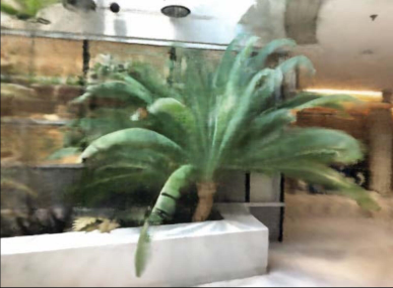}
        \includegraphics[height=0.113\textwidth]{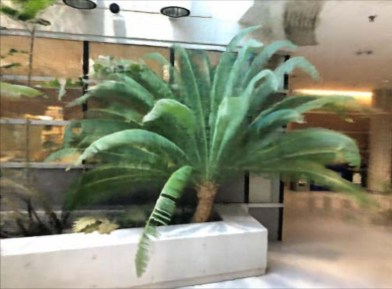}
        \includegraphics[height=0.113\textwidth]{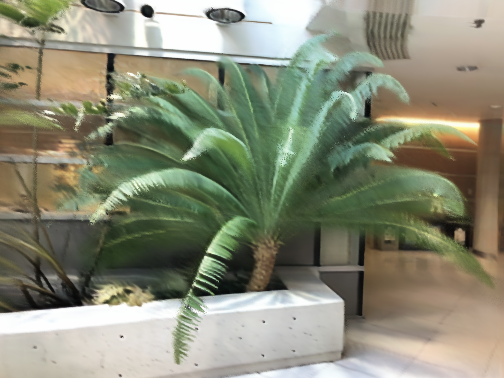}
        \\
        \begin{tikzpicture}
            [spy using outlines={rectangle, magnification=4, size=0.8cm}]
            \node[inner sep=0pt, outer sep=0pt] at (0,0){\includegraphics[height=0.1134\textwidth]{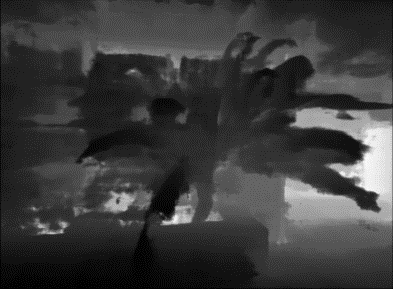}};
            \spy [red] on (1.03,0.26) in node at (-0.94, -0.6);
        \end{tikzpicture}
        \begin{tikzpicture}
            [spy using outlines={rectangle, magnification=4, size=0.8cm}]
            \node[inner sep=0pt, outer sep=0pt] at (0,0){\includegraphics[height=0.1134\textwidth]{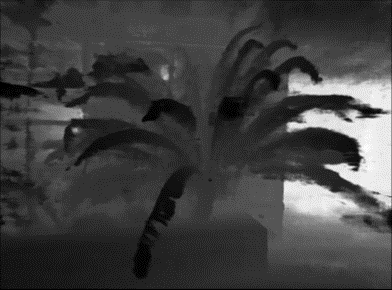}};
            \spy [red] on (1.03,0.26) in node at (-0.94, -0.6);
        \end{tikzpicture}
        \begin{tikzpicture}
            [spy using outlines={rectangle, magnification=4, size=0.8cm}]
            \node[inner sep=0pt, outer sep=0pt] at (0,0){\includegraphics[height=0.113\textwidth]{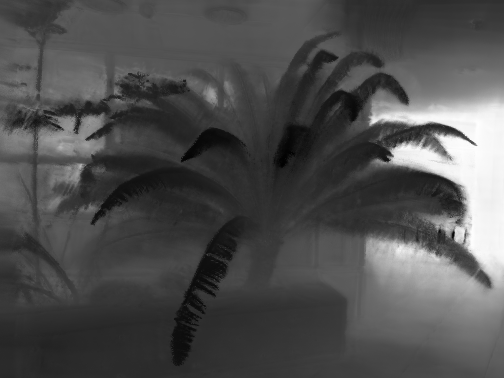}};
            \spy [red] on (1.03,0.26) in node at (-0.93, -0.6);
        \end{tikzpicture}

    \end{minipage}

    \caption{
    \textbf{Qualitative results on fern under 5 input views between NeRF, DS-NeRF and ColNeRF (ours).} Rendered depth maps reveal that our model achieves more accurate shapes than DS-NeRF, which relies on explicit depth labels.
    }
    \label{depth}
\end{figure}

\subsection{Ray Regularization}
\label{Ray Regularization}
\subsubsection{Geometry Regularization.} Our goal is to constrain the prediction of each points' density and improve the model's resilience to variations in view direction. Unlike InfoNeRF and RegNeRF, we opted for a more versatile ray regularization approach that better suits cross-scene training strategies. We employ a collaborative mutual-supervision for neighboring rays, pairing $N_{pairs}$ of the closest rays together and minimize the L1 Loss of the predicted depth for each pairs:
\begin{equation}
    \begin{aligned}
        \mathcal{L}_{geo} &=\sum_{i=1}^{N_{pairs}}{\rm M}({\rm r}_i) \odot (D({\rm r}_i)-D({\rm \hat{r}}_i)), \\
        {D}({\rm r}_i) &=\int_{t_n}^{t_f}T(t){\rm \sigma}({\rm r}_{i}(t))tdt, \\
         {\rm M}({\rm r}_i) & = \left\{
        \begin{array}{ll}
        0 \quad \text{if} \: {\rm Q}({\rm r}_i) < \tau \: \text{or} \: {\rm Q}({\rm \hat{r}}_i) < \tau\\
        1 \quad \rm otherwise
        \end{array}
        \right.,\\
    \end{aligned}
\end{equation}
${\rm r}_i$ and ${\rm \hat{r}}_i$ refer to the two neighboring target rays that are paired together. ${D}(\cdot)$ denotes the predicted depth of sampled rays. We employ a mask to exclude certain adjacent ray pairs that not need to be regularized. For instance, a pair that one ray hits the edge of an object while its corresponding ray does not hit anything. Applying geometric constraints to such pairs may introduce foggy artifacts. ${\rm Q}(\cdot)=\sum_{i=1}^N{1-\exp(-\sigma_i\delta_i)}$, represents the cumulative ray density, here $i$ refers to the $i$-th sampled point on a ray. We set $\tau=0.1$ in our experiments.


\subsubsection{Appearance Regularization.}
Epipolar plane is a plane determined in space by a spatial point and the optical centers of two distinct cameras, while the epipolar line arises from the intersection of the epipolar plane and the imaging plane. The RGB labels for each target ray are obtained from the color information along its corresponding epipolar lines on source views. Similar to the extraction of local features, epipolar lines are grabbed using projection and interpolation, their RGB are set as color label for target rays. However, due to occlusion, directly using $L_1$ loss for constraints can be overly restrictive. To address this, we minimize the KL-divergence between the color distributions $P_c$ of target rays and their corresponding color labels:
\begin{equation}
\begin{aligned}
\mathcal{L}_{app} &=\sum_{i=1}^{N_r}D_{KL}\left(P_c({\rm r}_{i})||P_c({\rm \hat{r}}_{i})\right)\\
&=\sum_{i=1}^{N_r}\sum_{j=1}^{N_p}{\rm p}_c({\rm M'}({\rm c}({\rm r}_{i,j})))\rm log\frac{p_c({\rm M'}(c({\rm r}_{i,j})))}{p_c({\rm M'}(c(\hat{{\rm r}}_{i,j})))}.
\end{aligned}
\end{equation}
Here, ${\rm \hat{r}_i}$ represents the target ray emitted during training, while ${\rm r}_i$ signifies the "pseudo-label" ray derived from averaging the color of target ray's corresponding epipolar lines across multiple source views. ${\rm p}_c({\rm \hat{r}}_{i, j})$ denotes the probability of j-th point on i-th target ray. A mask is employed to exclude inaccurately projected points from the computation:
\begin{equation}
 {\rm M'}(\cdot) = \left\{
    \begin{array}{ll}
    \rm false \quad &{\rm if} \: {\varepsilon}({\rm r}_{i,j}) >  {\varepsilon}_{\rm max} \\
    \rm true \quad &\rm otherwise
    \end{array}
    \right.,
\end{equation}
 ${\varepsilon}({\rm r}_{i,j})$ represents the pixel coordinate in the source image obtained by homography transformation from the point ${\rm r}_{i,j}$ on the target ray. The term ${\rm {\varepsilon}}_{max}$ refers to the maximum pixel coordinate in the source image.


\section{Experiments}
\subsection{Experimental Setups}
\subsubsection{Datasets.} We evaluate our method on two datasets: DTU~\cite{jensen2014large} and LLFF~\cite{mildenhall2019local}. We train on DTU and test the generalization capabilities of the pre-trained model on LLFF. For DTU, we follow the evaluation protocol established by PixelNeRF. For LLFF, we follow the evaluation standards set by NeRF and use it as an out-of-distribution test for conditional models. To evaluate our method's performance with sparse input, we conduct experiments with 3-view, 6-view, and 9-view configurations. 

\subsubsection{Metrics.} We report the mean of ${\rm PSNR}$, ${\rm SSIM}$~\cite{wang2004image}, and the ${\rm LPIPS}$ perceptual metric~\cite{zhang2018unreasonable}. To ease comparison, we also report the geometric mean of ${\rm MSE} = 10^{-{\rm PSNR}/10}$, $\sqrt{1 - {\rm SSIM}}$, and ${\rm LPIPS}$.

\subsubsection{Training Details.}
In line with PixelNeRF, we sample 128 training rays per iteration. To boost controllability, we randomly emit 112 rays and designate the final 16 of them as reference rays. The remaining 16 rays of all 128 rays are sampled from regions adjacent to reference rays. These freshly sampled rays share the same camera parameters and origin with the reference rays, but exhibit an offset of up to 7 pixels on the pixel plane. These last 32 rays are used as paired adjacent rays for geometry ray regularization. For the training of 3-view and 6-view, we set the batch size (BS) to 3, and for the 9-view training, BS is set to 2. We maintain a fixed learning rate of 1e-4 throughout our training process.

\subsubsection{Comparing Baselines.}
To facilitate comparison, we select several state-of-the-art (SOTA) methods that effectively address the challenge of limited input. These include PixelNeRF, SRF, MVSNeRF, DietNeRF, DS-NeRF, InfoNeRF, RegNeRF, and FreeNeRF. The first three, akin to our approach, are pre-trained across various scenes, while the remaining five are optimized for specific scenarios. Given the similarities in the compared methods, datasets, and experiment settings, we directly use the reported results in FreeNeRF and RegNeRF as the basis for our comparison with other methods. The results of DS-NeRF and InfoNeRF were taken from their published papers.

\begin{figure}[t]
  \centering
  \includegraphics[width=0.98\linewidth]{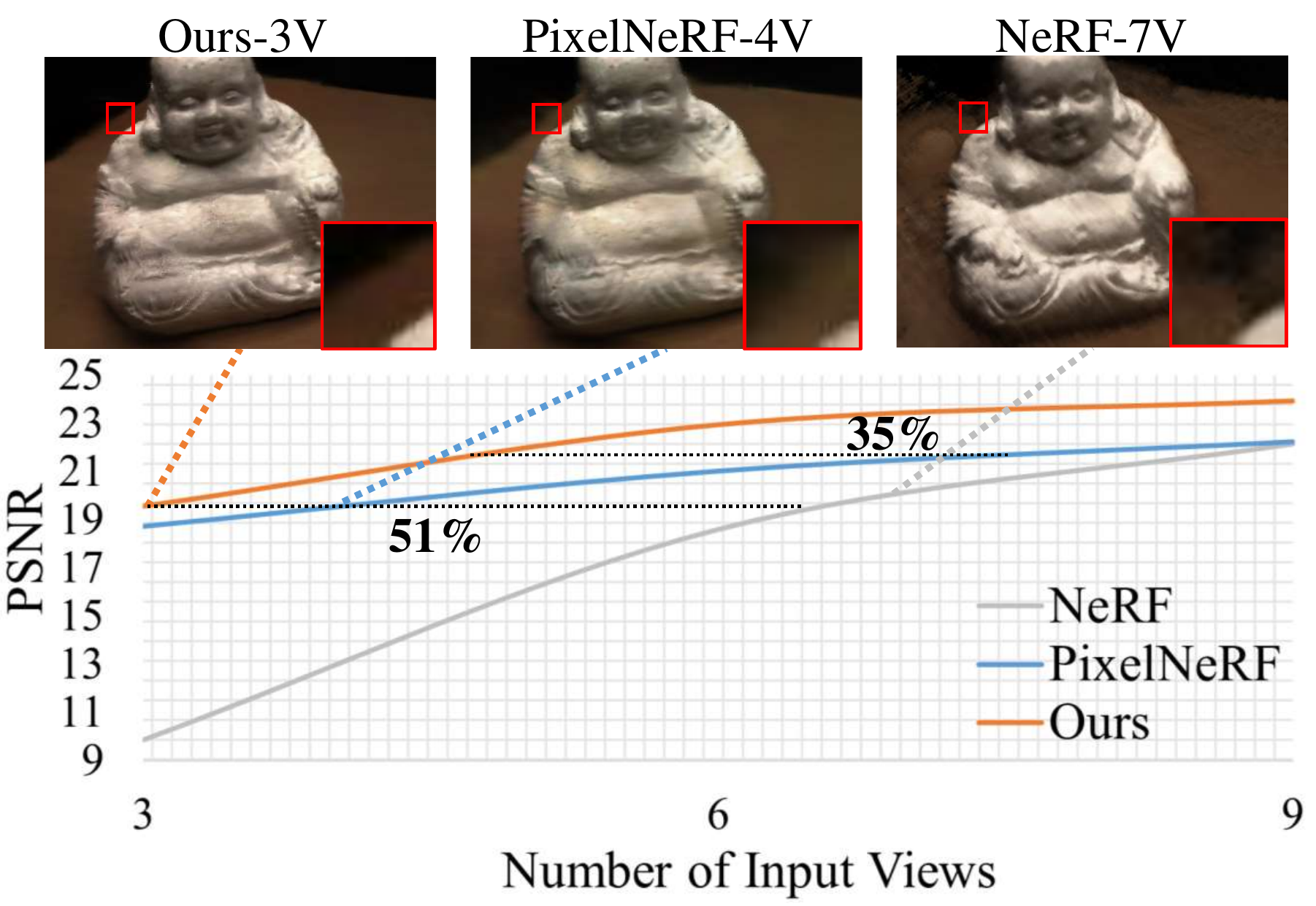}
  \caption{\textbf{Data Efficiency.} In sparse settings, our method requires an average of 51\% fewer images than NeRF and an average of 35\% fewer images than PixelNeRF to achieve a similar test set performance on DTU.}
  \label{data efficiency}
\end{figure}

\subsection{Quantitative Comparisons with SOTA Methods}
\subsubsection{Comparisons on DTU.} Tab.~\ref{Tab1} presents the quantitative results on the DTU dataset. Our model outperforms in most experimental settings, with the exception of the 9-view configuration, where it slightly lags behind FreeNeRF.

\subsubsection{Comparisons on LLFF.} To validate the model's generalization performance, we test our pre-trained model on the LLFF dataset. Following RegNeRF's comparison setup, we conduct extra fine-tuning iterations per scene for each method. The quantitative results of our experiments are presented in Tab.~\ref{Tab2}. While methods like FreeNeRF, which train separate models for each scene, may yield superior results, it's important to consider the overall performance. The LLFF dataset consists of 8 scenes, and FreeNeRF requires retraining for each scene over 250K iterations. In contrast, our model achieves comparable results with a total of no more than 15K fine-tuning iterations. This underscores our model's ability to produce realistic results across different scenes with significantly less computational effort.

\subsection{Qualitative Comparisons with SOTA Methods}
\subsubsection{Comparisons on DTU.} Fig.~\ref{DTU} provides a qualitative comparison between our direct baseline PixelNeRF and our ColNeRF. PixelNeRF exhibits blurriness and shape distortion as it directly feeds the mean of pixel features from all source views into the network, which can introduce negative biases, especially when the projection point falls into occluded regions. In contrast, our model rectifies these errors and implicitly reconstructs the object's geometric shape, leading to improved performance.

\subsubsection{Comparisons on LLFF.} Fig.~\ref{LLFF} presents a comparison of LLFF of our model with FreeNeRF, RegNeRF, and our baseline PixelNeRF. The results from the FreeNeRF and RegNeRF methods are noticeably marred by significant noise, while the PixelNeRF method exhibits a substantial blurring issue. In contrast, our method guarantees accurate and smooth geometric depiction while delivering high-quality rendered images. Further strengthening our claims, Fig.~\ref{depth} illustrates our model's precise geometric control in comparison to DS-NeRF, another method that incorporates explicit depth supervision. DS-NeRF employs depth labels generated with COLMAP\footnote{A universal motion structure (SfM) and multi-view stereo (MVS) pipeline, offering convenient tools for 3D reconstruction.} as constraints for rendering. It can be seen that the inaccuracy of these labels distorts DS-NeRF's geometry understanding. Conversely, our model achieves superior geometric reconstruction and multi-view consistency without additional supervision.

\subsection{Data Efficiency}To assess the data efficiency of our method, we perform a comparative analysis with NeRF and PixelNeRF using different numbers of input views, depicted in Fig.~\ref{data efficiency}. For sparse inputs, our method necessitates up to 51\% fewer input views to attain an equivalent mean PSNR on the test set as that of NeRF, with the disparity being more noticeable for fewer input views. Furthermore, our method delivers performance on par with PixelNeRF, averaging a 35\% reduction in the required input views to yield comparable results.

\begin{table}[t]
\small
\tabcolsep 0.12cm
\begin{tabular}{c|c c c c c|c c c}
\toprule
   & VI & $\mathcal{L}_{geo}$ & $\mathcal{L}_{app}$ & Info & Reg & PSNR↑ & SSIM ↑ & LPIPS ↓ \\ \midrule
   0 & & & & & &18.74 & 0.618 & 0.401 \\
   1 &\checkmark& & & & &19.21 & 0.698 & 0.384\\
   2 &\checkmark&\checkmark & & & &19.39 & \underline{0.714} & 0.375 \\
   3 &\checkmark& &\checkmark & & & \underline{19.48} & 0.710 & \underline{0.373} \\ \midrule
   4 &\checkmark& & & \checkmark & &19.32 & 0.707 & 0.380\\
   5 &\checkmark& & & & \checkmark &18.11 & 0.634 & 0.475\\ \midrule
\textbf{Full}&\checkmark&\checkmark &\checkmark & & & \textbf{19.55} &  \textbf{0.716} &  \textbf{0.362}\\ \bottomrule
\end{tabular}
\caption{\textbf{Ablative results of our model designs on 3-view input DTU.} VI denotes the cross-view volume integration module. Info and Reg respectively denote the Ray Regularization employed in InfoNeRF and RegNeRF.}
\label{Tab3}
\end{table}

\subsection{Ablation Study}
We evaluate the impact of our design choices on the 3-view input DTU dataset in Tab.~\ref{Tab3}. Adding our collaborative cross-view volume integration (CCVI) results in drastically better performance on all metrics. The Ray Regularization was designed to remove potential artifacts in the rendered results. We observe a slight improvement in the results after adding regularization. We also compared the impact of different ray regularization methods from InfoNeRF, RegNeRF and ColNeRF on the same backbone (ColNeRF w/o RayReg).

\section{Limitations and Conclusion}
ColNeRF is designed to have a lightweight network structure.
Based on this consideration, ColNeRF shoots fewer sampling points per ray and adopts a small dimension of the feature volume as well as average pooling.
These factors also lead to some downsides, \textit{i.e.} degrading the reconstruction accuracy with locally smooth renderings.
%
%
%
%
To mitigate these issues, future work can pay attention to various strategies, such as incorporating multi-scale feature volume representations, increasing the utilization of sampling points, or applying frequency regularization constraints.
To conclude, we have introduced ColNeRF, a method capable of achieving photorealistic renderings without using any external data. We have effectively integrated collaborative compensation and constraint into NeRF which leads to accurate 3D modeling with color and geometric consistency. This new route provides strong supervision for model training even in the absence of ground truth. 
Future research may explore faster and more detailed NeRF models.


\appendix

\clearpage


\twocolumn[{%
\centering

{\LARGE \textbf{Supplementary Materials for \\ColNeRF: Collaboration for Generalizable Sparse Input Neural Radiance Field} \par}
\vspace{4em}

}]

\begin{figure*}[h]
    \centering
    \small
    \begin{minipage}{\textwidth}%
        \centering
        \makebox[0.19\linewidth]{NeRF}
        \makebox[0.19\linewidth]{RegNeRF}
        \makebox[0.19\linewidth]{PixelNeRF}
        \makebox[0.19\linewidth]{\textbf{Ours}}
        \makebox[0.19\linewidth]{Ground Truth}
        \\
        \includegraphics[width=0.19\linewidth]{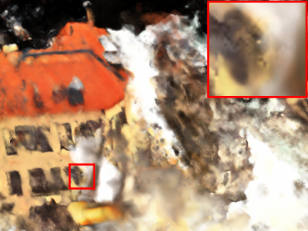}
        \includegraphics[width=0.19\linewidth]{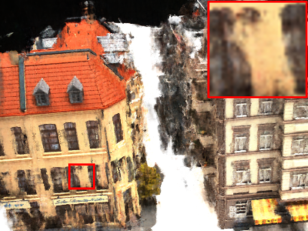}
        \includegraphics[width=0.19\linewidth]{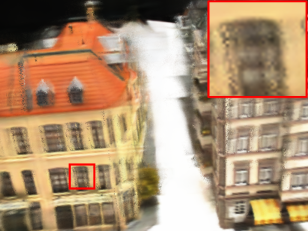}
        \includegraphics[width=0.19\linewidth]{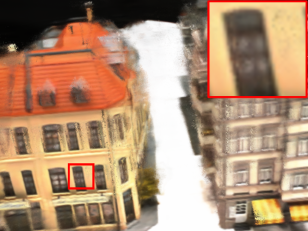}
        \includegraphics[width=0.19\linewidth]{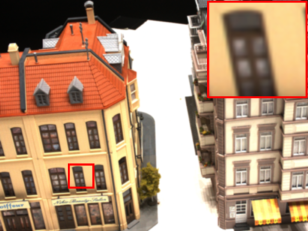}
        \\
        \includegraphics[width=0.19\linewidth]{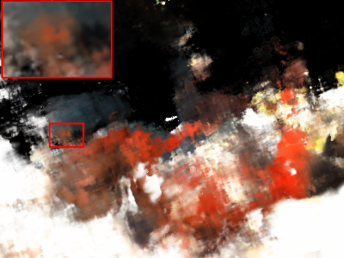}
        \includegraphics[width=0.19\linewidth]{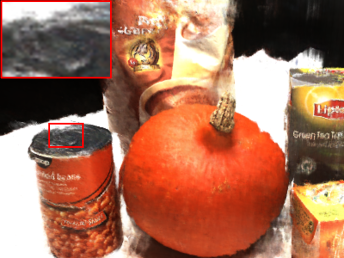}
        \includegraphics[width=0.19\linewidth]{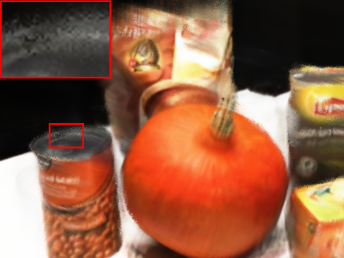}
        \includegraphics[width=0.19\linewidth]{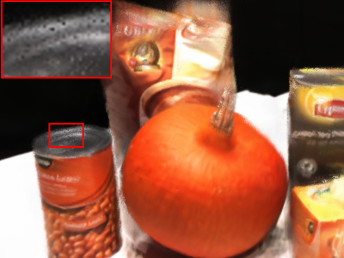}
        \includegraphics[width=0.19\linewidth]{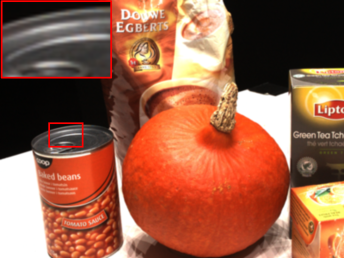}
        \\
        \includegraphics[width=0.19\linewidth]{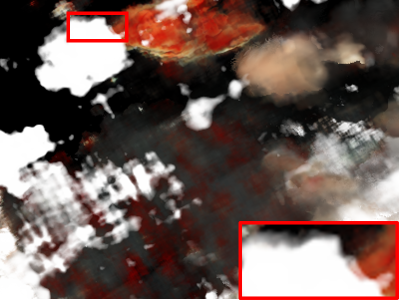}
        \includegraphics[width=0.19\linewidth]{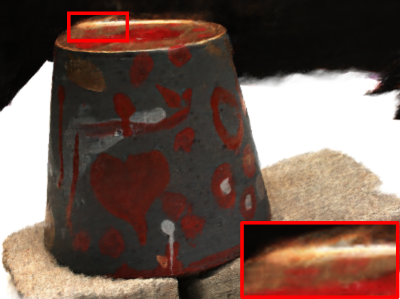}
        \includegraphics[width=0.19\linewidth]{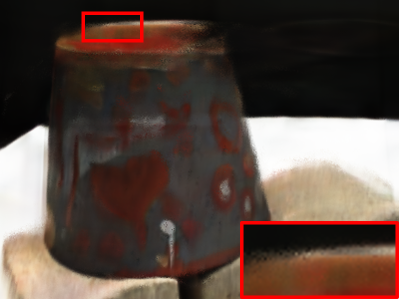}
        \includegraphics[width=0.19\linewidth]{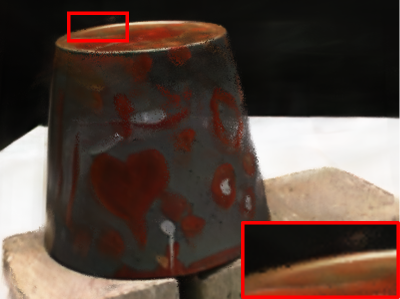}
        \includegraphics[width=0.19\linewidth]{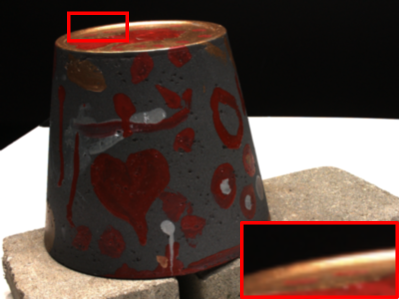}
        \\
        \includegraphics[width=0.19\linewidth]{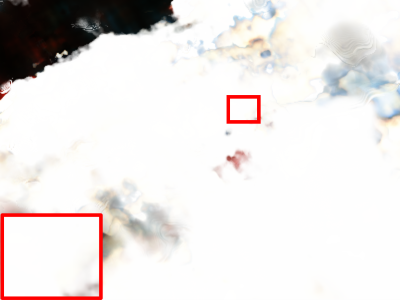}
        \includegraphics[width=0.19\linewidth]{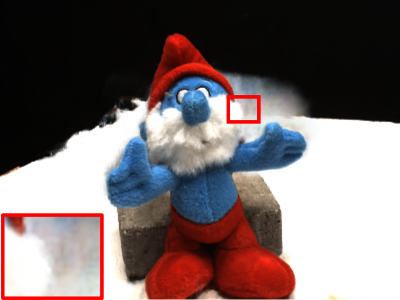}
        \includegraphics[width=0.19\linewidth]{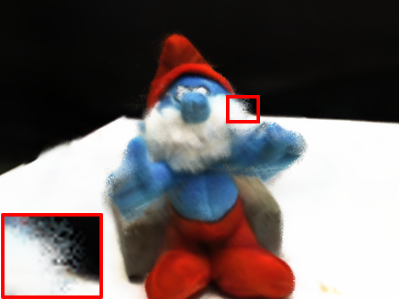}
        \includegraphics[width=0.19\linewidth]{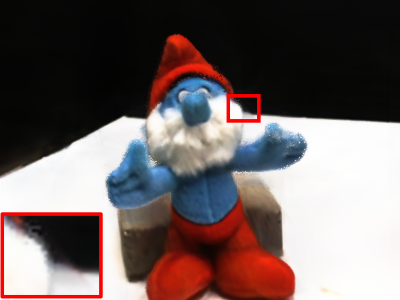}
        \includegraphics[width=0.19\linewidth]{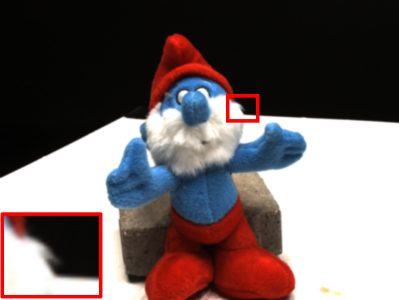}
        \\
        \includegraphics[width=0.19\linewidth]{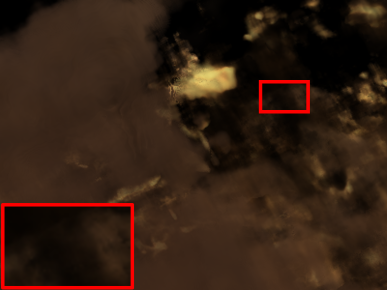}
        \includegraphics[width=0.19\linewidth]{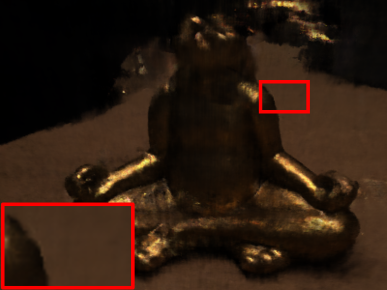}
        \includegraphics[width=0.19\linewidth]{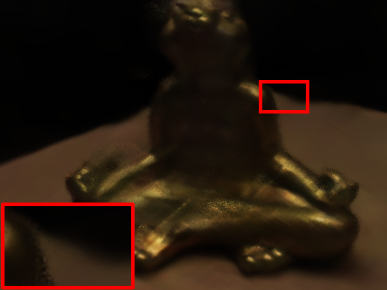}
        \includegraphics[width=0.19\linewidth]{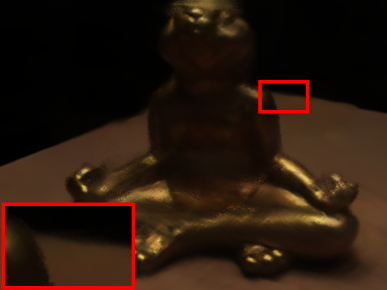}
        \includegraphics[width=0.19\linewidth]{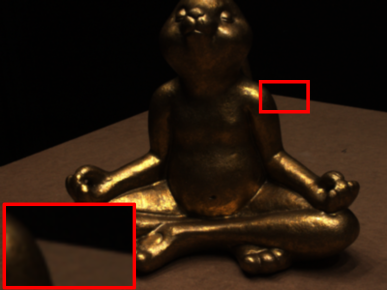}
        
    \end{minipage}
    \caption{
    \textbf{Additional 3-View Input Qualitative Comparison Results on DTU.}
    }
    \label{DTU_3V_comp}
\end{figure*}

\begin{figure*}[t] 
    \centering
    \small
    \begin{minipage}{\textwidth}%
        \centering       
        \makebox[0.19\linewidth]{NeRF}
        \makebox[0.19\linewidth]{RegNeRF}
        \makebox[0.19\linewidth]{PixelNeRF}
        \makebox[0.19\linewidth]{\textbf{Ours}}
        \makebox[0.19\linewidth]{Ground Truth}
        \\
        \includegraphics[width=0.19\linewidth]{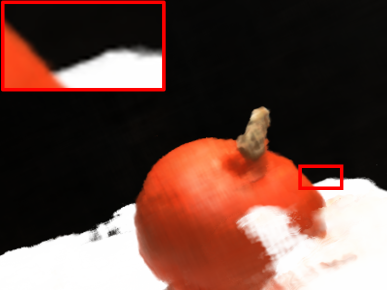}
        \includegraphics[width=0.19\linewidth]{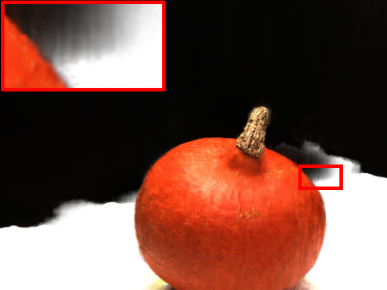}
        \includegraphics[width=0.19\linewidth]{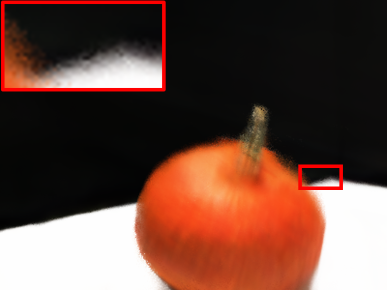}
        \includegraphics[width=0.19\linewidth]{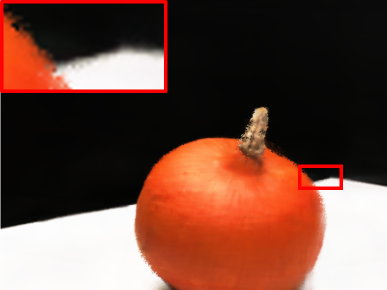}
        \includegraphics[width=0.19\linewidth]{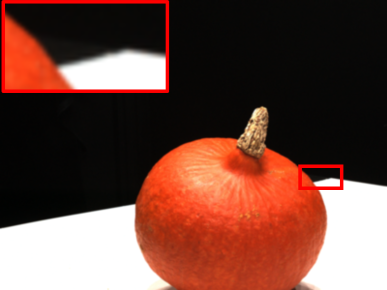}
        \\
        \includegraphics[width=0.19\linewidth]{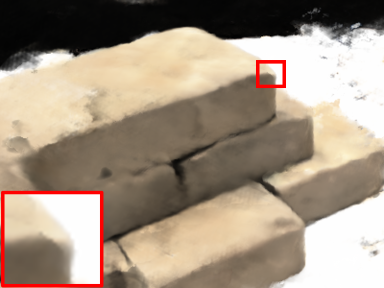}
        \includegraphics[width=0.19\linewidth]{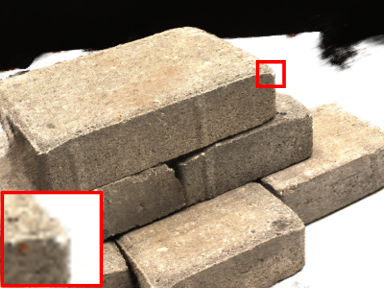}
        \includegraphics[width=0.19\linewidth]{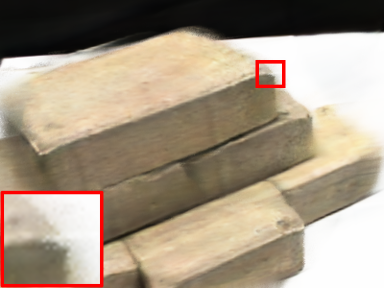}
        \includegraphics[width=0.19\linewidth]{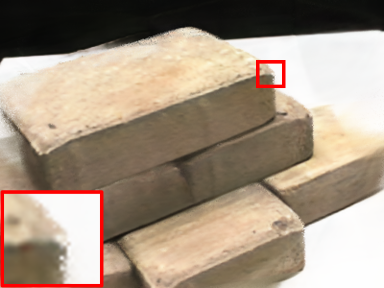}
        \includegraphics[width=0.19\linewidth]{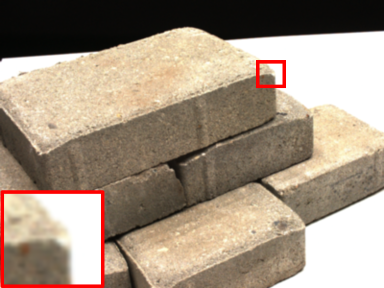}
        \\
        \includegraphics[width=0.19\linewidth]{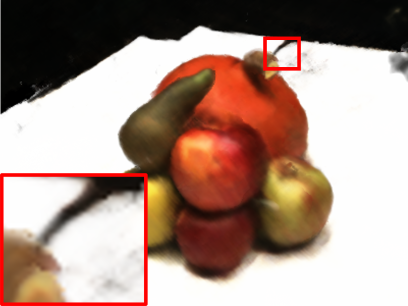}
        \includegraphics[width=0.19\linewidth]{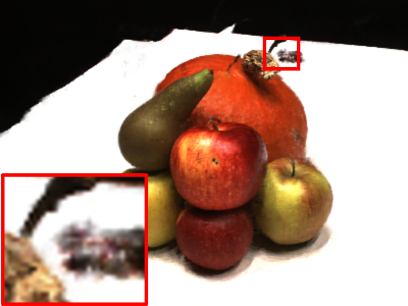}
        \includegraphics[width=0.19\linewidth]{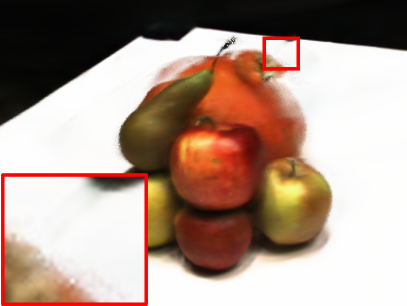}
        \includegraphics[width=0.19\linewidth]{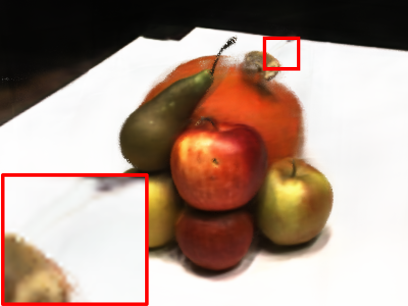}
        \includegraphics[width=0.19\linewidth]{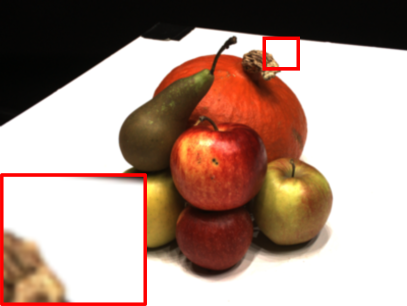}
        \\
        \includegraphics[width=0.19\linewidth]{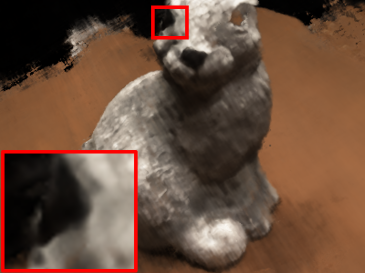}
        \includegraphics[width=0.19\linewidth]{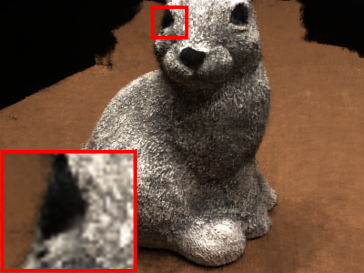}
        \includegraphics[width=0.19\linewidth]{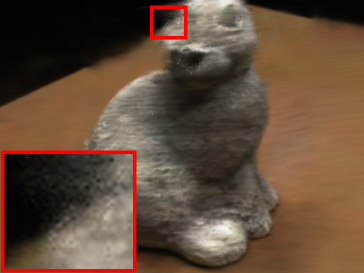}
        \includegraphics[width=0.19\linewidth]{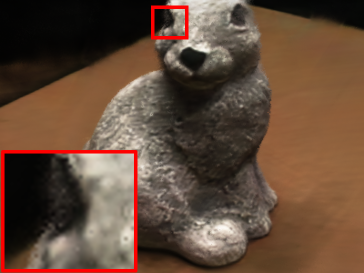}
        \includegraphics[width=0.19\linewidth]{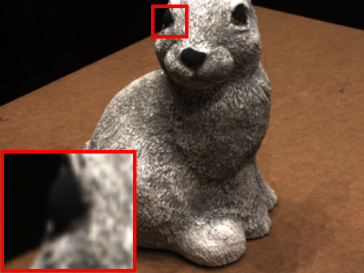}
        \\
        \includegraphics[width=0.19\linewidth]{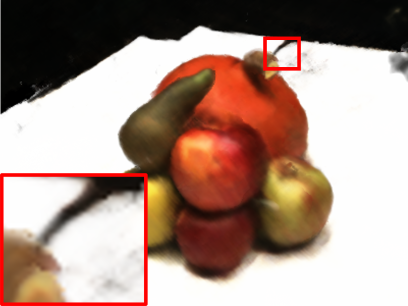}
        \includegraphics[width=0.19\linewidth]{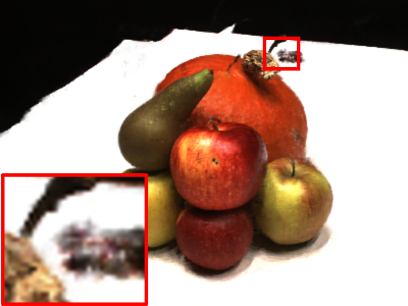}
        \includegraphics[width=0.19\linewidth]{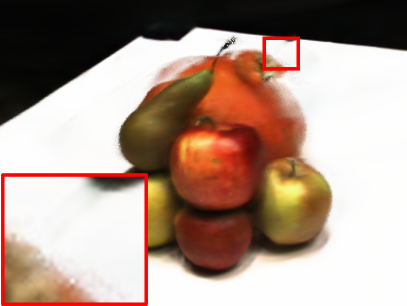}
        \includegraphics[width=0.19\linewidth]{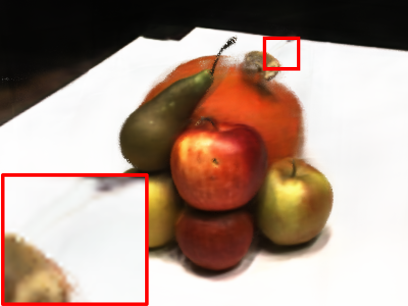}
        \includegraphics[width=0.19\linewidth]{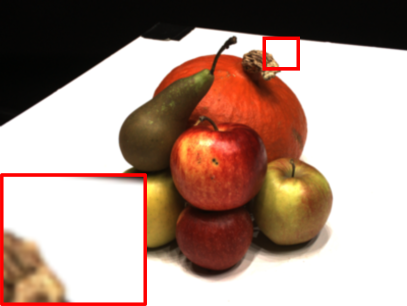}
        
	\end{minipage}
    \caption{
    \textbf{Additional 6-View Input Qualitative Comparison Results on DTU.}
    }
    \label{DTU_6V_comp}
\end{figure*}

\begin{table*}[t]
\small
\begin{tabular}{l|c|cc|cc|cc|cc}
    \toprule
      \multirow{2}{*}{Method}  & 
      \multirow{2}{*}{Setting} & 
      \multicolumn{2}{c|}{PSNR ↑} &
      \multicolumn{2}{c|}{SSIM ↑} &
      \multicolumn{2}{c|}{LPIPS ↓} &
      \multicolumn{2}{c}{Average ↓} \\
     & &3-view & 6-view & 3-view & 6-view & 3-view & 6-view & 3-view & 6-view   \\
      
    \midrule
    DietNeRF (ICCV 2021)& \multirow{3}{*}{Optimized per Scene} & 14.94 & 21.75 &0.370 &0.717&0.496 &0.248 &0.240 &0.105  \\
    RegNeRF (CVPR 2022)& & 19.08 &23.10  &\underline{0.587} &0.760 &\underline{0.336}&\underline{0.206}&0.149 &0.086 \\
    FreeNeRF (CVPR 2023)& & \underline{19.63} &\underline{23.73} & \underline{0.612} & \underline{0.779}&  \textbf{0.308}& \textbf{0.195}& \underline{0.134}& \textbf{0.075} \\ \midrule
     \textbf{ColNeRF ft (Ours)} & \multirow{1}{*}{Not Optimized per Scene} & \textbf{21.16} & \textbf{23.91} & \textbf{0.626} & \textbf{0.783} & 0.405 & 0.259 & \textbf{0.124} & \underline{0.078} \\
    \bottomrule
\end{tabular}
\caption{\textbf{Additional Quantitative Comparison on LLFF}. To show that our model does not falter in performance compared to scene-specific training methods, we conduct an additional trained on LLFF and evaluated on LLFF experiment. However, it is worth noting that our experiment is still not based on scene-by-scene training, meaning that our model not only retains generalization capabilities but can also render high-quality results. The table above presents the results under 3 and 6 views. As can be seen from the table, our model can generally achieve the best results compared to other methods.}
\label{Tab 1}
\end{table*}


This supplementary material presents (1) additional experimental details and (2) additional comparisons of quantitative and qualitative results between the proposed ColNeRF and other state-of-the-art methods.


\section{A. Experiment Details}

To comprehensively evaluate the performance of the proposed Collaborative Neural Radiance Fields (ColNeRF), we conduct extensive experiments following the experimental settings of PixelNeRF~\cite{yu2021pixelnerf}.
For a more detailed comparison, we provide some additional details in the following sections.


\subsection{A.1. Dataset and metrics} 

\subsubsection{DTU Dataset} 
The original DTU dataset~\cite{jensen2014large} consists of an extensive collection captured within controlled laboratory conditions, captured in 128 scenes. Utilizing a structured light scanner, this dataset generates models by scanning each scene from 49 or 64 identical camera positions, in seven distinct lighting conditions.
Therefore, RGB images with a resolution of $1200\times 1600$ pixels are obtained.
The DTU dataset is diverse, containing a large number of objects with diverse textures, materials, and geometries.
However, the analysis in PixelNeRF~\cite{yu2021pixelnerf} illustrates that the DTU dataset shows significant overlap between the original training, validation and test sets.
To address this, PixelNeRF establish an alternative split: designating scans 8, 21, 30, 31, 34, 38, 40, 41, 45, 55, 63, 82, 103, 110, 114 for testing, while all other scans, excluding 1, 2, 7, 25, 26, 27, 29, 39, 51, 54, 56, 57, 58, 73, 83, 111, 112, 113, 115, 116, and 117, are adopted for training.
Moreover, for efficient training, PixelNeRF downsizes all DTU images to $400\times 300$ resolution and scales the world coordinates of all scans by a factor of 1/300.
In our experiments, we also adopt these experimental settings from PixelNeRF.

\subsubsection{LLFF Dataset} 
The Light Field from Frames (LLFF) dataset serves as a comprehensive repository of image sets tailored for tasks such as light field rendering and depth estimation. 
With a diverse range of indoor and outdoor scenarios, this dataset is particularly distinguished by its focus on ``forward-facing'' scenes.
Including 8 scenes captured using a handheld cellphone, five sourced from the LLFF paper~\cite{mildenhall2019local} and three from the original NeRF study~\cite{mildenhall2021nerf}, the LLFF dataset includes from 20 to 62 images per scene. 
In accordance with previous studies, we designate every 8-th image as part of the held-out test set, while the input views are randomly selected from the remaining images. 
All images have been downsampled to $504\times 378$ pixels consistently.

\subsubsection{Metrics} In our experiments, we employ the API of scikit-image to calculate the Structural Similarity Index Measure (SSIM) and the Peak Signal-to-Noise Ratio (PSNR) scores. 
Additionally, we use the code provided by the original authors that take a trained VGG model to compute a Learned Perceptual Image Patch Similarity (LPIPS) score.
Furthermore, we compute the geometric mean of $\rm MSE = 10^{-\rm PSNR/10}$, $\sqrt{1 - \rm SSIM}$, and $\rm LPIPS$ similar to the approach employed in RegNeRF~\cite{niemeyer2022regnerf}. 
The geometric mean, a form of averaging that considers the product of values rather than their sum, proves particularly valuable in scenarios where the numbers exhibit varying ranges or signify distinct measures. 
By computing the geometric mean of these three measures, we provide a unified score that captures the quality of the reconstruction (via PSNR), the similarity to the original image (via SSIM), and the perceptual similarity (via LPIPS). 


\subsection{A.2. Implementations}


\subsubsection{Encoder} As briefly discussed in the main paper, we use a ResNet34 backbone as the encoder. For an image of size $H\times W$, we extract a feature pyramid by taking the feature maps before the first pooling operation and after the first three ResNet stages. The feature representations obtained from these stages, in four distinct dimensions, are upsampled to match the dimensions of $H/2\times W/2$ and then concatenated. This concatenation results in final feature maps with dimensions of $512\times H/2\times W/2$.

\subsubsection{Collaborative Integration Network} 
It integrates feature maps from different input images in a collaborative manner, enhancing the scene modeling capability of the model.
In each fusion operation, one of the feature maps is selected as the anchor, with the rest serving as auxiliaries.
We first aggregate the auxiliary feature maps through summation and concatenate them with the anchor feature map. 
This aggregation results in a combined feature volume with dimensions of 1024.
Subsequently, we employ a convolutional layer to reduce this combined feature volume to 512 dimensions. 
The obtained feature volume is utilized as both query and key components in our proposed collaborative cross-view volume integration module (CCVI), while the original anchor feature volume serves as the values of CCVI.
It is noted that with our well-designed CCVI, we can achieve the collaborative fusion of arbitrary input image features.

The proposed CCVI conducts attention computation at the patch level.
We use local attention to emphasize the most important areas within each patch. 
However, simply dividing each feature map into supplementary patches and performing attention within them prevents the transfer of information between adjacent blocks.
For pixels near the boundary of a block, despite being very close to pixels in adjacent blocks, their attention computation cannot gather information from those blocks. 
To address this, we opt to expand each patch by a band of pixels, resulting in a block size of $(s + 2a) \cdot (s + 2a)$, where $s$ represents the patch width, and $a$ is the band width.
In cases where adjacent blocks exist, the band pixels are copied from these neighbouring blocks; otherwise, they are zero-padded.
This expansion allows for information exchange between adjacent blocks.
Notably, this expansion is only applied to the query and key feature volumes.
Specifically, for the DTU dataset, we utilize a patch size $s$ of 5 and a band width $a$ of 3. As for the LLFF dataset, we employ a patch size $s$ of 9 along with a band width $a$ of 3.
The feature volume after the attention operation is combined with the residual of the anchor feature to get $\rm \hat{F}_{ref }$.
Subsequently, after processing through a feed-forward network consisting of two convolutional layers and adding residuals of $\rm \hat{F}_{ref}$, we arrive at the final affined feature volume of the anchor.

\subsubsection{Position Encoding} 
Similar to the approach of the vanilla NeRF,  we apply position encoding to the coordinates before feeding them into the neural radiance field to capture finer high-frequency details, which can be formulated as:
\begin{multline}
    \gamma(x) = (sin(2^0 \omega x),cos(2^0 \omega x),...,\\
    sin(2^{L-1} \omega x),cos(2^{L-1} \omega x)).
\end{multline}
In our experiments, we utilize the configuration of PixelNeRF~\cite{yu2021pixelnerf} to set the parameter $L=6$, $\omega=1.5$, and only process the original three-dimensional spatial coordinates $\rm x =(x,y,z)$ to obtain a new 39 dimension coordinate vector, while the original 3-dimensional directional features are not position encoded.

\subsubsection{Neural Radiance Network} 
We adopt the neural radiance field network structure of PixelNeRF, which employs a fully connected ResNet architecture with 5 ResNet blocks and a width of 512. 
The inputs to the neural radiance field are composed of three distinct components: a 39-dimensional coordinate vector, a 3-dimensional directional vector, and 512-dimensional pixel features. 
The extraction of pixel features involves projecting the original three-dimensional coordinate vector onto the affined feature volume, thereby generating a $1 \times 1 \times 512$ pixel feature representation.
For $N$ source views' feature volumes, pixel features of each source view are first fused with the coordinate and direction vectors in the first three ResNet blocks.
The fused results for each source view feature are averaged across the N-dimension after the third block, and further fused in the final two ResNet blocks.

\subsubsection{Sampling Strategy} 
We have observed that too concentrated ray casting during training will lead to overfitting to a particular scene or background section. 
Therefore, we follow PixelNeRF to adopt the random sampling strategy, including arbitrary camera origins and shooting directions.
In each training iteration, we first randomly select a training scene, randomly select source views from all views in the scene, and randomly cast 128 rays into the scene for training.

We adopt a two-stage training strategy. Firstly, we use a coarse network training stage to capture the scene structure roughly.
For this purpose, we utilize a stratified sampling approach where we partition $[ t_n, t_f ]$ into 64 evenly-spaced bins, and then draw one sample uniformly at random from within each bin:
\begin{equation}
    t_i \sim \mathcal{U}\left[t_n+\frac{i-1}{N}(t_f-t_n),t_n+\frac{i}{N}(t_f-t_n)\right].
\end{equation}
Subsequently, in the second stage of fine network training, we employ the outcomes of the coarse network to determine the sampling probability density in a refinement stage. This probability density determines the most critical sampling regions along each ray.

To facilitate more effective comparative experiments, we generate pairs of adjacent rays for the purpose of geometric self-supervision between these rays.
Adding extra training rays directly would result in elevated training expenses, and generating an excessive number of neighboring ray pairs might compromise the generalization capacity of our ColNeRF.
Therefore, we choose to randomly sample 112 rays from the 128 rays needed for sampling. 
Then, we select the last 16 of these 112 rays as reference rays and emit the remaining 16 rays within a region offset by no more than 7 pixels from these reference rays. 
This approach ensures that the final 32 out of the 128 rays are employed as mutually supervisory pairs for the purpose of Geometry Regularization.

\subsubsection{Hyperparameters} 
For all experiments, we set the learning rate at $10^{-4}$. 
The batch size is set to 3, 3, and 2 for 3-view, 6-view, and 9-view settings, respectively.
Our proposed ColNeRF is trained on the DTU dataset.
In the case of input viewpoints is 3, the ColNeRF is trained for 250K iterations.
The network converges quickly when the number of input viewpoints is 6 and 9, and we train the network for approximately 150K and 100K iterations, respectively. For model selection, we choose the latest model.

When testing on the DTU dataset, we choose a fixed sequence of source views. Specifically, we opt for source views 25, 22, 28, 40, 44, 48, 0, 8, 13. This sequence is designated for utilization with varying quantities of source views, in cases where the number of source views $N$ is less than that of a 9-view setup, we employ the initial sequence's first $N$ viewpoints as the set of source views.
We exclude views with poor exposure (specifically, views 3, 4, 5, 6, 7, 16, 17, 18, 19, 20, 21, 36, 37, 38, 39) from the testing. When testing on the LLFF dataset, choosing a fixed rendered view is challenging due to the varying number of views for different scenes. Therefore, we choose a randomly selected sequence of source views.

\begin{figure*}[htp] 
    \centering
    \small
    \begin{minipage}{\textwidth}%
        \centering      
        \makebox[0.19\linewidth]{RegNeRF}
        \makebox[0.19\linewidth]{FreeNeRF}
        \makebox[0.19\linewidth]{PixelNeRF}
        \makebox[0.19\linewidth]{\textbf{Ours}}
        \makebox[0.19\linewidth]{Ground Truth}
        \\
        \includegraphics[width=0.19\linewidth]{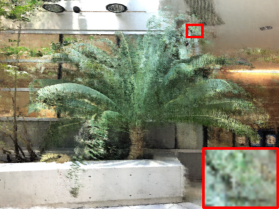}
        \includegraphics[width=0.19\linewidth]{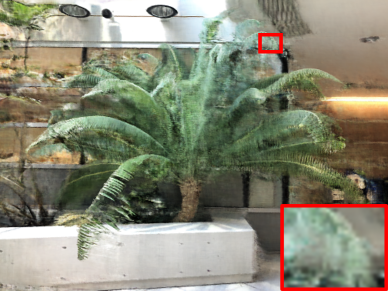}
        \includegraphics[width=0.19\linewidth]{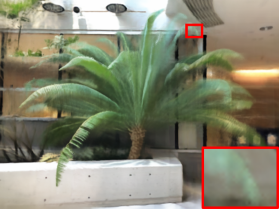}
        \includegraphics[width=0.19\linewidth]{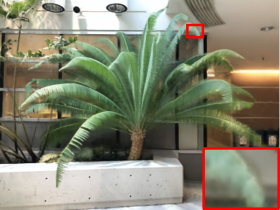}
        \includegraphics[width=0.19\linewidth]{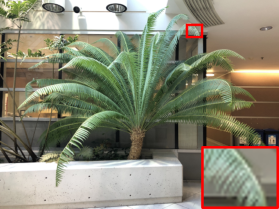}
        \\
        \includegraphics[width=0.19\linewidth]{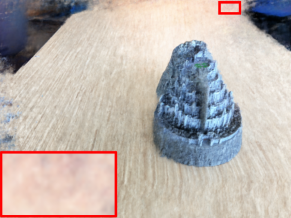}
        \includegraphics[width=0.19\linewidth]{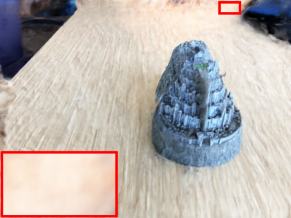}
        \includegraphics[width=0.19\linewidth]{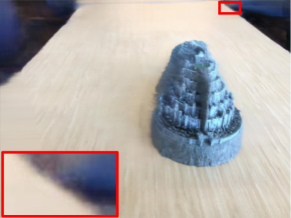}
        \includegraphics[width=0.19\linewidth]{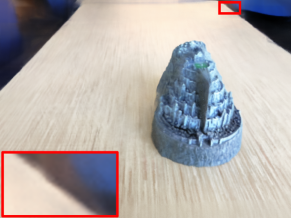}
        \includegraphics[width=0.19\linewidth]{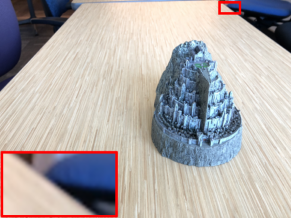}
        \\
        \includegraphics[width=0.19\linewidth]{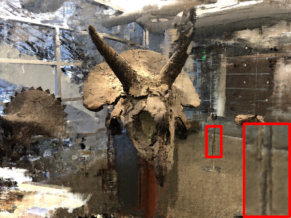}
        \includegraphics[width=0.19\linewidth]{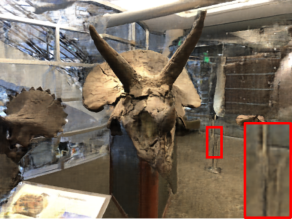}
        \includegraphics[width=0.19\linewidth]{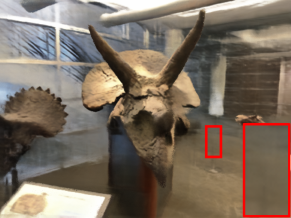}
        \includegraphics[width=0.19\linewidth]{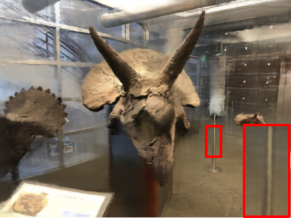}
        \includegraphics[width=0.19\linewidth]{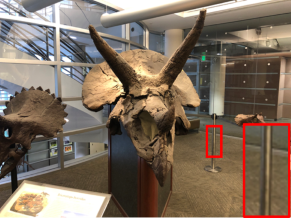}
        \\
        \includegraphics[width=0.19\linewidth]{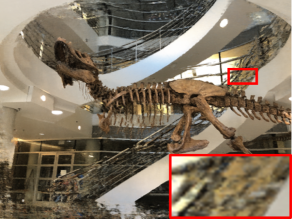}
        \includegraphics[width=0.19\linewidth]{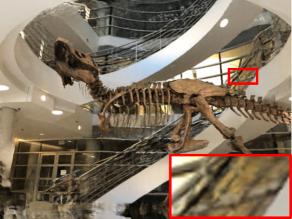}
        \includegraphics[width=0.19\linewidth]{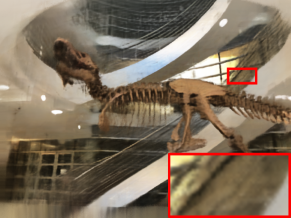}
        \includegraphics[width=0.19\linewidth]{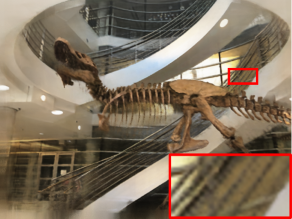}
        \includegraphics[width=0.19\linewidth]{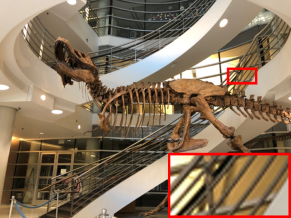}
        
	\end{minipage}
    \caption{
    \textbf{Additional 3-View Input Qualitative Comparison Results on LLFF.}
    }
    \label{LLFF_3V_comp}
\end{figure*}

\begin{figure*}[htp] 
    \centering
    \small
    \begin{minipage}{\textwidth}
        \centering
        \makebox[0.05\linewidth]{ }
        \makebox[0.22\linewidth]{DS-NeRF}
        \makebox[0.22\linewidth]{Ours}
        \hspace{0.5em}
        \makebox[0.22\linewidth]{DS-NeRF}
        \makebox[0.22\linewidth]{Ours}
        \\
        \raisebox{4.2em}{
          \begin{minipage}{0.05\linewidth}
            RGB
          \end{minipage}
        }
        \includegraphics[width=0.215\linewidth]{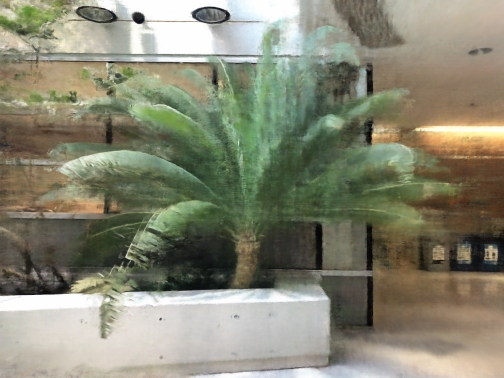}
        \includegraphics[width=0.215\linewidth]{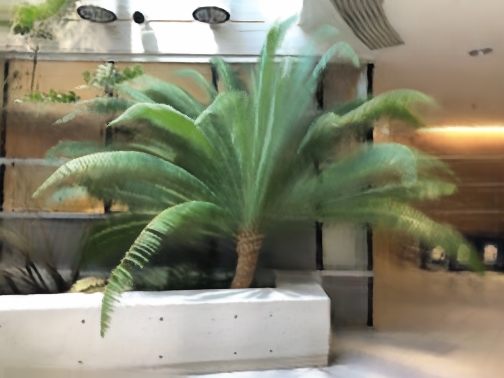}
        \hspace{0.5em}
        \includegraphics[width=0.215\linewidth]{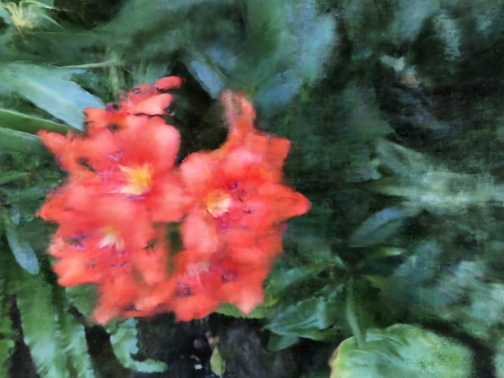}
        \includegraphics[width=0.215\linewidth]{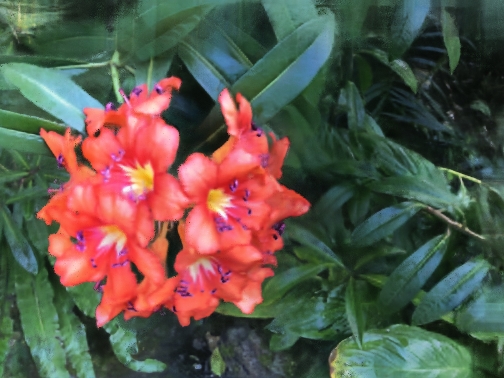}
        \\
        \raisebox{4.2em}{
          \begin{minipage}{0.0535\linewidth}
            Depth
          \end{minipage}
        }
        \includegraphics[width=0.215\linewidth]{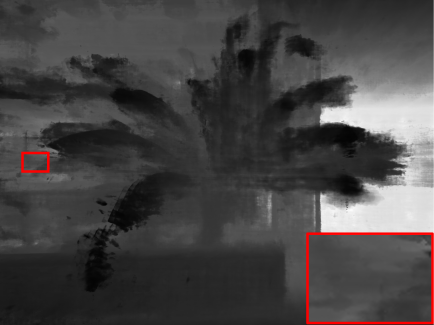}
        \includegraphics[width=0.215\linewidth]{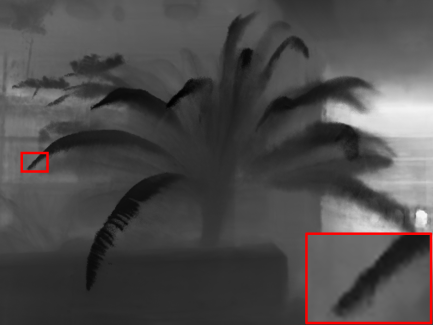}
        \hspace{0.5em}
        \includegraphics[width=0.215\linewidth]{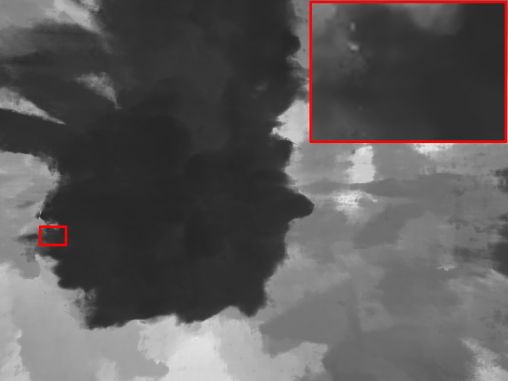}
        \includegraphics[width=0.215\linewidth]{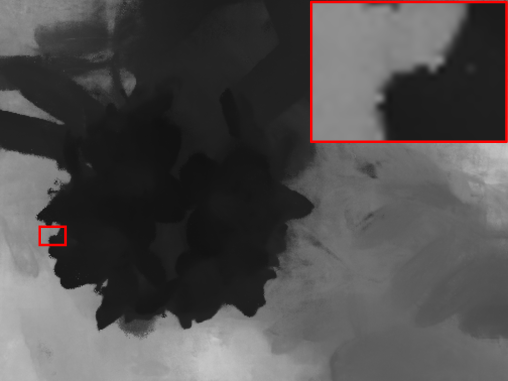}
        \vspace{0.5em}
        \\
        \raisebox{4.2em}{
          \begin{minipage}{0.05\linewidth}
            RGB
          \end{minipage}
        }
        \includegraphics[width=0.215\linewidth]{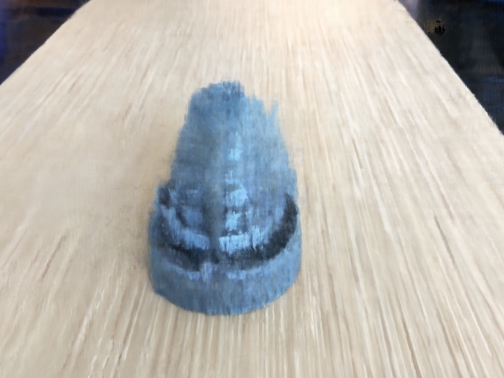}
        \includegraphics[width=0.215\linewidth]{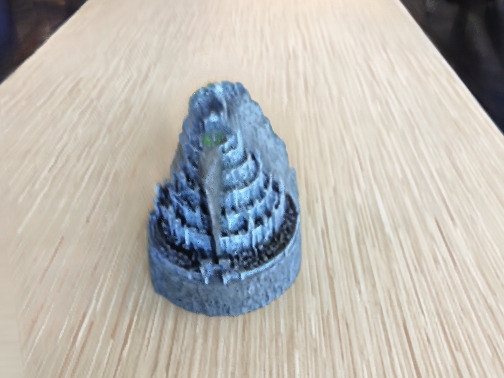}
        \hspace{0.5em}
        \includegraphics[width=0.215\linewidth]{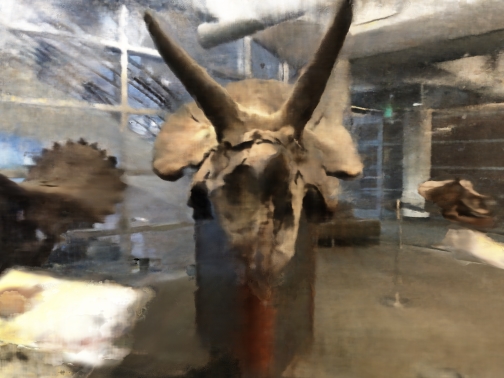}
        \includegraphics[width=0.215\linewidth]{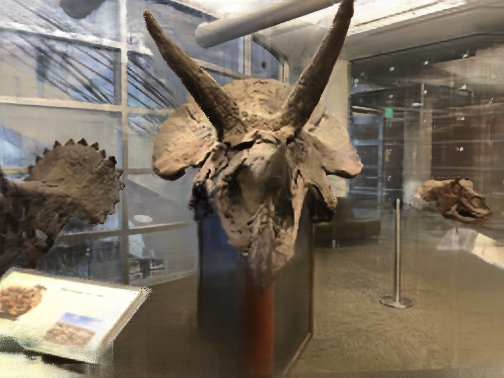}
        \\
        \raisebox{4.2em}{
          \begin{minipage}{0.0535\linewidth}
            Depth
          \end{minipage}
        }
        \includegraphics[width=0.215\linewidth]{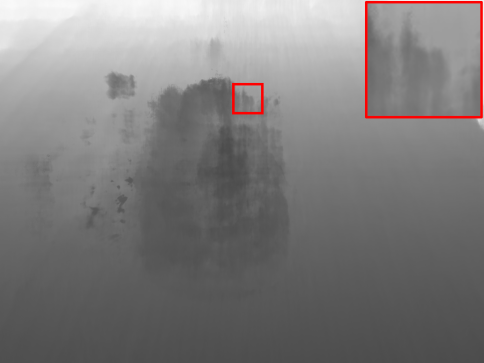}
        \includegraphics[width=0.215\linewidth]{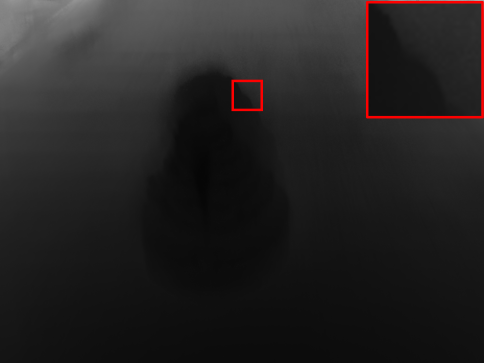}
        \hspace{0.5em}
        \includegraphics[width=0.215\linewidth]{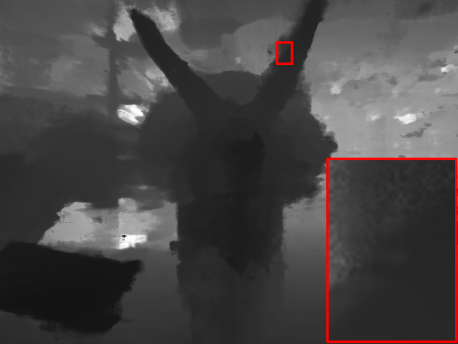}
        \includegraphics[width=0.215\linewidth]{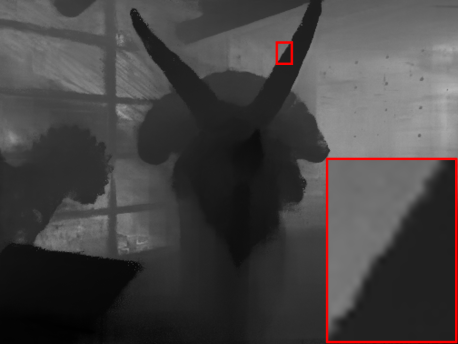}
        \vspace{0.5em}
        \\
        \raisebox{4.2em}{
          \begin{minipage}{0.05\linewidth}
            RGB
          \end{minipage}
        }
        \includegraphics[width=0.215\linewidth]{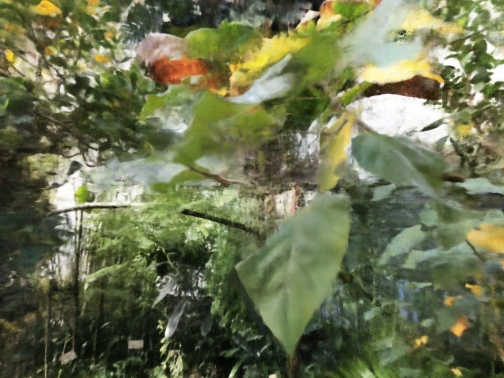}
        \includegraphics[width=0.215\linewidth]{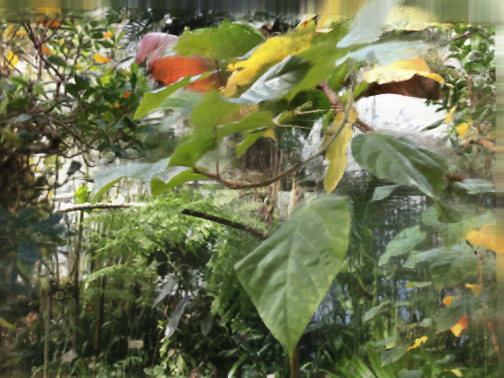}
        \hspace{0.5em}
        \includegraphics[width=0.215\linewidth]{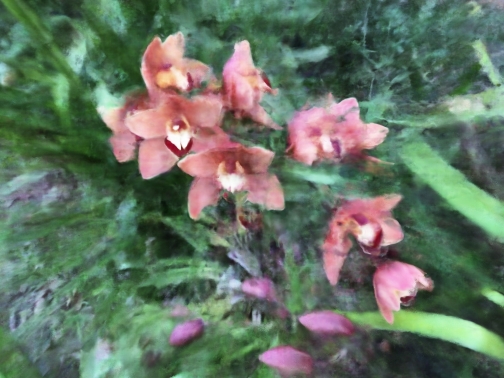}
        \includegraphics[width=0.215\linewidth]{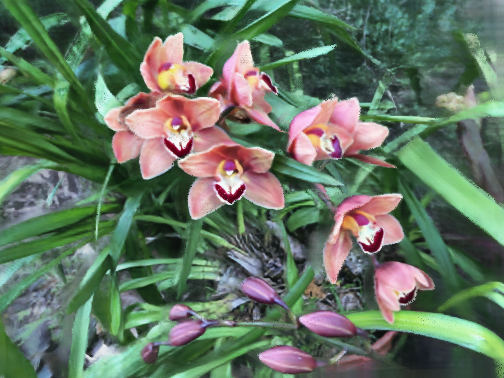}
        \\
        \raisebox{4.2em}{
          \begin{minipage}{0.0535\linewidth}
            Depth
          \end{minipage}
        }
        \includegraphics[width=0.215\linewidth]{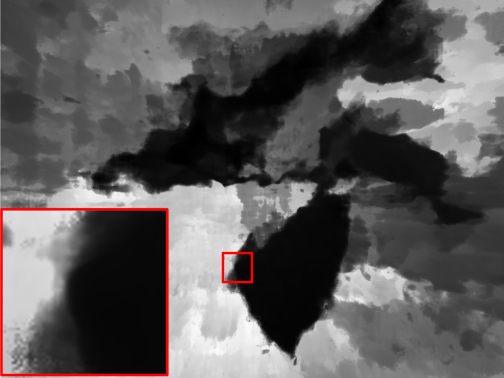}
        \includegraphics[width=0.215\linewidth]{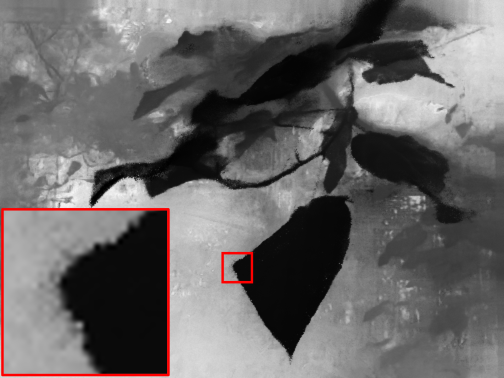}
        \hspace{0.5em}
        \includegraphics[width=0.215\linewidth]{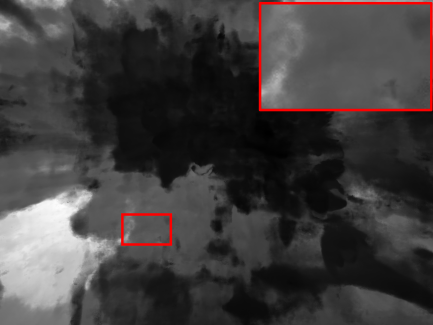}
        \includegraphics[width=0.215\linewidth]{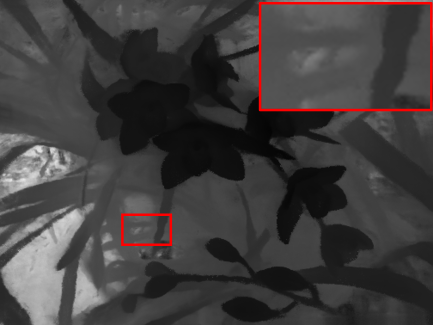}
        \vspace{0.5em}
        \\
        \raisebox{4.2em}{
          \begin{minipage}{0.05\linewidth}
            RGB
          \end{minipage}
        }
        \includegraphics[width=0.215\linewidth]{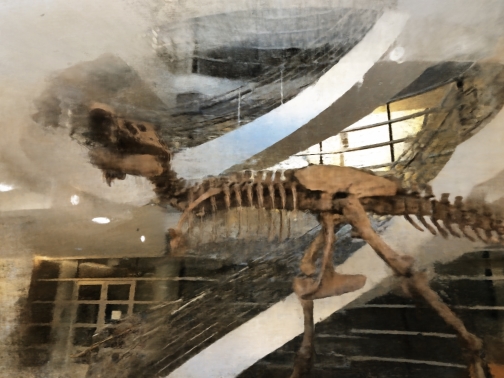}
        \includegraphics[width=0.215\linewidth]{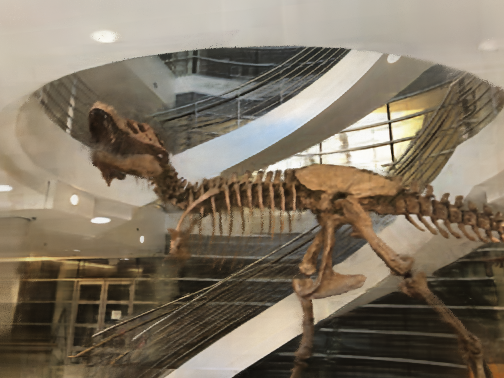}
        \hspace{0.5em}
        \includegraphics[width=0.215\linewidth]{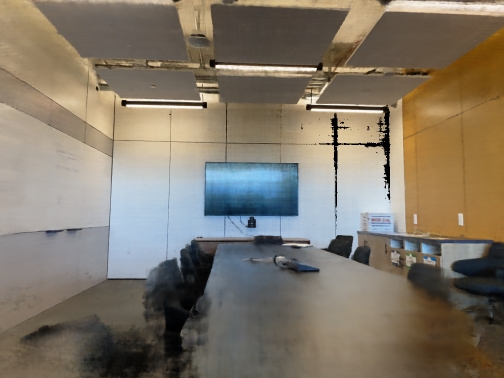}
        \includegraphics[width=0.215\linewidth]{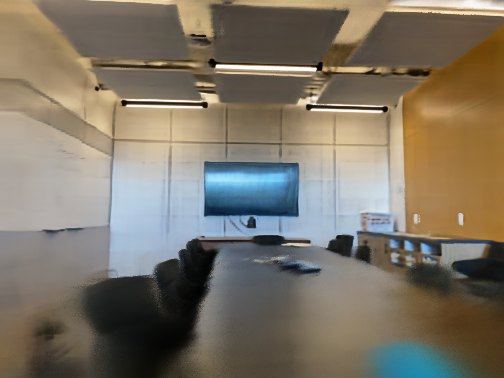}
        \\
        \raisebox{4.2em}{
          \begin{minipage}{0.0535\linewidth}
            Depth
          \end{minipage}
        }
        \includegraphics[width=0.215\linewidth]{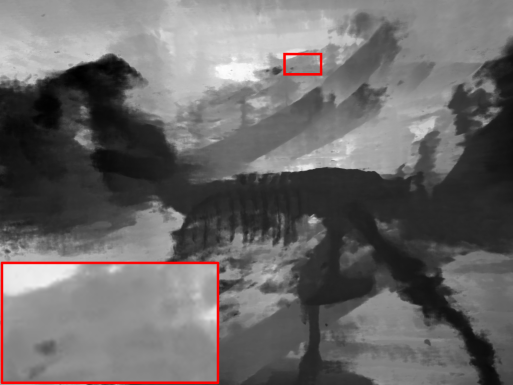}
        \includegraphics[width=0.215\linewidth]{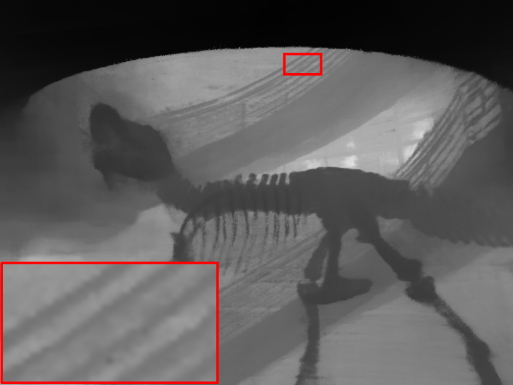}
        \hspace{0.5em}
        \includegraphics[width=0.215\linewidth]{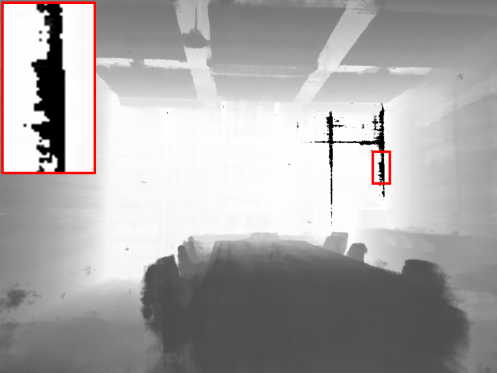}
        \includegraphics[width=0.215\linewidth]{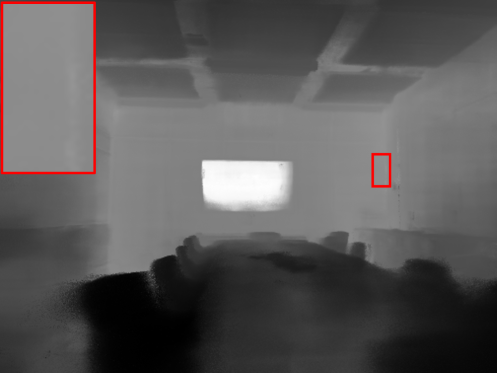}
        \vspace{0.5em}
    \end{minipage}
    \caption{
    \textbf{Additional 2-View Input Rendered Depth Comparisons on LLFF.}
    }
    \label{LLFF_depth}
\end{figure*}

\begin{figure*}[htp] 
    \centering
    \small
    \begin{minipage}{\textwidth}%
        \centering
        \makebox[0.32\textwidth]{3 Views}
        \hspace{0.0015\textwidth}
        \makebox[0.32\textwidth]{6 Views}
        \hspace{0.0015\textwidth}
        \makebox[0.32\textwidth]{9 Views}
        \\
        \includegraphics[height=0.12\linewidth]{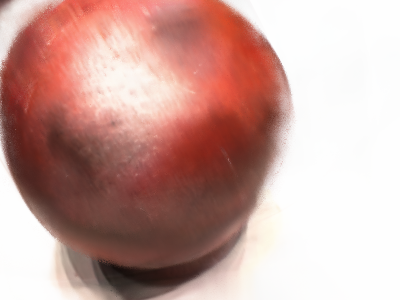}
        \includegraphics[height=0.12\linewidth]{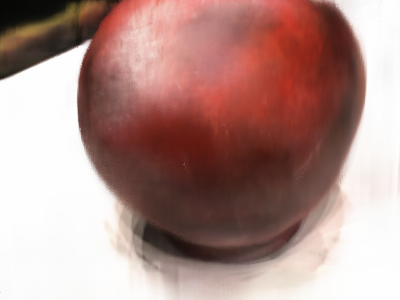}
        \hspace{0.0015\textwidth}
        \includegraphics[height=0.12\linewidth]{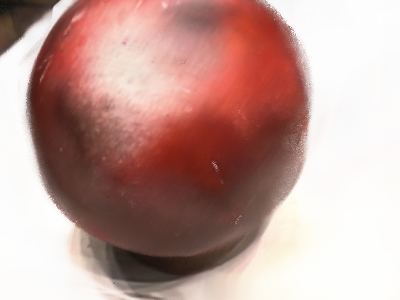}
        \includegraphics[height=0.12\linewidth]{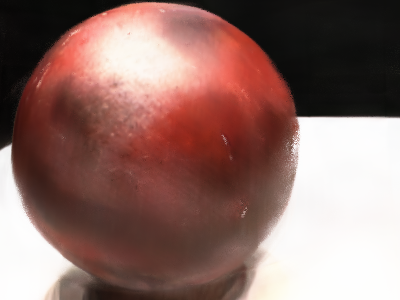}
        \hspace{0.0015\textwidth}
        \includegraphics[height=0.12\linewidth]{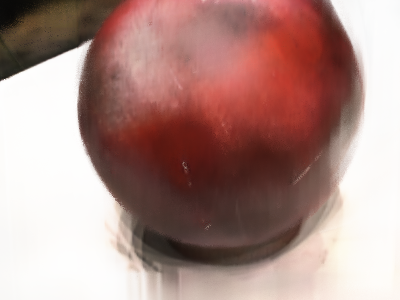}
        \includegraphics[height=0.12\linewidth]{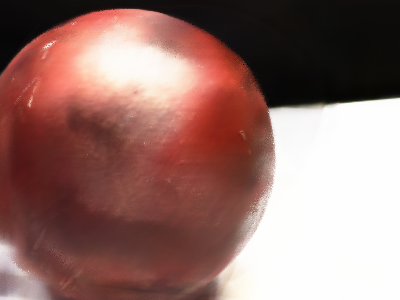}
        \\
        \includegraphics[height=0.12\linewidth]{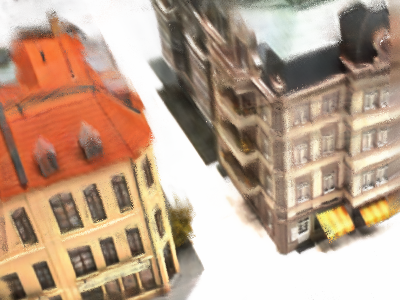}
        \includegraphics[height=0.12\linewidth]{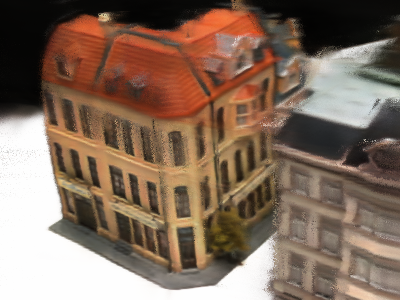}
        \hspace{0.0015\textwidth}
        \includegraphics[height=0.12\linewidth]{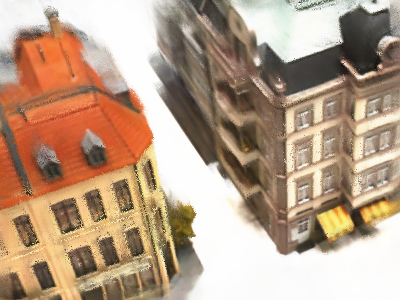}
        \includegraphics[height=0.12\linewidth]{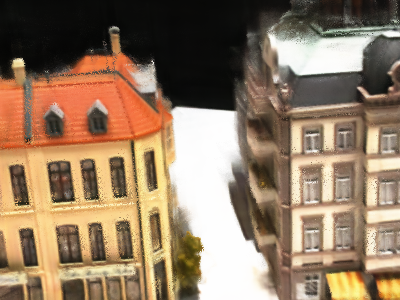}
        \hspace{0.0015\textwidth}
        \includegraphics[height=0.12\linewidth]{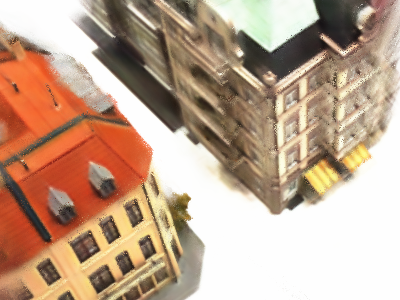}
        \includegraphics[height=0.12\linewidth]{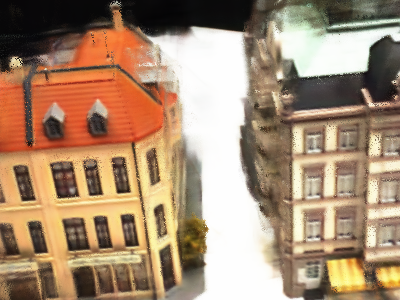}
        \\
        \includegraphics[height=0.12\linewidth]{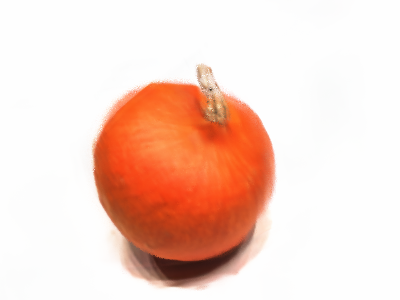}
        \includegraphics[height=0.12\linewidth]{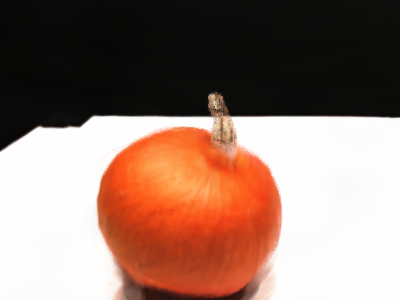}
        \hspace{0.0015\textwidth}
         \includegraphics[height=0.12\linewidth]{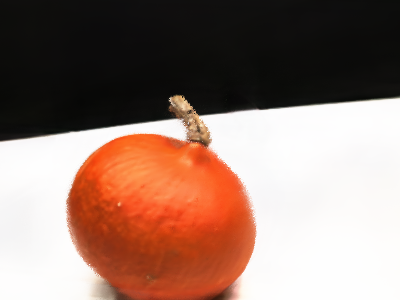}
        \includegraphics[height=0.12\linewidth]{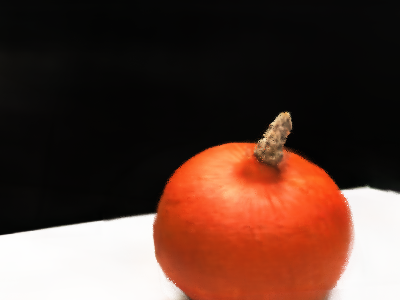}
        \hspace{0.0015\textwidth}
        \includegraphics[height=0.12\linewidth]{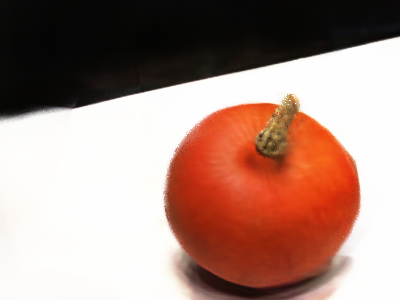}
        \includegraphics[height=0.12\linewidth]{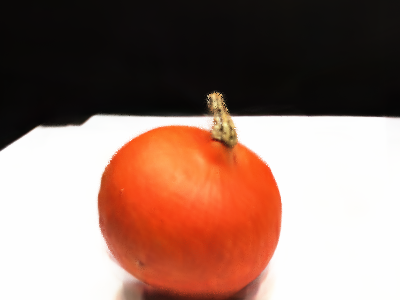}
        \\
        \includegraphics[height=0.12\linewidth]{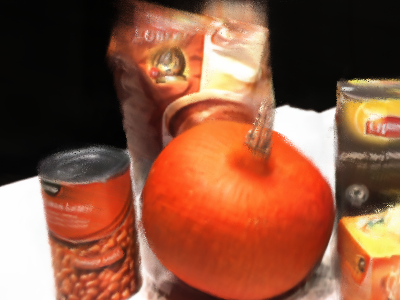}
        \includegraphics[height=0.12\linewidth]{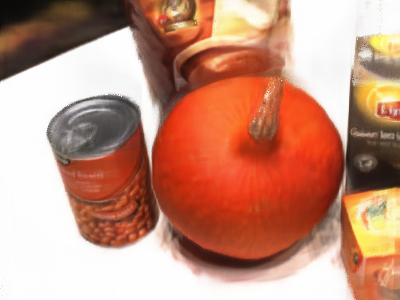}
        \hspace{0.0015\textwidth}
        \includegraphics[height=0.12\linewidth]{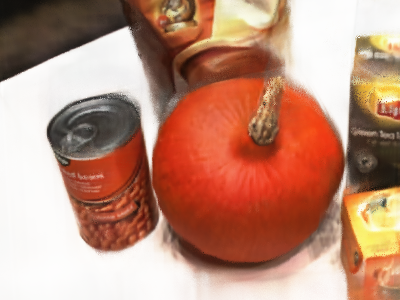}
        \includegraphics[height=0.12\linewidth]{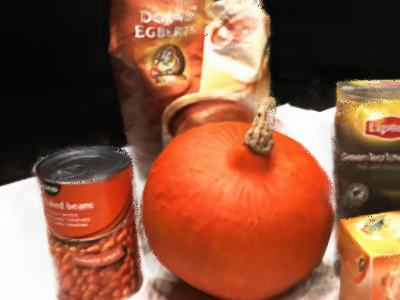}
        \hspace{0.0015\textwidth}
        \includegraphics[height=0.12\linewidth]{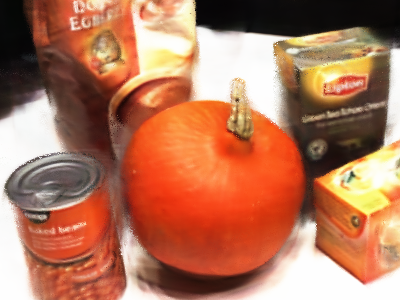}
        \includegraphics[height=0.12\linewidth]{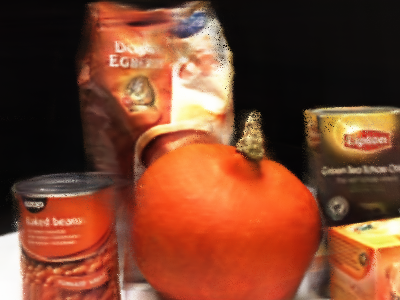}
        \\
        \includegraphics[height=0.12\linewidth]{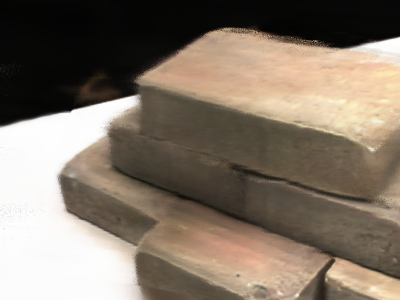}
        \includegraphics[height=0.12\linewidth]{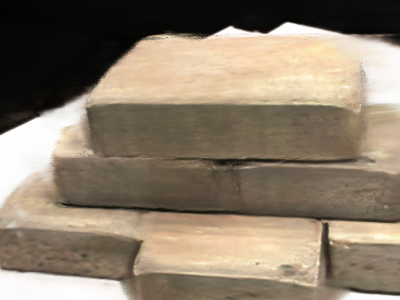}
        \hspace{0.0015\textwidth}
        \includegraphics[height=0.12\linewidth]{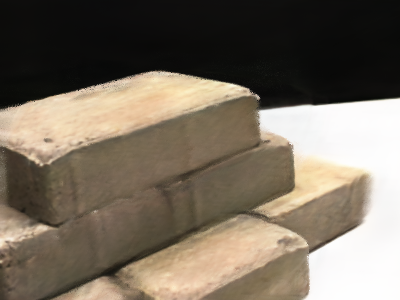}
        \includegraphics[height=0.12\linewidth]{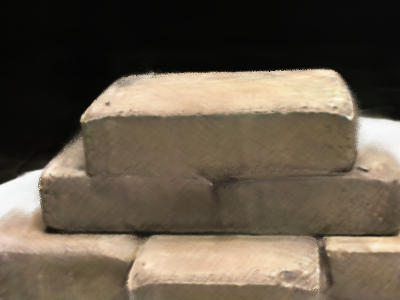}
        \hspace{0.0015\textwidth}
        \includegraphics[height=0.12\linewidth]{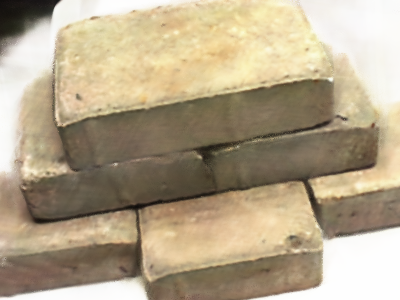}
        \includegraphics[height=0.12\linewidth]{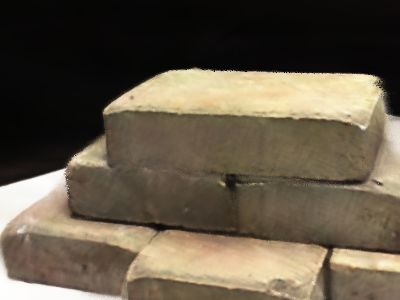}
        \\
        \includegraphics[height=0.12\linewidth]{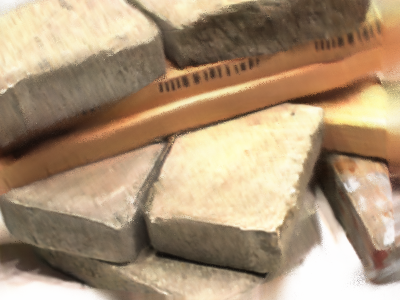}
        \includegraphics[height=0.12\linewidth]{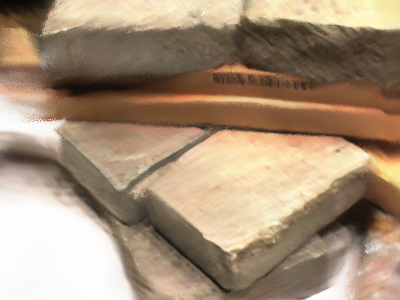}
        \hspace{0.0015\textwidth}
        \includegraphics[height=0.12\linewidth]{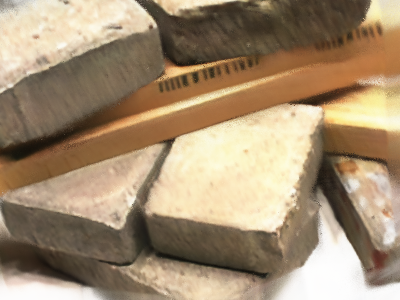}
        \includegraphics[height=0.12\linewidth]{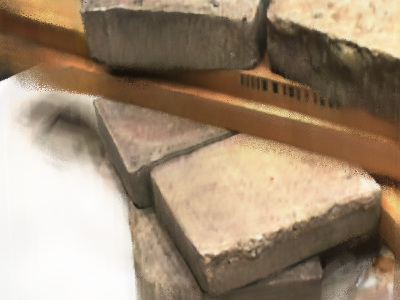}
        \hspace{0.0015\textwidth}
        \includegraphics[height=0.12\linewidth]{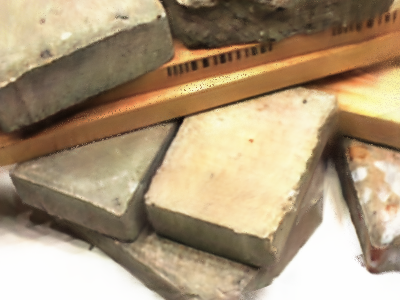}
        \includegraphics[height=0.12\linewidth]{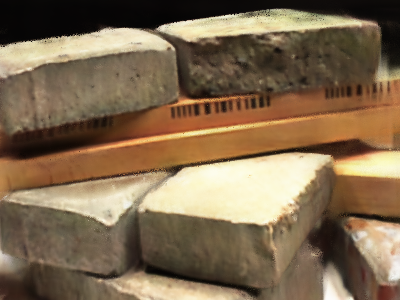}
        \\
        \includegraphics[height=0.12\linewidth]{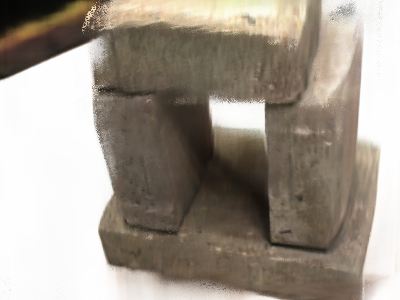}
        \includegraphics[height=0.12\linewidth]{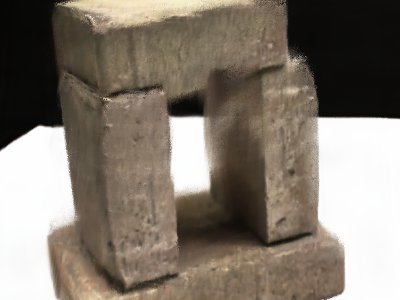}
        \hspace{0.0015\textwidth}
        \includegraphics[height=0.12\linewidth]{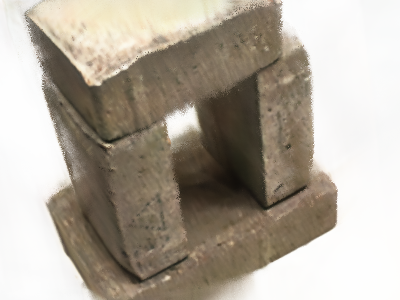}
        \includegraphics[height=0.12\linewidth]{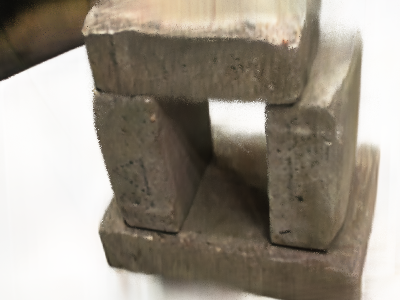}
        \hspace{0.0015\textwidth}
        \includegraphics[height=0.12\linewidth]{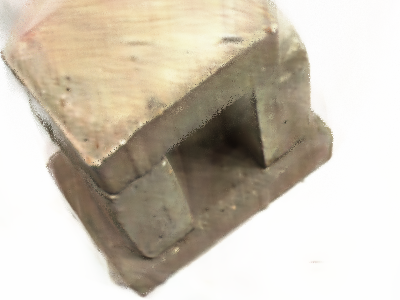}
        \includegraphics[height=0.12\linewidth]{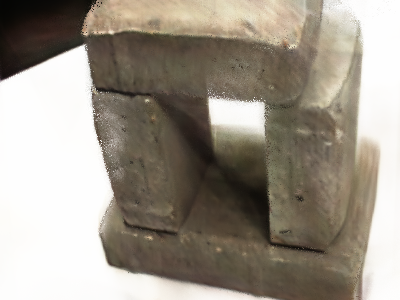}
        \\
        \includegraphics[height=0.12\linewidth]{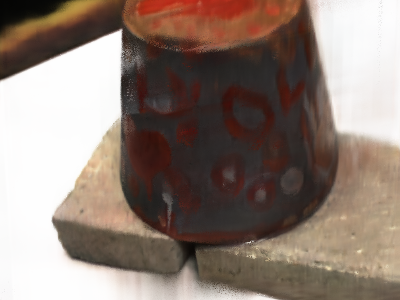}
        \includegraphics[height=0.12\linewidth]{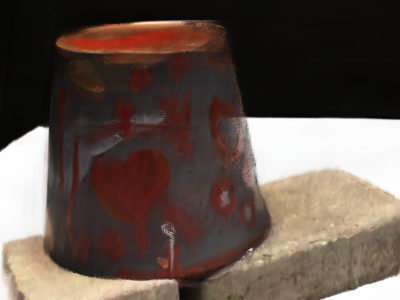}
        \hspace{0.0015\textwidth}
        \includegraphics[height=0.12\linewidth]{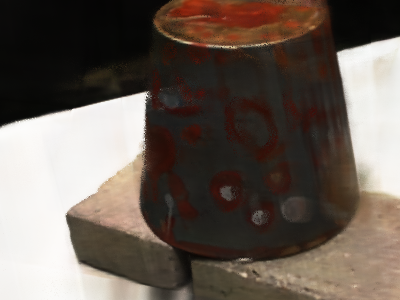}
        \includegraphics[height=0.12\linewidth]{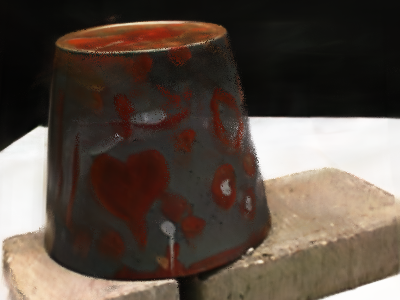}
        \hspace{0.0015\textwidth}
        \includegraphics[height=0.12\linewidth]{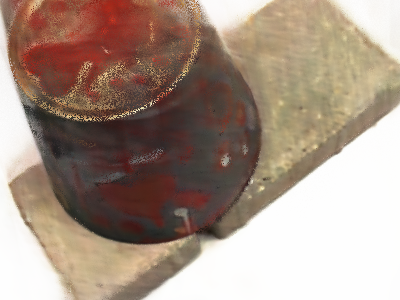}
        \includegraphics[height=0.12\linewidth]{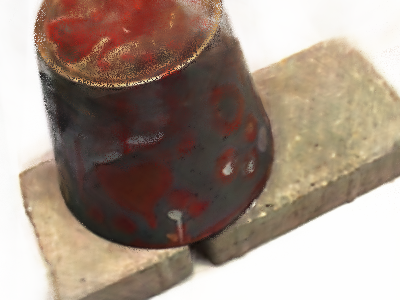}
        \\
	\end{minipage}
    \caption{
    \textbf{Additional Qualitative Results for the First Eight Scenes on DTU.}
    }
    \label{DTU_1}
\end{figure*}

\begin{figure*}[htp] 
    \centering
    \small
    \begin{minipage}{\textwidth}%
        \centering
        \makebox[0.32\textwidth]{3 Views}
        \hspace{0.0015\textwidth}
        \makebox[0.32\textwidth]{6 Views}
        \hspace{0.0015\textwidth}
        \makebox[0.32\textwidth]{9 Views}
        \\
        \includegraphics[height=0.12\linewidth]{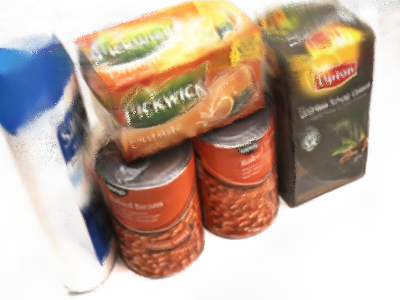}
        \includegraphics[height=0.12\linewidth]{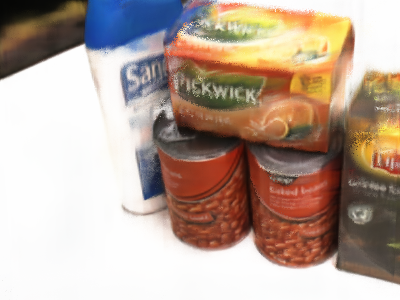}
        \hspace{0.0015\textwidth}
        \includegraphics[height=0.12\linewidth]{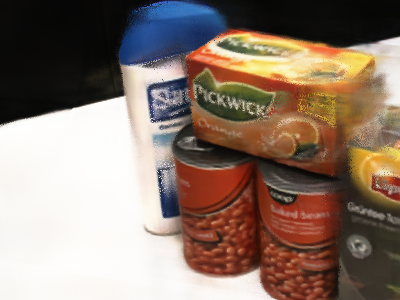}
        \includegraphics[height=0.12\linewidth]{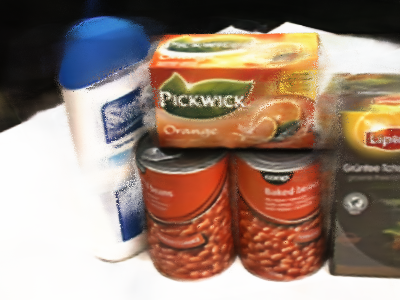}
        \hspace{0.0015\textwidth}
        \includegraphics[height=0.12\linewidth]{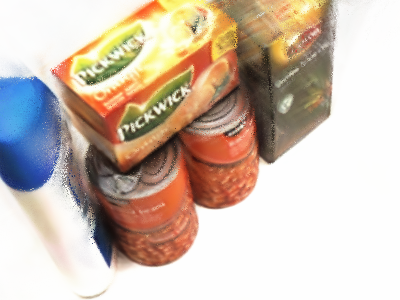}
        \includegraphics[height=0.12\linewidth]{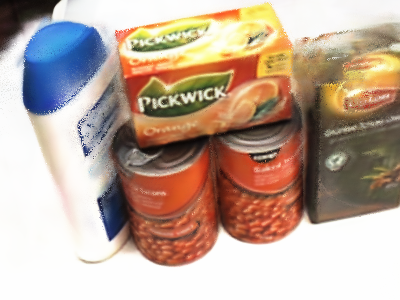}
        \\
        \includegraphics[height=0.12\linewidth]{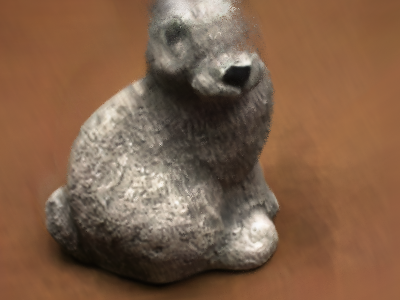}
        \includegraphics[height=0.12\linewidth]{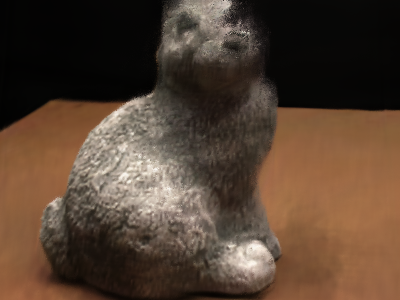}
        \hspace{0.0015\textwidth}
        \includegraphics[height=0.12\linewidth]{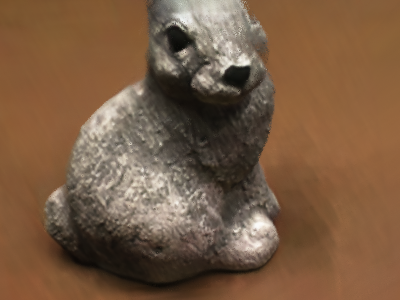}
        \includegraphics[height=0.12\linewidth]{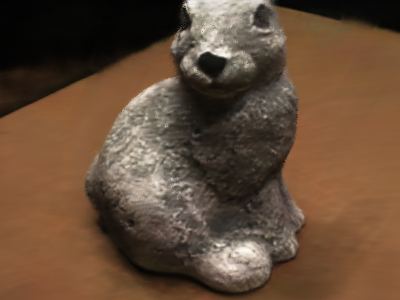}
        \hspace{0.0015\textwidth}
        \includegraphics[height=0.12\linewidth]{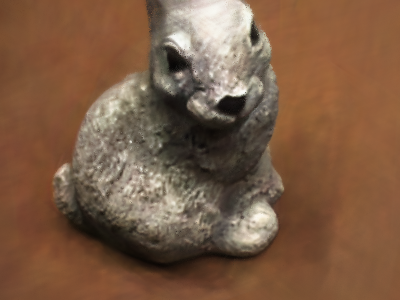}
        \includegraphics[height=0.12\linewidth]{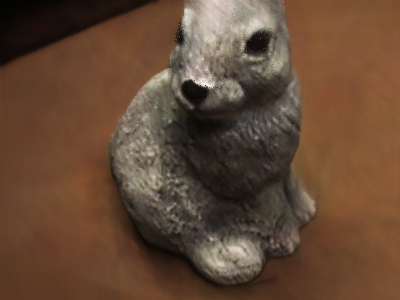}
        \\
        \includegraphics[height=0.12\linewidth]{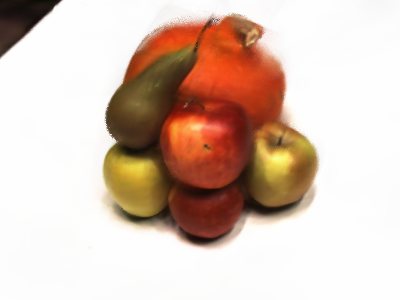}
        \includegraphics[height=0.12\linewidth]{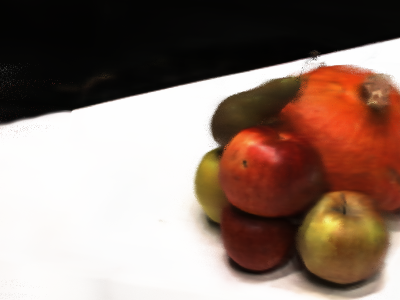}
        \hspace{0.0015\textwidth}
        \includegraphics[height=0.12\linewidth]{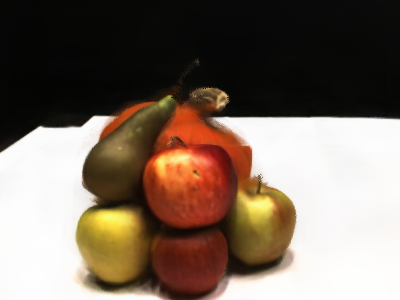}
        \includegraphics[height=0.12\linewidth]{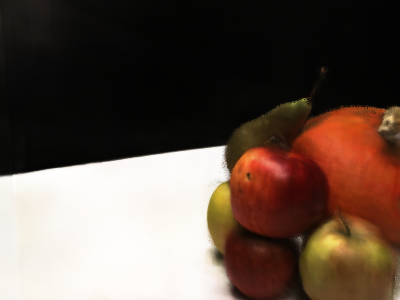}
        \hspace{0.0015\textwidth}
        \includegraphics[height=0.12\linewidth]{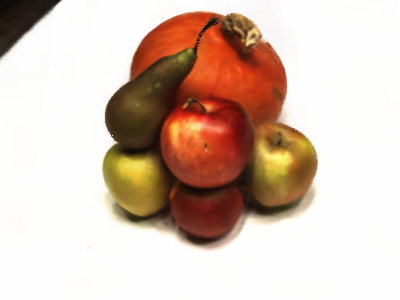}
        \includegraphics[height=0.12\linewidth]{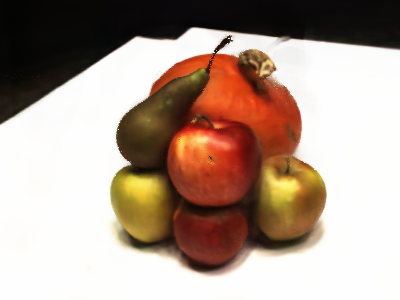}
        \\
        \includegraphics[height=0.12\linewidth]{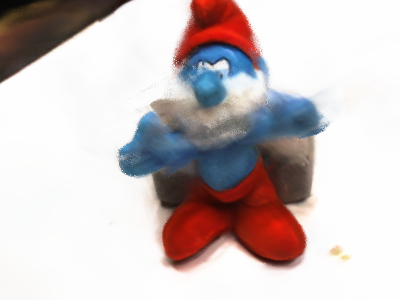}
        \includegraphics[height=0.12\linewidth]{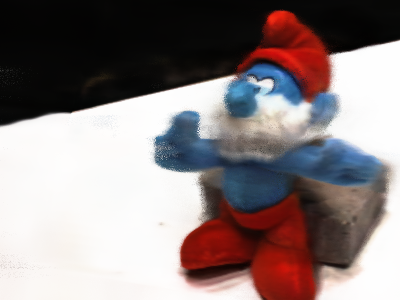}
        \hspace{0.0015\textwidth}
        \includegraphics[height=0.12\linewidth]{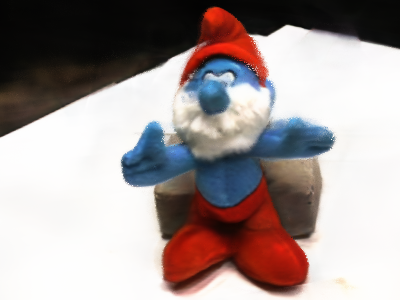}
        \includegraphics[height=0.12\linewidth]{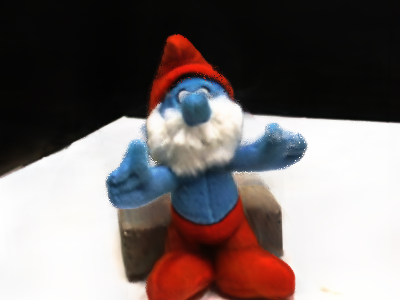}
        \hspace{0.0015\textwidth}
        \includegraphics[height=0.12\linewidth]{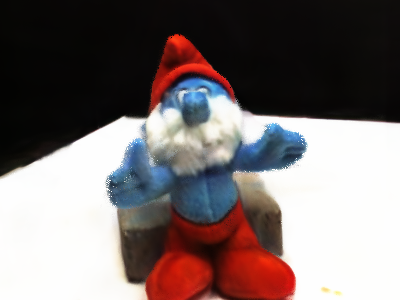}
        \includegraphics[height=0.12\linewidth]{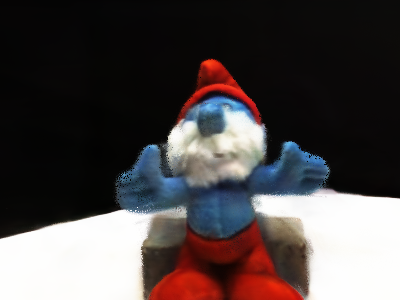}
        \\
        \includegraphics[height=0.12\linewidth]{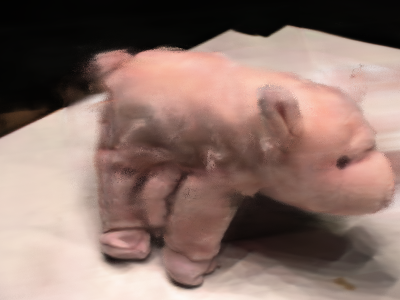}
        \includegraphics[height=0.12\linewidth]{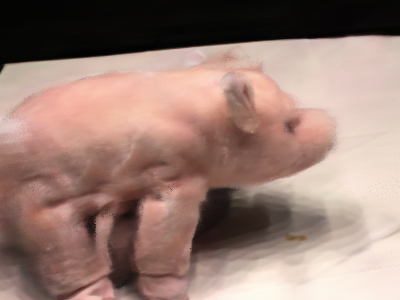}
        \hspace{0.0015\textwidth}
        \includegraphics[height=0.12\linewidth]{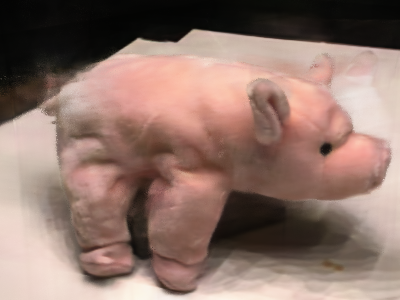}
        \includegraphics[height=0.12\linewidth]{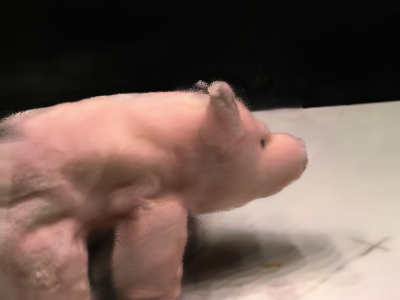}
        \hspace{0.0015\textwidth}
        \includegraphics[height=0.12\linewidth]{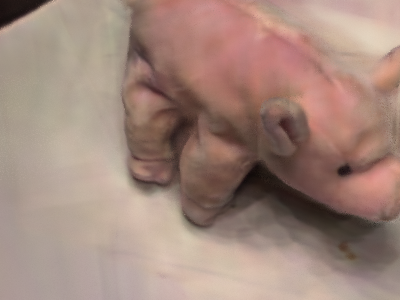}
        \includegraphics[height=0.12\linewidth]{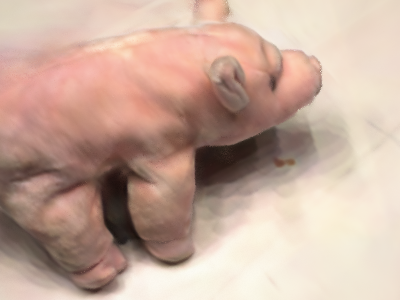}
        \\
        \includegraphics[height=0.12\linewidth]{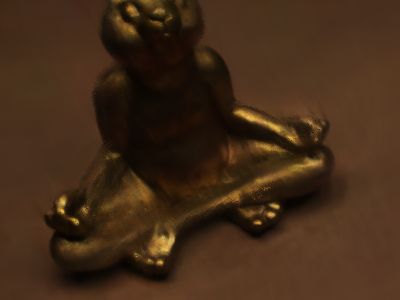}
        \includegraphics[height=0.12\linewidth]{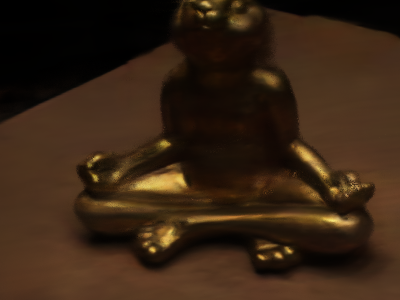}
        \hspace{0.0015\textwidth}
        \includegraphics[height=0.12\linewidth]{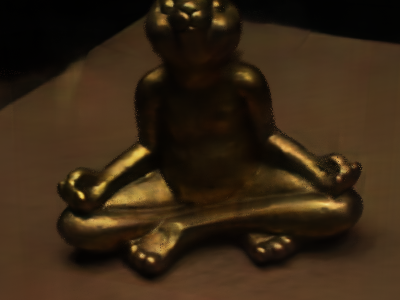}
        \includegraphics[height=0.12\linewidth]{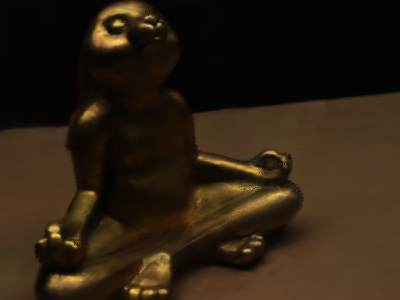}
        \hspace{0.0015\textwidth}
        \includegraphics[height=0.12\linewidth]{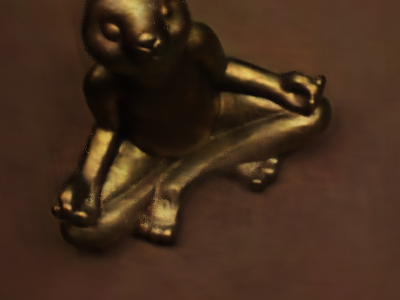}
        \includegraphics[height=0.12\linewidth]{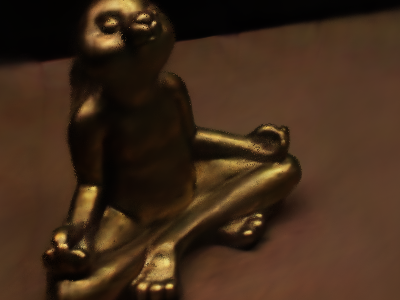}
        \\
        \includegraphics[height=0.12\linewidth]{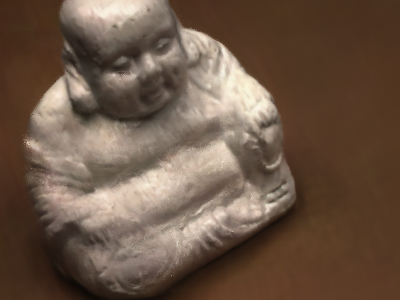}
        \includegraphics[height=0.12\linewidth]{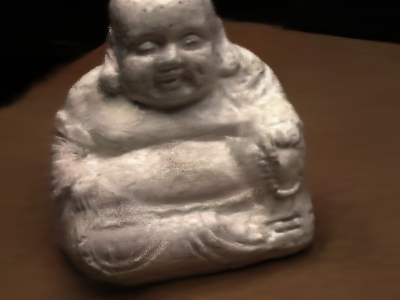}
        \hspace{0.0015\textwidth}
        \includegraphics[height=0.12\linewidth]{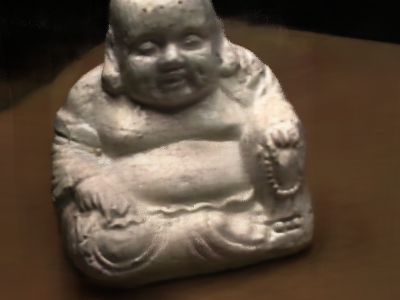}
        \includegraphics[height=0.12\linewidth]{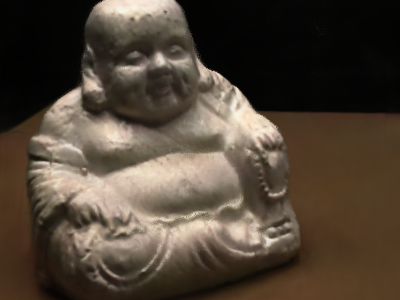}
        \hspace{0.0015\textwidth}
        \includegraphics[height=0.12\linewidth]{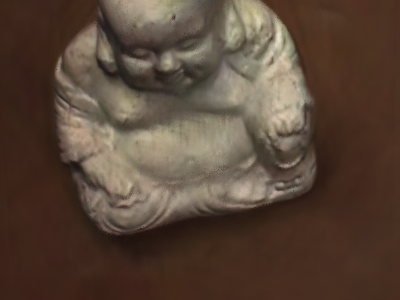}
        \includegraphics[height=0.12\linewidth]{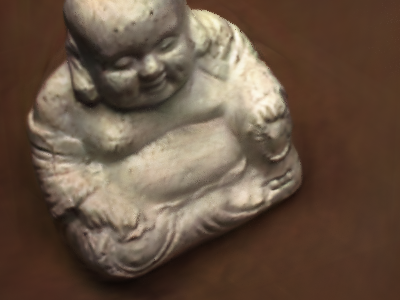}
        \\
        \end{minipage}
    \caption{
    \textbf{Additional Qualitative Results for the Last Seven Scenes on DTU.}
    }
    \label{DTU_2}
\end{figure*}

\begin{figure*}[htp] 
    \centering
    \small
    \begin{minipage}{\textwidth}%
        \centering
        \raisebox{3em}{
          \begin{minipage}{0.04\textwidth}
            RGB
          \end{minipage}
        }
        \includegraphics[height=0.115\linewidth]{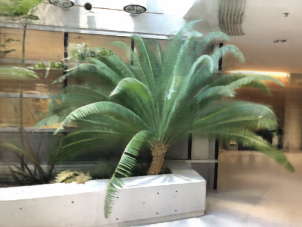}
        \includegraphics[height=0.115\linewidth]{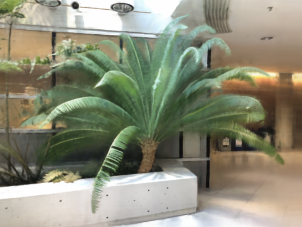}
        \includegraphics[height=0.115\linewidth]{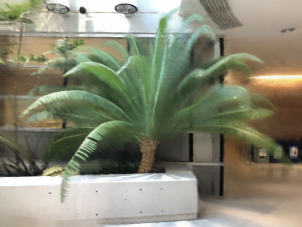}
        \includegraphics[height=0.115\linewidth]{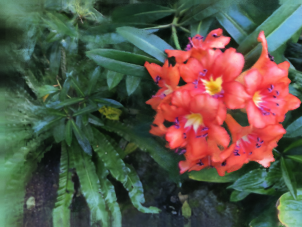}
        \includegraphics[height=0.115\linewidth]{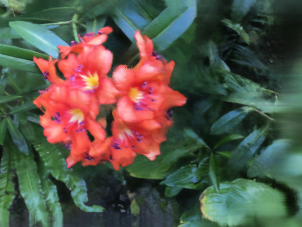}
        \includegraphics[height=0.115\linewidth]{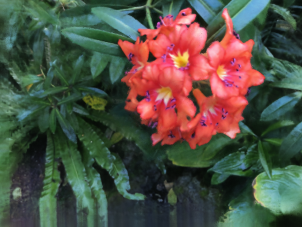}
        \\
        \raisebox{3em}{
          \begin{minipage}{0.0426\textwidth}
            Depth
          \end{minipage}
        }
        \includegraphics[height=0.115\linewidth]{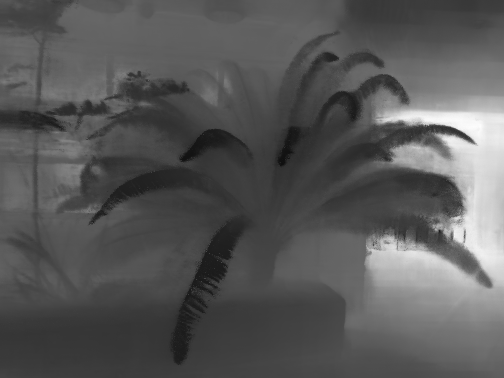}
        \includegraphics[height=0.115\linewidth]{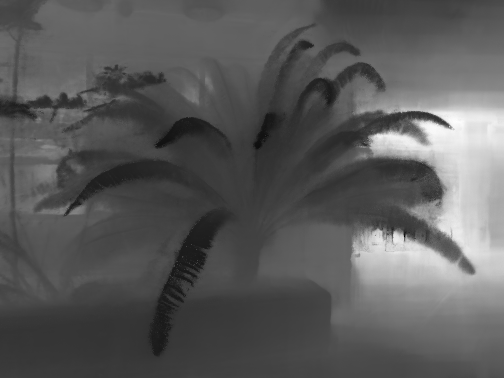}
        \includegraphics[height=0.115\linewidth]{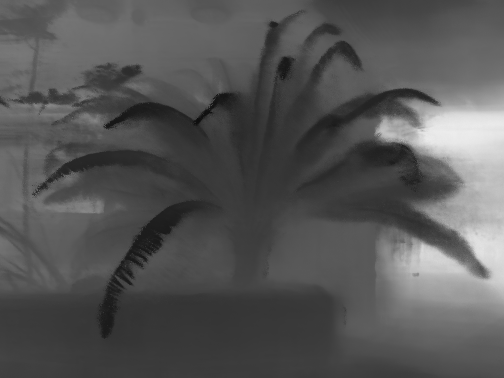}
        \includegraphics[height=0.115\linewidth]{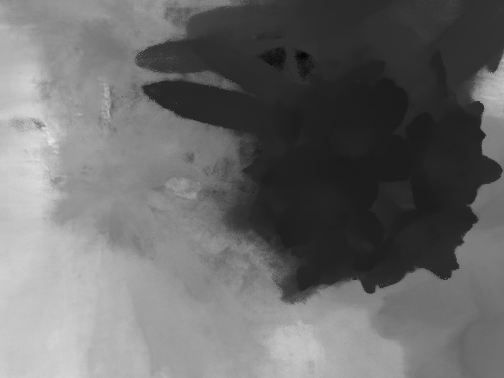}
        \includegraphics[height=0.115\linewidth]{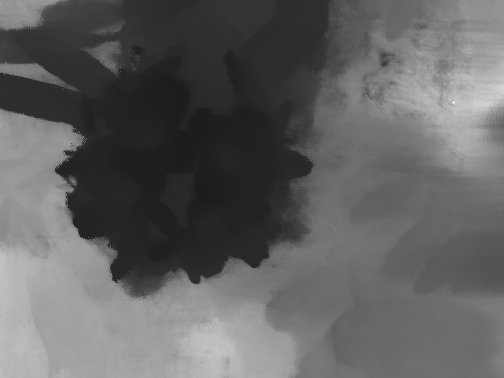}
        \includegraphics[height=0.115\linewidth]{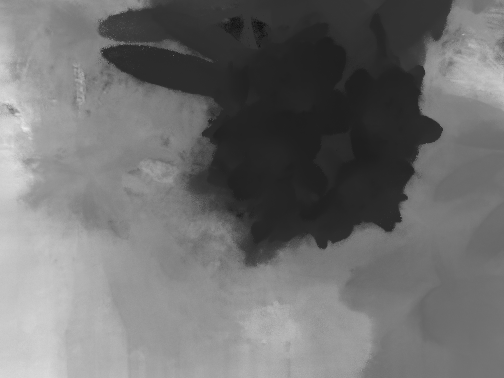}
        \vspace{1em}
        \\
        \raisebox{3em}{
          \begin{minipage}{0.04\textwidth}
            RGB
          \end{minipage}
        }
        \includegraphics[height=0.115\linewidth]{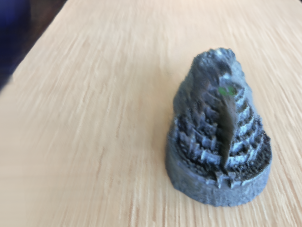}
        \includegraphics[height=0.115\linewidth]{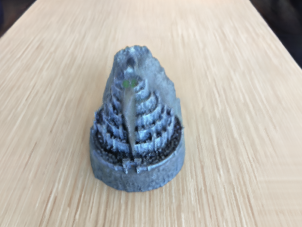}
        \includegraphics[height=0.115\linewidth]{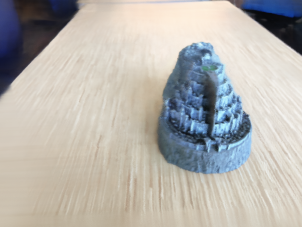}
        \includegraphics[height=0.115\linewidth]{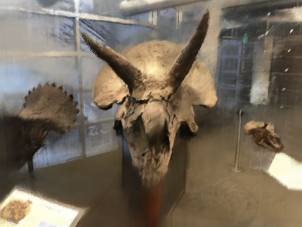}
        \includegraphics[height=0.115\linewidth]{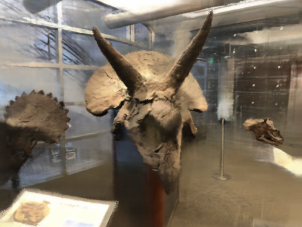}
        \includegraphics[height=0.115\linewidth]{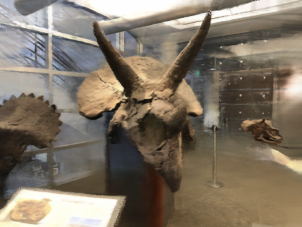}
        \\
        \raisebox{3em}{
          \begin{minipage}{0.0426\textwidth}
            Depth
          \end{minipage}
        }
        \includegraphics[height=0.115\linewidth]{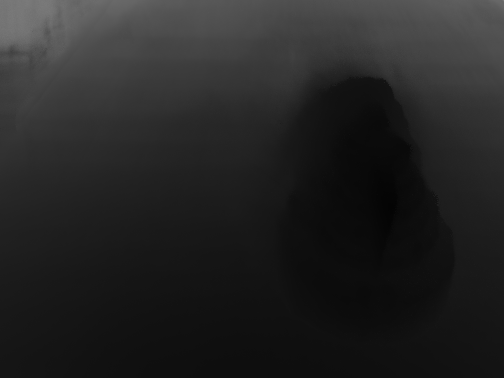}
        \includegraphics[height=0.115\linewidth]{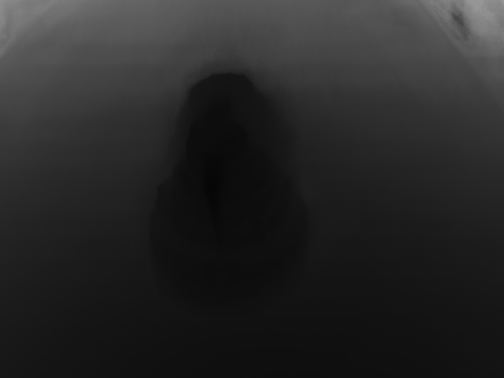}
        \includegraphics[height=0.115\linewidth]{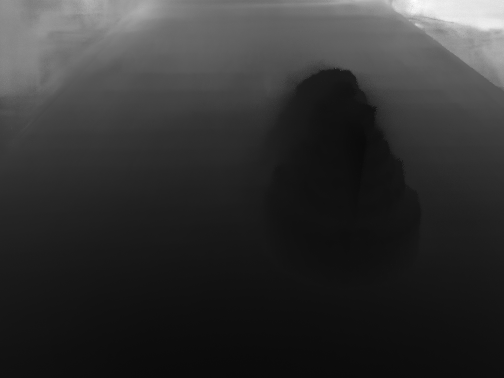}
        \includegraphics[height=0.115\linewidth]{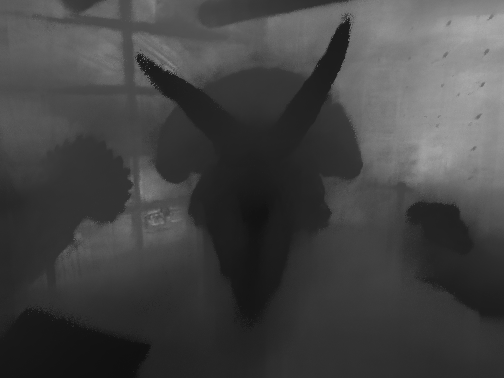}
        \includegraphics[height=0.115\linewidth]{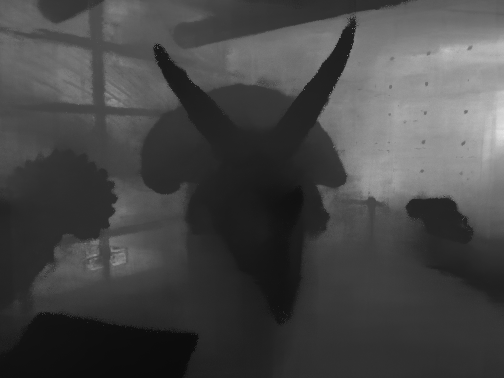}
        \includegraphics[height=0.115\linewidth]{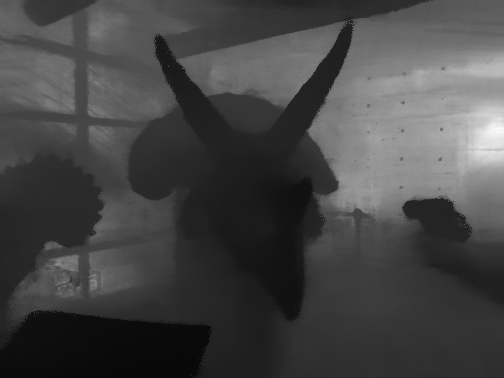}
        \vspace{1em}
        \\
        \raisebox{3em}{
          \begin{minipage}{0.04\textwidth}
            RGB
          \end{minipage}
        }
        \includegraphics[height=0.115\linewidth]{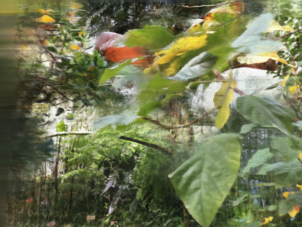}
        \includegraphics[height=0.115\linewidth]{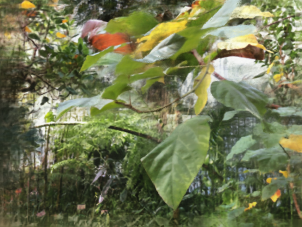}
        \includegraphics[height=0.115\linewidth]{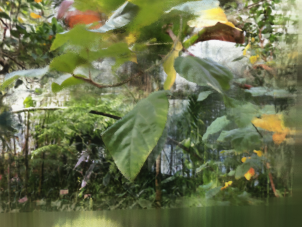}
        \includegraphics[height=0.115\linewidth]{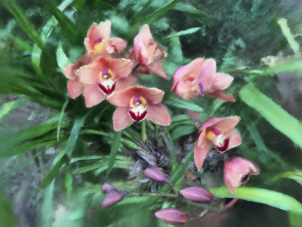}
        \includegraphics[height=0.115\linewidth]{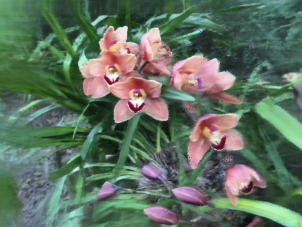}
        \includegraphics[height=0.115\linewidth]{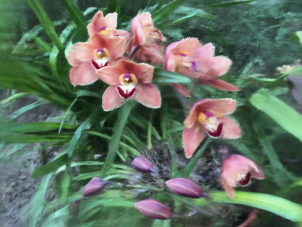}
        \\
        \raisebox{3em}{
          \begin{minipage}{0.0426\textwidth}
            Depth
          \end{minipage}
        }
        \includegraphics[height=0.115\linewidth]{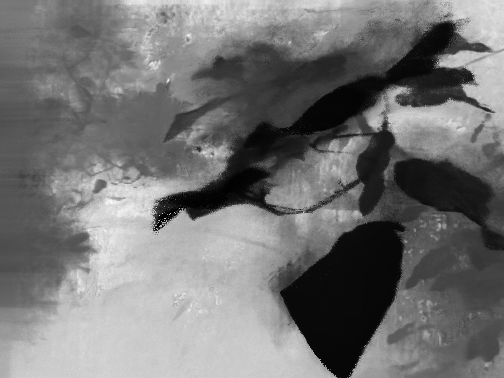}
        \includegraphics[height=0.115\linewidth]{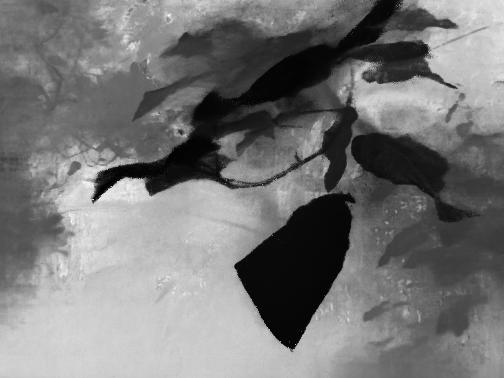}
        \includegraphics[height=0.115\linewidth]{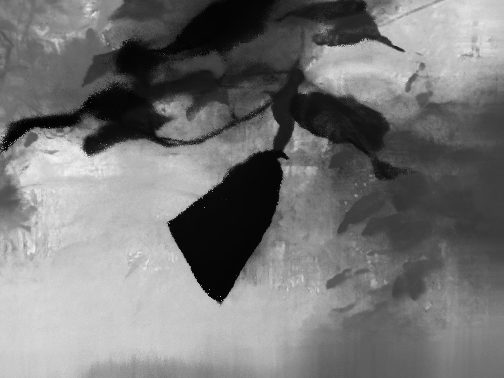}
        \includegraphics[height=0.115\linewidth]{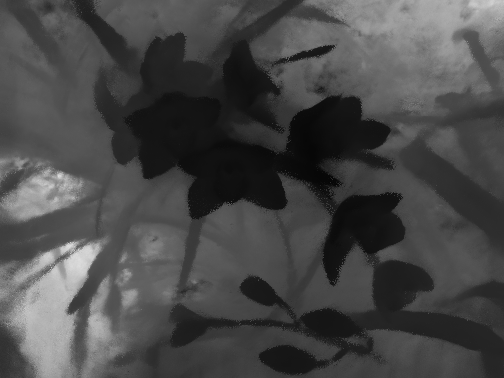}
        \includegraphics[height=0.115\linewidth]{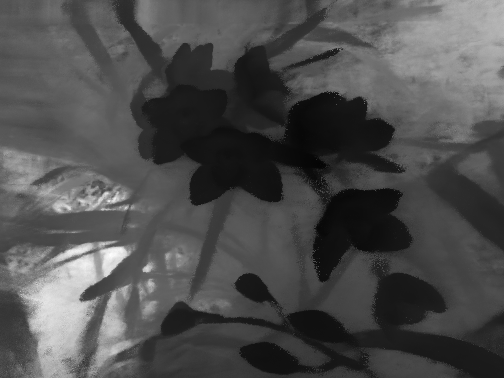}
        \includegraphics[height=0.115\linewidth]{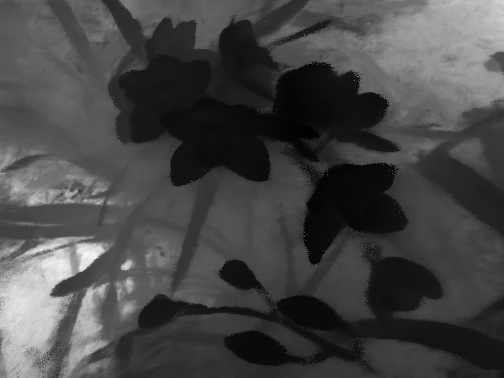}
        \vspace{1em}
        \\
        \raisebox{3em}{
          \begin{minipage}{0.04\textwidth}
            RGB
          \end{minipage}
        }
        \includegraphics[height=0.115\linewidth]{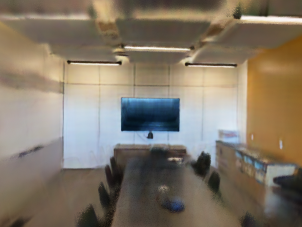}
        \includegraphics[height=0.115\linewidth]{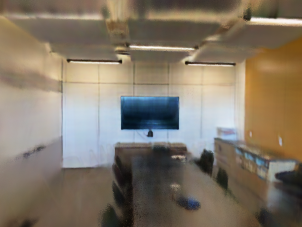}
        \includegraphics[height=0.115\linewidth]{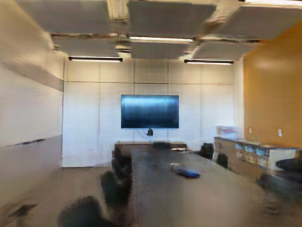}
        \includegraphics[height=0.115\linewidth]{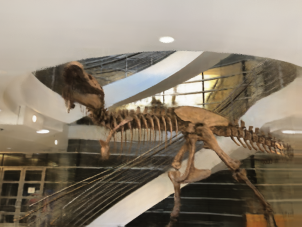}
        \includegraphics[height=0.115\linewidth]{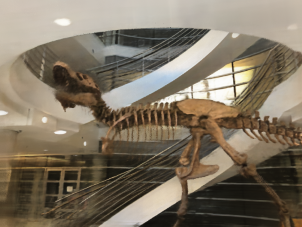}
        \includegraphics[height=0.115\linewidth]{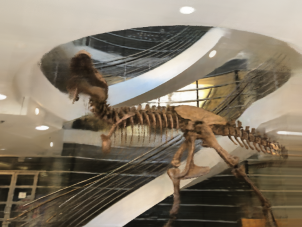}
        \\
        \raisebox{3em}{
          \begin{minipage}{0.04\textwidth}
            Depth
          \end{minipage}
        }
        \includegraphics[height=0.115\linewidth]{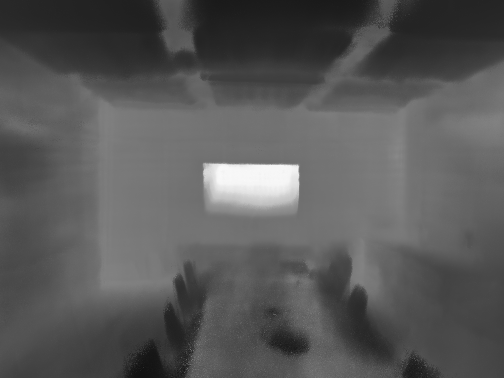}
        \includegraphics[height=0.115\linewidth]{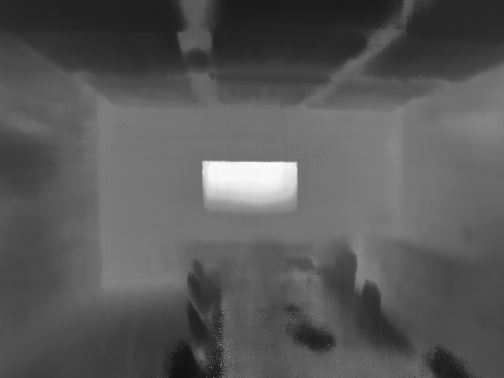}
        \includegraphics[height=0.115\linewidth]{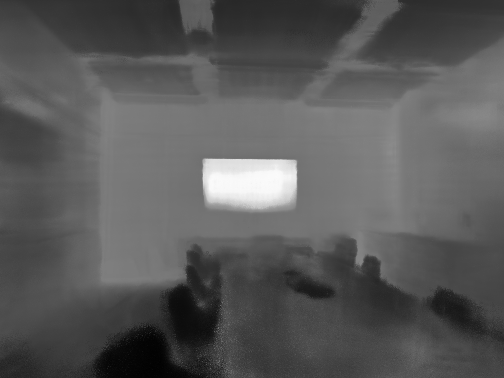}
        \includegraphics[height=0.115\linewidth]{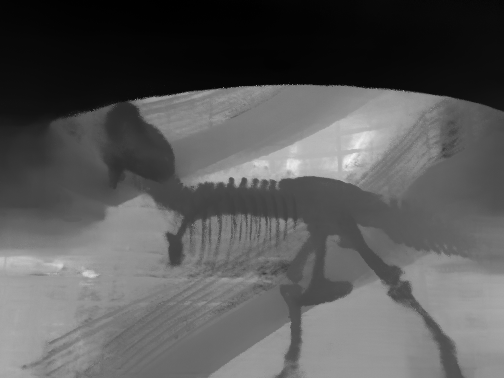}
        \includegraphics[height=0.115\linewidth]{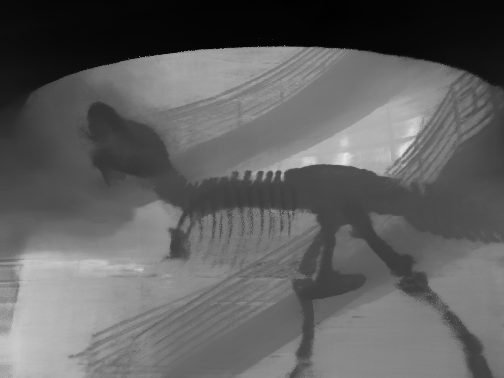}
        \includegraphics[height=0.115\linewidth]{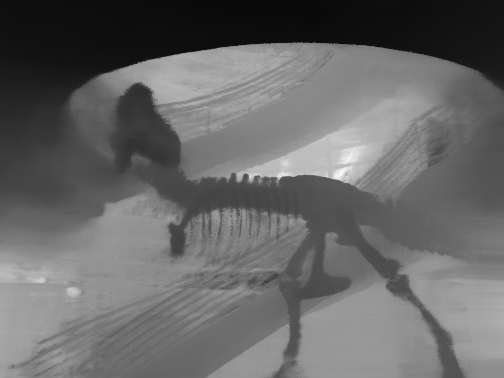}

    \end{minipage}
    \caption{
    \textbf{Additional Qualitative Results and Rendered Depth on LLFF for 3 Input Views.}
    }
    \label{LLFF_3V}
\end{figure*}

\begin{figure*}[htp] 
    \centering
    \small
    \begin{minipage}{\textwidth}%
        \centering
        \raisebox{3em}{
          \begin{minipage}{0.04\textwidth}
            RGB
          \end{minipage}
        }
        \includegraphics[height=0.115\linewidth]{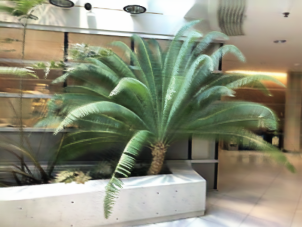}
        \includegraphics[height=0.115\linewidth]{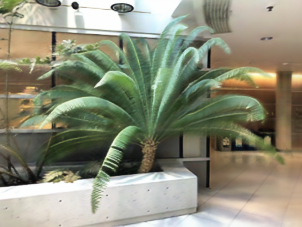}
        \includegraphics[height=0.115\linewidth]{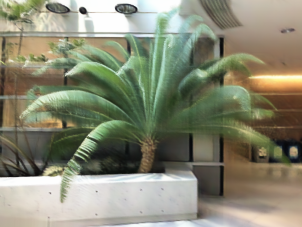}
        \includegraphics[height=0.115\linewidth]{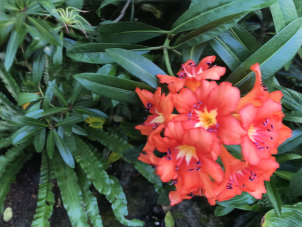}
        \includegraphics[height=0.115\linewidth]{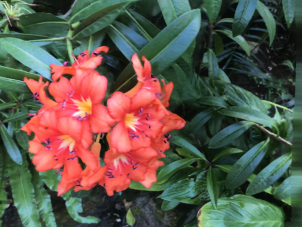}
        \includegraphics[height=0.115\linewidth]{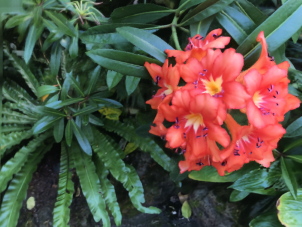}
        \\
        \raisebox{3em}{
          \begin{minipage}{0.0426\textwidth}
            Depth
          \end{minipage}
        }
        \includegraphics[height=0.115\linewidth]{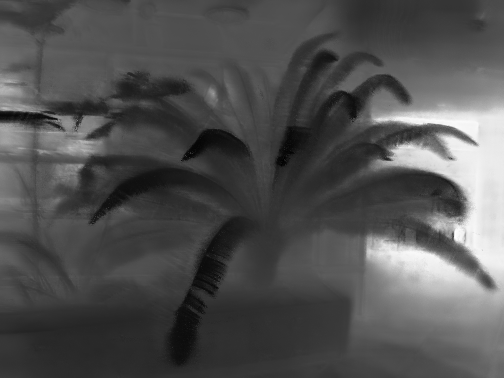}
        \includegraphics[height=0.115\linewidth]{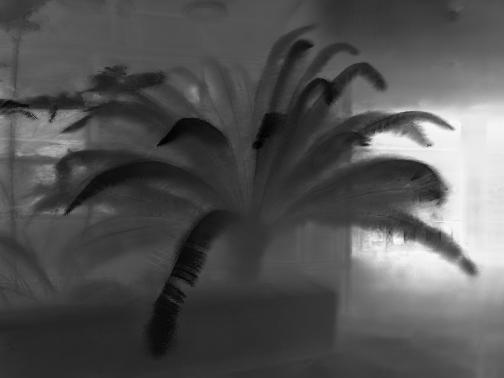}
        \includegraphics[height=0.115\linewidth]{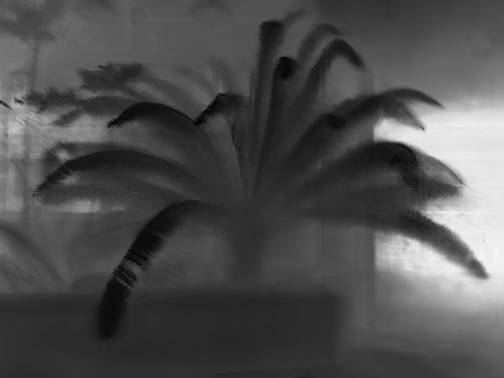}
        \includegraphics[height=0.115\linewidth]{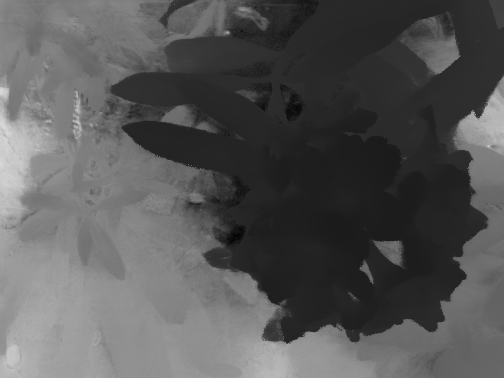}
        \includegraphics[height=0.115\linewidth]{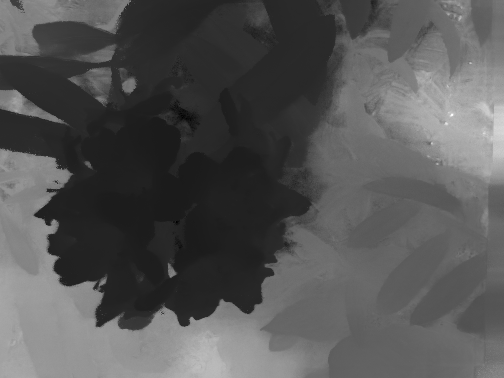}
        \includegraphics[height=0.115\linewidth]{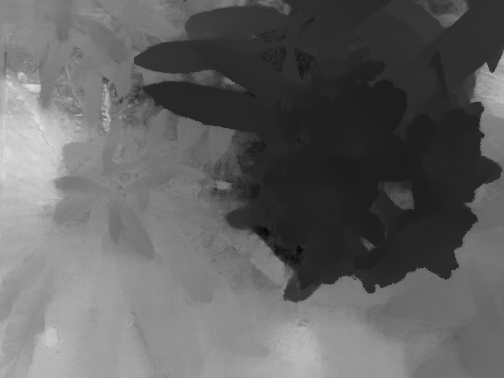}
        \vspace{1em}
        \\
        \raisebox{3em}{
          \begin{minipage}{0.04\textwidth}
            RGB
          \end{minipage}
        }
        \includegraphics[height=0.115\linewidth]{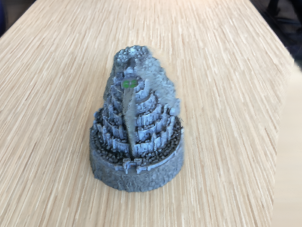}
        \includegraphics[height=0.115\linewidth]{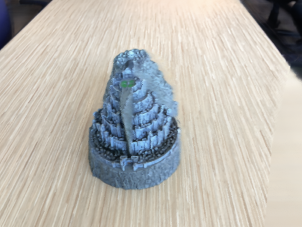}
        \includegraphics[height=0.115\linewidth]{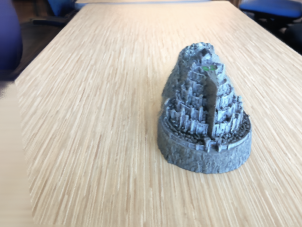}
        \includegraphics[height=0.115\linewidth]{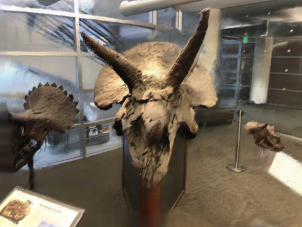}
        \includegraphics[height=0.115\linewidth]{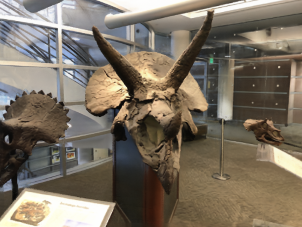}
        \includegraphics[height=0.115\linewidth]{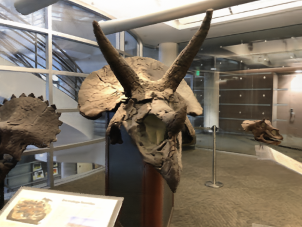}
        \\
        \raisebox{3em}{
          \begin{minipage}{0.0426\textwidth}
            Depth
          \end{minipage}
        }
        \includegraphics[height=0.115\linewidth]{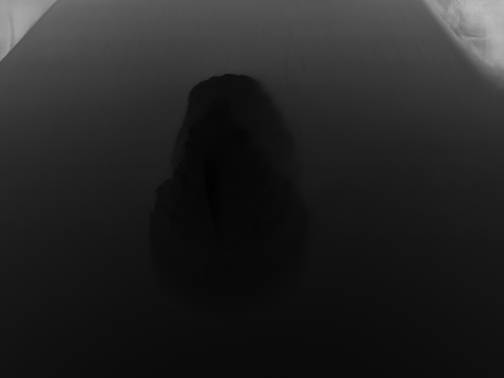}
        \includegraphics[height=0.115\linewidth]{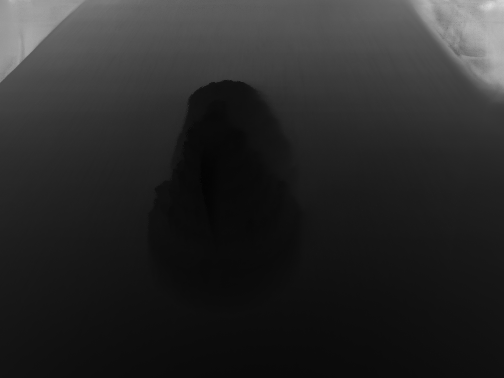}
        \includegraphics[height=0.115\linewidth]{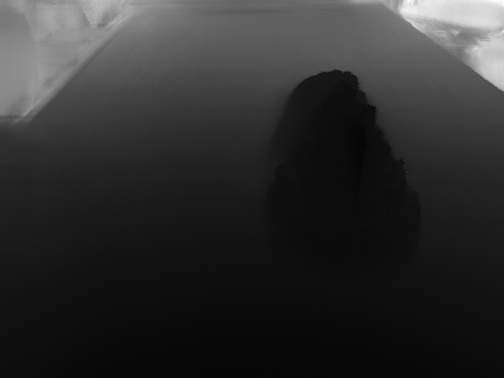}
        \includegraphics[height=0.115\linewidth]{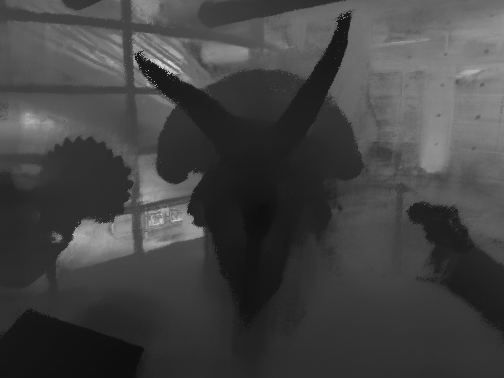}
        \includegraphics[height=0.115\linewidth]{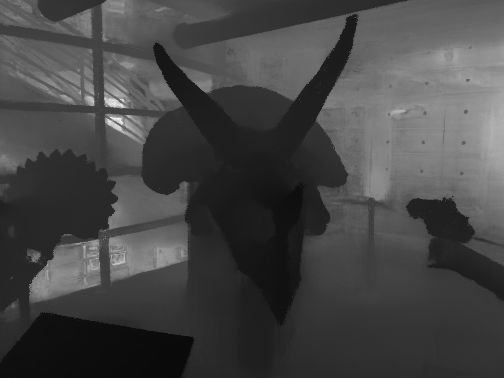}
        \includegraphics[height=0.115\linewidth]{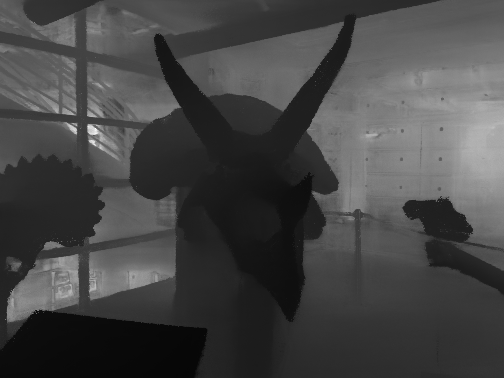}
        \vspace{1em}
        \\
        \raisebox{3em}{
          \begin{minipage}{0.04\textwidth}
            RGB
          \end{minipage}
        }
        \includegraphics[height=0.115\linewidth]{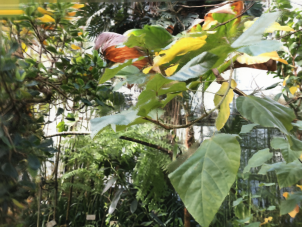}
        \includegraphics[height=0.115\linewidth]{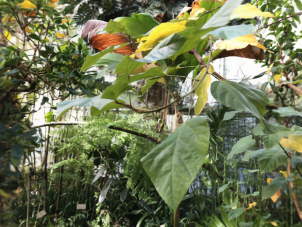}
        \includegraphics[height=0.115\linewidth]{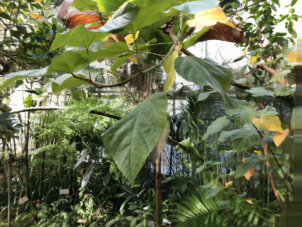}
        \includegraphics[height=0.115\linewidth]{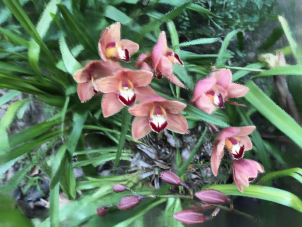}
        \includegraphics[height=0.115\linewidth]{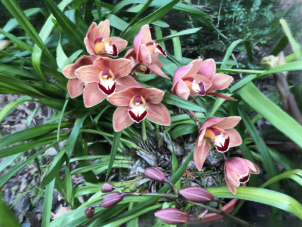}
        \includegraphics[height=0.115\linewidth]{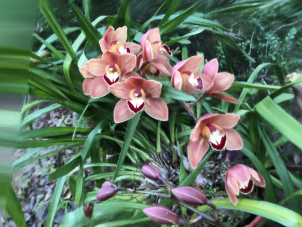}
        \\
        \raisebox{3em}{
          \begin{minipage}{0.0426\textwidth}
            Depth
          \end{minipage}
        }
        \includegraphics[height=0.115\linewidth]{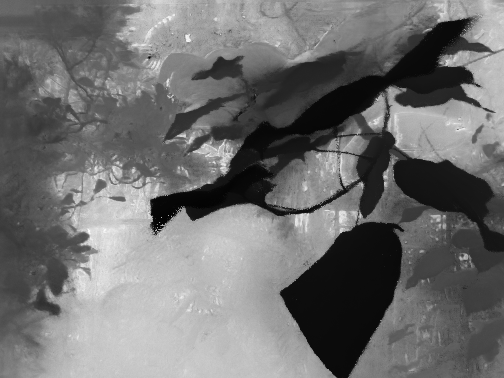}
        \includegraphics[height=0.115\linewidth]{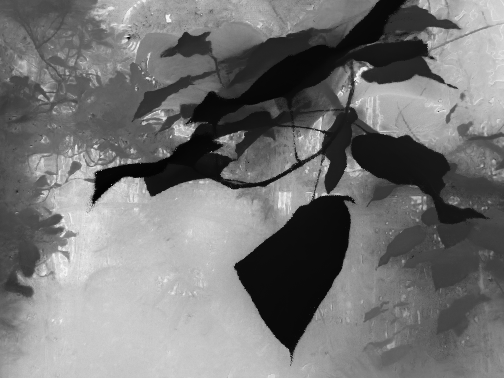}
        \includegraphics[height=0.115\linewidth]{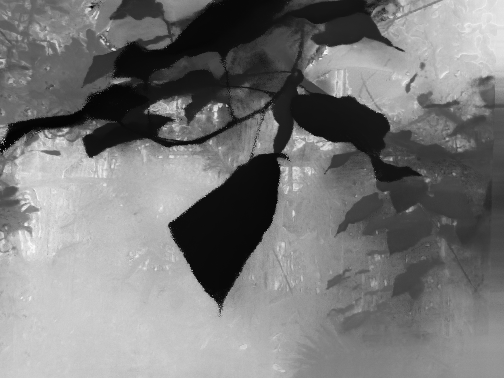}
        \includegraphics[height=0.115\linewidth]{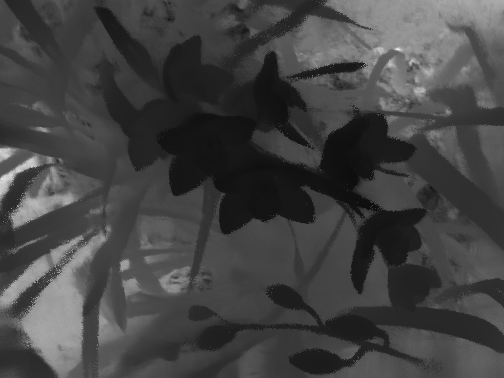}
        \includegraphics[height=0.115\linewidth]{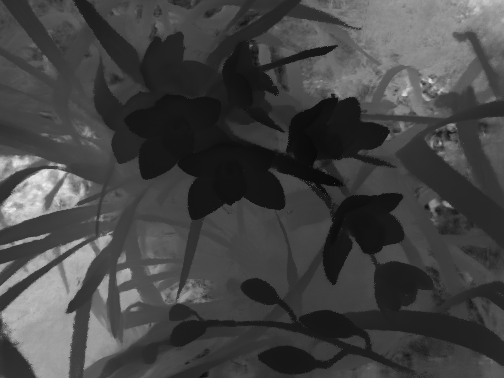}
        \includegraphics[height=0.115\linewidth]{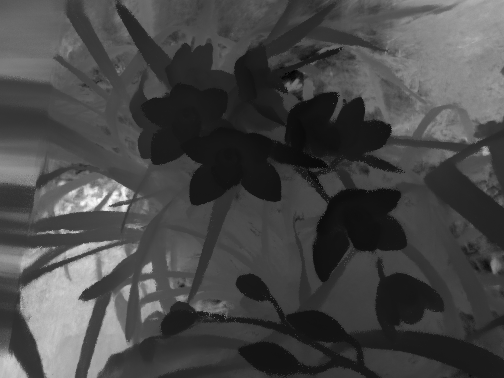}
        \vspace{1em}
        \\
        \raisebox{3em}{
          \begin{minipage}{0.04\textwidth}
            RGB
          \end{minipage}
        }
        \includegraphics[height=0.115\linewidth]{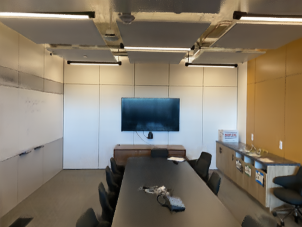}
        \includegraphics[height=0.115\linewidth]{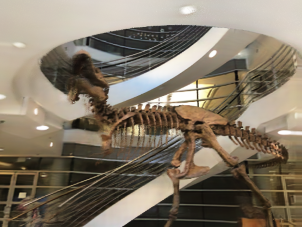}
        \includegraphics[height=0.115\linewidth]{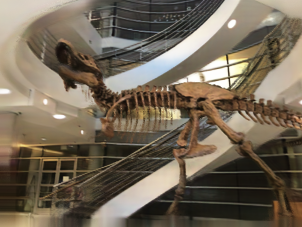}
        \includegraphics[height=0.115\linewidth]{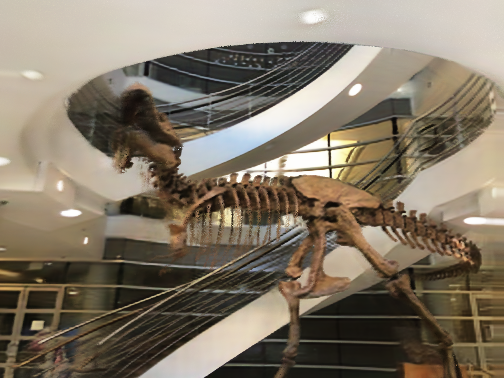}
        \includegraphics[height=0.115\linewidth]{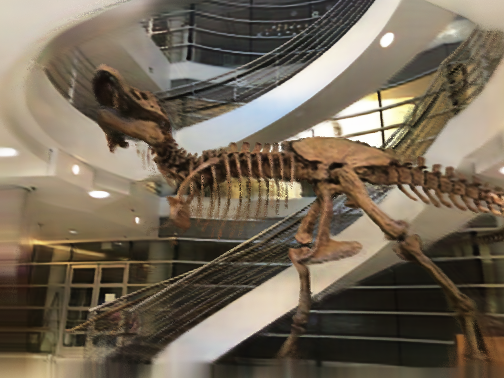}
        \includegraphics[height=0.115\linewidth]{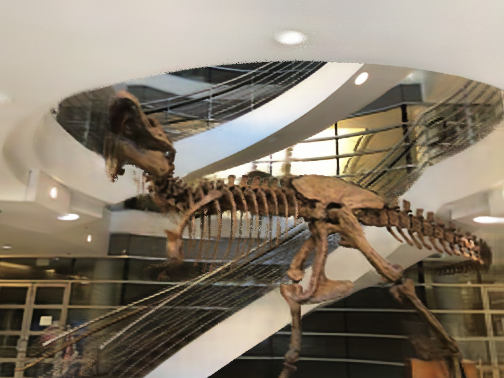}
        \\
        \raisebox{3em}{
          \begin{minipage}{0.04\textwidth}
            Depth
          \end{minipage}
        }
        \includegraphics[height=0.115\linewidth]{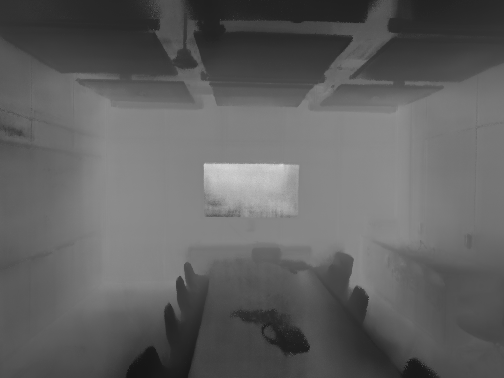}
        \includegraphics[height=0.115\linewidth]{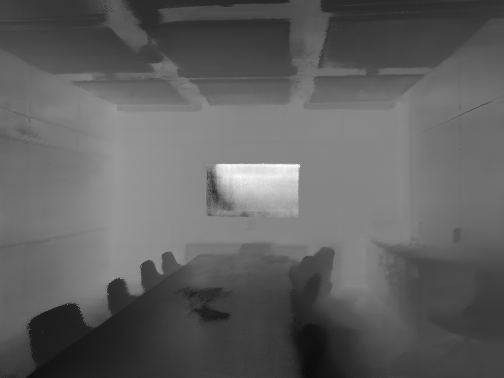}
        \includegraphics[height=0.115\linewidth]{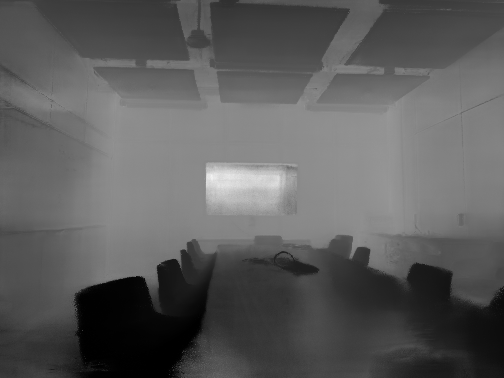}
        \includegraphics[height=0.115\linewidth]{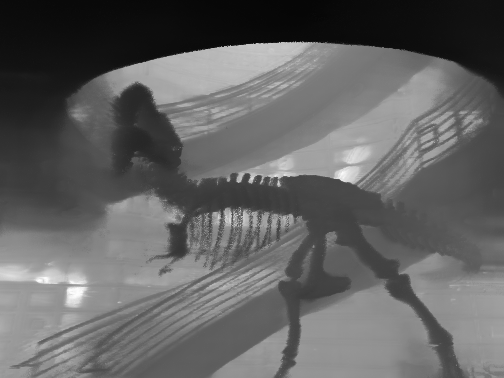}
        \includegraphics[height=0.115\linewidth]{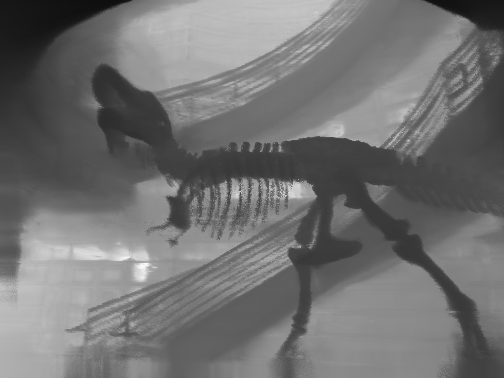}
        \includegraphics[height=0.115\linewidth]{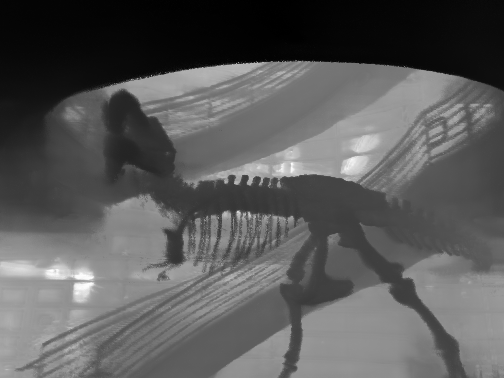}

    \end{minipage}
    \caption{
    \textbf{Additional Qualitative Results and Rendered Depth on LLFF for 6 Input Views.}
    }
    \label{LLFF_6V}
\end{figure*}

\section{B. Additional Results}
\subsection{B.1. Quantitative Results}
\label{sec_method}

\subsubsection{Trained on LLFF} 
In the main paper, the result presented under the LLFF dataset serves as an out-of-distribution validation for our pre-trained DTU model.
It is might unfair to compare our method directly with scene-specific training models such as FreeNeRF~\cite{yang2023freenerf} and RegNeRF~\cite{niemeyer2022regnerf}.
Thus, to demonstrate that our model can achieve comparable results even with identical training datasets, we conduct an additional experiment, ``Trained on LLFF and Evaluated on LLFF''.
It is noted that, our model is still not trained on a single scene, which means that our LLFF-trained model still retains its generalizability.
We use an experimental setup similar to the ``Trained on DTU'' scenario. 
In each optimization iteration, we randomly select a scene and launch 128 rays in random directions for training. 
Cross-scene training can provide scene priors for our model. 
However, considering that the 8 scenes in the LLFF dataset have significant differences in the number of viewpoints, distances and scene distribution, we still perform a brief fine-tuning of 5K iterations for each scene before testing.
The final test results for the 3-view and 6-view scenarios are presented in Table~\ref{Tab 1}.


\subsection{B.2. Qualitative Results}
\label{sec_method}
\subsubsection{Additional Comparisons on DTU}
We present additional rendering results from our ColNeRF model in comparison with other methods under 3 and 6 input view settings in Figure~\ref{DTU_3V_comp} and Figure~\ref{DTU_6V_comp}.
Methods like RegNeRF, which integrate scene-specific optimization, often show visual degradation, \textit{e.g.} floater and noise in the spatial representation.
One reason for this phenomenon is mainly attributed to the strong reliance of these methods' networks on coordinates, leading to deviations in coordinate mapping and consequently introducing noise and incorrect rendering in certain spatial regions of the scene. 
This kind of deficiency tends to be prevalent in specific localized spatial areas.
In contrast, our approach excels in preserving the overall structural integrity of the scene.

\subsubsection{Additional Comparisons on LLFF}
We present additional visualization results demonstrating the effective generalizability of our model, which is trained on the DTU dataset to the LLFF dataset. The additional qualitative comparison results for 3-views input are shown in Fig.~\ref{LLFF_3V_comp}.

\subsubsection{Depth Estimation Comparison}
We further provide a comparison of the depth maps predicted by our method and DS-NeRF,
to demonstrate our method is capable to reconstruct the geometric structure of the scene.
The main innovation of DS-NeRF lies in its use of additional depth labels to supervise training of network.
As can be seen from Fig.~\ref{LLFF_depth}, our method can still accurately predict the geometric structure of the scene even without the need for additional supervision.

\subsubsection{More Rendered results} 
To more comprehensively demonstrate the effectiveness of our model, we present more rendering results. Fig.~\ref{DTU_1} and Fig.~\ref{DTU_2} show the rendering results of our model on DTU with 3-view input, 6-view input and 9-view input, while Fig.~\ref{LLFF_3V} and Fig.~\ref{LLFF_6V} respectively display the rendering results of our model on LLFF with 3-view input and 6-view input.


\bibliography{ColNeRF}

\begin{thebibliography}{34}
\providecommand{\natexlab}[1]{#1}

\bibitem[{Chen et~al.(2021)Chen, Xu, Zhao, Zhang, Xiang, Yu, and
  Su}]{chen2021mvsnerf}
Chen, A.; Xu, Z.; Zhao, F.; Zhang, X.; Xiang, F.; Yu, J.; and Su, H. 2021.
\newblock MVSNeRF: Fast Generalizable Radiance Field Reconstruction from
  Multi-View Stereo.
\newblock In \emph{Proceedings of the IEEE/CVF International Conference on
  Computer Vision}, 14124--14133.

\bibitem[{Chen et~al.(2023{\natexlab{a}})Chen, Yan, Sang, Chen, Wang, Guo,
  Zhong, and Wan}]{chen2023bidirectional}
Chen, S.; Yan, B.; Sang, X.; Chen, D.; Wang, P.; Guo, X.; Zhong, C.; and Wan,
  H. 2023{\natexlab{a}}.
\newblock Bidirectional Optical Flow NeRF: High Accuracy and High Quality under
  Fewer Views.
\newblock In \emph{Proceedings of the AAAI Conference on Artificial
  Intelligence}, volume~37, 359--368.

\bibitem[{Chen et~al.(2023{\natexlab{b}})Chen, Xu, Wu, Zheng, Cham, and
  Cai}]{chen2023explicit}
Chen, Y.; Xu, H.; Wu, Q.; Zheng, C.; Cham, T.-J.; and Cai, J.
  2023{\natexlab{b}}.
\newblock Explicit Correspondence Matching for Generalizable Neural Radiance
  Fields.
\newblock \emph{arXiv preprint arXiv:2304.12294}.

\bibitem[{Chibane et~al.(2021)Chibane, Bansal, Lazova, and
  Pons-Moll}]{chibane2021stereo}
Chibane, J.; Bansal, A.; Lazova, V.; and Pons-Moll, G. 2021.
\newblock Stereo Radiance Fields (SRF): Learning View Synthesis for Sparse
  Views of Novel Scenes.
\newblock In \emph{Proceedings of the IEEE/CVF Conference on Computer Vision
  and Pattern Recognition}, 7911--7920.

\bibitem[{Deng et~al.(2021)Deng, Shi, Li, Zhou, Zhang, and Li}]{deng2021voxel}
Deng, J.; Shi, S.; Li, P.; Zhou, W.; Zhang, Y.; and Li, H. 2021.
\newblock Voxel R-CNN: Towards High Performance Voxel-Based 3D Object
  detection.
\newblock In \emph{Proceedings of the AAAI Conference on Artificial
  Intelligence}, volume~35, 1201--1209.

\bibitem[{Deng et~al.(2022)Deng, Liu, Zhu, and Ramanan}]{deng2022depth}
Deng, K.; Liu, A.; Zhu, J.-Y.; and Ramanan, D. 2022.
\newblock Depth-Supervised NeRF: Fewer Views and Faster Training for Free.
\newblock In \emph{Proceedings of the IEEE/CVF Conference on Computer Vision
  and Pattern Recognition}, 12882--12891.

\bibitem[{Fang et~al.(2023)Fang, Xu, Wang, Yang, Wang, and Zhou}]{fang2023one}
Fang, S.; Xu, W.; Wang, H.; Yang, Y.; Wang, Y.; and Zhou, S. 2023.
\newblock One is All: Bridging the Gap between Neural Radiance Fields
  Architectures with Progressive Volume Distillation.
\newblock In \emph{Proceedings of the AAAI Conference on Artificial
  Intelligence}, volume~37, 597--605.

\bibitem[{Fontaine et~al.(2022)Fontaine, Carpenter, Gross, Leon-Saval, Jung,
  Richardson, and Amezcua-Correa}]{fontaine2022photonic}
Fontaine, N.~K.; Carpenter, J.; Gross, S.; Leon-Saval, S.; Jung, Y.;
  Richardson, D.~J.; and Amezcua-Correa, R. 2022.
\newblock Photonic Lanterns, 3-D Waveguides, Multiplane Light Conversion, and
  Other Components that Enable Space-Division Multiplexing.
\newblock \emph{Proceedings of the IEEE}, 110(11): 1821--1834.

\bibitem[{He et~al.(2016)He, Zhang, Ren, and Sun}]{he2016deep}
He, K.; Zhang, X.; Ren, S.; and Sun, J. 2016.
\newblock Deep Residual Learning for Image Recognition.
\newblock In \emph{Proceedings of the IEEE Conference on Computer Vision and
  Pattern Recognition}, 770--778.

\bibitem[{Huang et~al.(2019)Huang, Lai, Xu, and Tu}]{huang20193d}
Huang, W.; Lai, B.; Xu, W.; and Tu, Z. 2019.
\newblock 3D Volumetric Modeling with Introspective Neural Networks.
\newblock In \emph{Proceedings of the AAAI Conference on Artificial
  Intelligence}, volume~33, 8481--8488.

\bibitem[{Jain, Tancik, and Abbeel(2021)}]{jain2021putting}
Jain, A.; Tancik, M.; and Abbeel, P. 2021.
\newblock Putting NeRF on a Diet: Semantically Consistent Few-Shot View
  Synthesis.
\newblock In \emph{Proceedings of the IEEE/CVF International Conference on
  Computer Vision}, 5885--5894.

\bibitem[{Jensen et~al.(2014)Jensen, Dahl, Vogiatzis, Tola, and
  Aan{\ae}s}]{jensen2014large}
Jensen, R.; Dahl, A.; Vogiatzis, G.; Tola, E.; and Aan{\ae}s, H. 2014.
\newblock Large Scale Multi-View Stereopsis Evaluation.
\newblock In \emph{Proceedings of the IEEE Conference on Computer Vision and
  Pattern Recognition}, 406--413.

\bibitem[{Kim, Seo, and Han(2022)}]{kim2022infonerf}
Kim, M.; Seo, S.; and Han, B. 2022.
\newblock InfoNeRF: Ray Entropy Minimization for Few-Shot Neural Volume
  Rendering.
\newblock In \emph{Proceedings of the IEEE/CVF Conference on Computer Vision
  and Pattern Recognition}, 12912--12921.

\bibitem[{Li et~al.(2021)Li, Feng, She, Ding, Wang, and Lee}]{li2021mine}
Li, J.; Feng, Z.; She, Q.; Ding, H.; Wang, C.; and Lee, G.~H. 2021.
\newblock MINE: Towards Continuous Depth MPI with NeRF for Novel View
  Synthesis.
\newblock In \emph{Proceedings of the IEEE/CVF International Conference on
  Computer Vision}, 12578--12588.

\bibitem[{Lombardi et~al.(2019)Lombardi, Simon, Saragih, Schwartz, Lehrmann,
  and Sheikh}]{lombardi2019neural}
Lombardi, S.; Simon, T.; Saragih, J.; Schwartz, G.; Lehrmann, A.; and Sheikh,
  Y. 2019.
\newblock Neural Volumes: Learning Dynamic Renderable Volumes from Images.
\newblock \emph{ACM Transactions on Graphics}, 38(4).

\bibitem[{Maturana and Scherer(2015)}]{maturana2015voxnet}
Maturana, D.; and Scherer, S. 2015.
\newblock VoxNet: A 3D Convolutional Neural Network for Real-Time Object
  recognition.
\newblock In \emph{Proceedings of the IEEE/RSJ International Conference on
  Intelligent Robots and Systems}, 922--928.

\bibitem[{Mildenhall et~al.(2019)Mildenhall, Srinivasan, Ortiz-Cayon,
  Kalantari, Ramamoorthi, Ng, and Kar}]{mildenhall2019local}
Mildenhall, B.; Srinivasan, P.~P.; Ortiz-Cayon, R.; Kalantari, N.~K.;
  Ramamoorthi, R.; Ng, R.; and Kar, A. 2019.
\newblock Local Light Field Fusion: Practical View Synthesis With Prescriptive
  Sampling Guidelines.
\newblock \emph{ACM Transactions on Graphics}, 38(4): 1--14.

\bibitem[{Mildenhall et~al.(2021)Mildenhall, Srinivasan, Tancik, Barron,
  Ramamoorthi, and Ng}]{mildenhall2021nerf}
Mildenhall, B.; Srinivasan, P.~P.; Tancik, M.; Barron, J.~T.; Ramamoorthi, R.;
  and Ng, R. 2021.
\newblock NeRF: Representing Scenes as Neural Radiance Fields for View
  Synthesis.
\newblock \emph{Communications of the ACM}, 65(1): 99--106.

\bibitem[{Ni et~al.(2020{\natexlab{a}})Ni, Yang, Wang, Ma, and Kwong}]{9204448}
Ni, Z.; Yang, W.; Wang, S.; Ma, L.; and Kwong, S. 2020{\natexlab{a}}.
\newblock Towards Unsupervised Deep Image Enhancement With Generative
  Adversarial Network.
\newblock \emph{IEEE Transactions on Image Processing}, 29: 9140--9151.

\bibitem[{Ni et~al.(2020{\natexlab{b}})Ni, Yang, Wang, Ma, and
  Kwong}]{10.1145/3394171.3413839}
Ni, Z.; Yang, W.; Wang, S.; Ma, L.; and Kwong, S. 2020{\natexlab{b}}.
\newblock Unpaired Image Enhancement with Quality-Attention Generative
  Adversarial Network.
\newblock In \emph{Proceedings of the 28th ACM International Conference on
  Multimedia}, 1697--1705.

\bibitem[{Niemeyer et~al.(2022)Niemeyer, Barron, Mildenhall, Sajjadi, Geiger,
  and Radwan}]{niemeyer2022regnerf}
Niemeyer, M.; Barron, J.~T.; Mildenhall, B.; Sajjadi, M.~S.; Geiger, A.; and
  Radwan, N. 2022.
\newblock RegNeRF: Regularizing Neural Radiance Fields for View Synthesis from
  Sparse Inputs.
\newblock In \emph{Proceedings of the IEEE/CVF Conference on Computer Vision
  and Pattern Recognition}, 5480--5490.

\bibitem[{Shih et~al.(2020)Shih, Su, Kopf, and Huang}]{shih20203d}
Shih, M.-L.; Su, S.-Y.; Kopf, J.; and Huang, J.-B. 2020.
\newblock 3D Photography Using Context-Aware Layered Depth Inpainting.
\newblock In \emph{Proceedings of the IEEE/CVF Conference on Computer Vision
  and Pattern Recognition}, 8028--8038.

\bibitem[{Somraj, Karanayil, and Soundararajan(2023)}]{somraj2023simplenerf}
Somraj, N.; Karanayil, A.; and Soundararajan, R. 2023.
\newblock SimpleNeRF: Regularizing Sparse Input Neural Radiance Fields with
  Simpler Solutions.
\newblock \emph{arXiv preprint arXiv:2309.03955}.

\bibitem[{Somraj and Soundararajan(2023)}]{somraj2023VipNeRF}
Somraj, N.; and Soundararajan, R. 2023.
\newblock {ViP-NeRF}: Visibility Prior for Sparse Input Neural Radiance Fields.
\newblock In \emph{Proceedings of the ACM Special Interest Group on Computer
  Graphics and Interactive Techniques}.

\bibitem[{Sun, Sun, and Chen(2022)}]{sun2022direct}
Sun, C.; Sun, M.; and Chen, H.-T. 2022.
\newblock Direct Voxel Grid Optimization: Super-Fast Convergence for Radiance
  Fields Reconstruction.
\newblock In \emph{Proceedings of the IEEE/CVF Conference on Computer Vision
  and Pattern Recognition}, 5459--5469.

\bibitem[{Tulsiani, Tucker, and Snavely(2018)}]{tulsiani2018layer}
Tulsiani, S.; Tucker, R.; and Snavely, N. 2018.
\newblock Layer-Structured 3D Scene Inference via View Synthesis.
\newblock In \emph{Proceedings of the European Conference on Computer Vision},
  302--317.

\bibitem[{Wang et~al.(2023)Wang, Chen, Loy, and Liu}]{wang2023sparsenerf}
Wang, G.; Chen, Z.; Loy, C.~C.; and Liu, Z. 2023.
\newblock Sparsenerf: Distilling depth ranking for few-shot novel view
  synthesis.
\newblock \emph{arXiv preprint arXiv:2303.16196}.

\bibitem[{Wang et~al.(2021)Wang, Wang, Genova, Srinivasan, Zhou, Barron,
  Martin-Brualla, Snavely, and Funkhouser}]{wang2021ibrnet}
Wang, Q.; Wang, Z.; Genova, K.; Srinivasan, P.~P.; Zhou, H.; Barron, J.~T.;
  Martin-Brualla, R.; Snavely, N.; and Funkhouser, T. 2021.
\newblock IBRNet: Learning Multi-View Image-Based Rendering.
\newblock In \emph{Proceedings of the IEEE/CVF Conference on Computer Vision
  and Pattern Recognition}, 4690--4699.

\bibitem[{Wang et~al.(2004)Wang, Bovik, Sheikh, and Simoncelli}]{wang2004image}
Wang, Z.; Bovik, A.~C.; Sheikh, H.~R.; and Simoncelli, E.~P. 2004.
\newblock Image Quality Assessment: From Error Visibility to Structural
  Similarity.
\newblock \emph{IEEE Transactions on Image Processing}, 13(4): 600--612.

\bibitem[{Xu, Zhong, and Neumann(2022)}]{xu2022behind}
Xu, Q.; Zhong, Y.; and Neumann, U. 2022.
\newblock Behind the Curtain: Learning Occluded Shapes for 3D Object Detection.
\newblock In \emph{Proceedings of the AAAI Conference on Artificial
  Intelligence}, volume~36, 2893--2901.

\bibitem[{Yang, Pavone, and Wang(2023)}]{yang2023freenerf}
Yang, J.; Pavone, M.; and Wang, Y. 2023.
\newblock FreeNeRF: Improving Few-Shot Neural Rendering with Free Frequency
  Regularization.
\newblock In \emph{Proceedings of the IEEE/CVF Conference on Computer Vision
  and Pattern Recognition}, 8254--8263.

\bibitem[{Yu et~al.(2021)Yu, Ye, Tancik, and Kanazawa}]{yu2021pixelnerf}
Yu, A.; Ye, V.; Tancik, M.; and Kanazawa, A. 2021.
\newblock PixelNeRF: Neural Radiance Fields from One or Few Images.
\newblock In \emph{Proceedings of the IEEE/CVF Conference on Computer Vision
  and Pattern Recognition}, 4578--4587.

\bibitem[{Zhang et~al.(2018)Zhang, Isola, Efros, Shechtman, and
  Wang}]{zhang2018unreasonable}
Zhang, R.; Isola, P.; Efros, A.~A.; Shechtman, E.; and Wang, O. 2018.
\newblock The Unreasonable Effectiveness of Deep Features as a Perceptual
  Metric.
\newblock In \emph{Proceedings of the IEEE Conference on Computer Vision and
  Pattern recognition}, 586--595.

\bibitem[{Zhu, Xie, and Fang(2018)}]{zhu2018learning}
Zhu, J.; Xie, J.; and Fang, Y. 2018.
\newblock Learning Adversarial 3D Model Generation with 2D Image Enhancer.
\newblock In \emph{Proceedings of the AAAI Conference on Artificial
  Intelligence}, volume~32, 7615–7622.

\end{thebibliography}

\end{document}